\documentclass{article}

\usepackage{microtype}
\usepackage{graphicx}
\usepackage{subfigure}
\usepackage{booktabs} 
\usepackage{placeins}
\usepackage{amsthm,amsmath,amssymb}
\usepackage{mathrsfs}
\usepackage{multirow}
\usepackage{hyperref}

\newtheorem{theorem}{Theorem}
\newtheorem{assumption}{Assumption}
\newtheorem{definition}{Definition}
\newtheorem{corollary}{Corollary}

\usepackage{hyperref}

\usepackage[accepted]{icml2021}

\icmltitlerunning{Asymmetric Loss Functions for Learning with Noisy Labels}

\begin{document}
\twocolumn[
\icmltitle{Asymmetric Loss Functions for Learning with Noisy Labels}




\begin{icmlauthorlist}
\icmlauthor{Xiong Zhou}{hit,pcl}
\icmlauthor{Xianming Liu}{hit,pcl}
\icmlauthor{Junjun Jiang}{hit,pcl}
\icmlauthor{Xin Gao}{kaust,pcl}
\icmlauthor{Xiangyang Ji}{thu}
\end{icmlauthorlist}

\icmlaffiliation{hit}{Harbin Institute of Technology}
\icmlaffiliation{pcl}{Peng Cheng Laboratory}
\icmlaffiliation{kaust}{King Abdullah University of Science and Technology}
\icmlaffiliation{thu}{Tsinghua University}

\icmlcorrespondingauthor{Xianming Liu}{csxm@hit.edu.cn}

\icmlkeywords{Machine Learning, ICML}

\vskip 0.3in
]
 


\printAffiliationsAndNotice{}  

\begin{abstract}
Robust loss functions are essential for training deep neural networks with better generalization power in the presence of noisy labels. Symmetric loss functions are confirmed to be robust to label noise. However, the symmetric condition is overly restrictive. In this work, we propose a new class of loss functions, namely \textit{asymmetric loss functions}, which are robust to learning with noisy labels for various types of noise. We investigate general theoretical properties of asymmetric loss functions, including classification calibration, excess risk bound, and noise tolerance. Meanwhile, we introduce the asymmetry ratio to measure the asymmetry of a loss function. The empirical results show that higher ratio would provide better noise tolerance. Moreover, we modify several commonly-used loss functions and establish the necessary and sufficient conditions for them to be asymmetric. Experimental results on benchmark datasets demonstrate that asymmetric loss functions can outperform state-of-the-art methods. The code is available at \href{https://github.com/hitcszx/ALFs}{https://github.com/hitcszx/ALFs}
\end{abstract}

\section{Introduction}
The success of deep neural networks based supervised learning largely relies on massive high-quality labeled data. However, in practice, the annotation process inevitably introduces wrong labels, due to the lack of experts involved or data from public crowdsourcing platforms \cite{liu2011noise, arpit2017closer}. Empirical studies show that over-parameterized deep networks can even fit random labels \cite{2017Understanding}. When samples are mis-labeled, the network would memorize wrong patterns, leading to impaired performance in the subsequent inference tasks. Accordingly, robust learning of 
classifier in the presence of label noise has received a lot of attention.

To alleviate the impact of label noise to classifier learning, one popular research line is to design noise-tolerant loss functions. This approach has been pursued in a large body of work \cite{nonconvex, Wang2019ImprovingMA, liu2020peer, Lyu2020CurriculumLR, menon2020can, Feng2020CanCE} that embraces new losses, especially symmetric loss functions and their variants \cite{Manwani, unhingedloss, symmetric, GCE, sce, ma2020normalized}. 

Symmetric loss functions were proposed as a sufficient condition such that the risk minimization with respect to the loss becomes noise-tolerant for binary classification \cite{Manwani}. Subsequently, the unhinged loss \cite{unhingedloss}, which is equivalent to a scaled Mean Absolute Error (MAE) \cite{symmetric}, was proved to be the only convex loss function that is strongly robust for symmetric label noise (SLN). Ghosh \textit{et al.} \cite{symmetric} theoretically demonstrated that a loss function would be inherently tolerant to SLN as long as it satisfies the symmetric condition. The sufficient condition was then extended for multi-class classification \cite{symmetric} and was emphasized in the BER minimization and AUC maximization from corrupted labels \cite{BERAUC}. However, MAE treats every sample equally, leading to significantly longer training time before convergence. This drawback motivates some works to improve MAE, which follows the principle of combining the robustness of MAE and the fast convergence of Cross Entropy (CE). For instance, \cite{GCE} advocated the use of a more general class of noise-robust loss functions, called Generalized Cross Entropy (GCE), which encompasses both MAE and CE. Inspired by the symmetric KL-divergence, the symmetric cross entropy (SCE) \cite{sce} was proposed to combine CE with 
a noise tolerance term, namely Reverse Cross Entropy (RCE). Ma \textit{et al.} \cite{ma2020normalized} theoretically demonstrated that by applying a simple normalization, any loss can be made robust to noisy labels. However, the normalized loss functions are not sufficient to train accurate DNNs and are prone to encounter the gradient explosion problem. 

From the above review, it can be found that the fitting ability of the existing symmetric loss functions is restricted by the symmetric condition \cite{GCE, BERAUC}. However, the symmetric condition is too stringent to find a convex loss function \cite{pmlr-v37-plessis15, GHOSH201593, unhingedloss}, leading to difficulties in optimization. Thus, learning with symmetric loss function usually suffers from underfitting issues.

In this paper, we propose a new class of robust loss functions, namely \textit{asymmetric loss functions}, which are tailored to satisfy that the Bayes-optimal prediction under the loss is a point-mass on the highest scoring label, \textit{i.e.}, the loss has Bayes-optimal prediction that matches that of the 0-1 loss. Specifically, our scheme is based on a reasonable assumption that in a training dataset samples have higher probability to be annotated with true semantic labels than any other class labels. According to this \textit{clean-labels-domination assumption}, the proposed asymmetric loss is derived. It indicates that, minimizing the $L$-risk under noisy case, which can be formulated as the weighted form, would make the optimization direction shift to the loss term with the maximum weight. In this way, the contribution of noisy labels in the process of classifier learning is eliminated, and thus the proposed asymmetric loss functions are inherently noise-tolerant.  Furthermore, we offer a complete theoretical analysis about the properties of asymmetric loss functions, including classification calibration, excess risk bound, noise-tolerance, and asymmetry ratio. We show that several commonly-used loss functions can be modified to be asymmetric, and establish the corresponding necessary and sufficient conditions for them. The main contributions of our work are highlighted as follows:
\begin{itemize}
    \item We propose a new family of robust loss functions, namely asymmetric loss functions, which are noise-tolerant with an appropriate model for various types of noise. We theoretically prove that completely asymmetric losses, which include symmetric losses as a special case, are classification-calibrated, and have an excess risk bound when they are strictly asymmetric.
    \item We introduce the asymmetric ratio to measure the asymmetry of a loss function, which, together with the clean level of labels, can be associated with noise-tolerance. The empirical results show that higher ratio will provide better noise robustness.
    \item We generalize several commonly-used loss functions, and establish the necessary and sufficient conditions for them to be asymmetric. The experimental results demonstrate that the new loss functions can outperform the state-of-the-art methods.
\end{itemize}

\section{Preliminaries}
\subsection{Risk Minimization}
 Define $\mathcal X\subset \mathbb R^d$ as the feature space from which the samples are drawn, and $\mathcal Y=[k]=\{1,...,k\}$  as the class label space, \textit{i.e.}, we consider a $k$-classification problem, where $k\ge 2$. In an ideal classifier learning problem, we are given a clean training set, $\mathcal S=\{(\mathbf x_1,y_1),...,(\mathbf x_{N}, y_{N})\}$, where $(\mathbf x_i, y_i)$ is drawn \textit{i.i.d.} from an unknown distribution $\mathcal D$ over $\mathcal X\times\mathcal Y$. The classifier is a mapping function from feature space to label space $h(\mathbf x)=\arg\max_i f(\mathbf x)_i$, where $f:\mathcal X\rightarrow \mathcal C$, $\mathcal C\subseteq [0,1]^k$, $\forall\ \mathbf c\in\mathcal C$, $\mathbf{1}^T\mathbf c=1$. $f(\mathbf x)$ denotes an approximation of $p(\cdot|\mathbf x)$, which is considered as a neural network ending with a softmax layer in this work.

We define a loss function as a mapping $L:\mathcal C\times \mathcal Y\rightarrow \mathbb{R}$, where $\arg\min_{\mathbf u\in \mathcal C}L(\mathbf u,y)=\mathbf{e}_y$, $L(\mathbf u,y)$ is monotonically decreasing on the prediction probability $u_y$ of class $y$, and $\mathbf e_y$ denotes a one-hot vector. The $L$-risk for the hypothesis $f$ is defined as
\begin{equation}
\label{L-risk}
    R_L(f)=
    \mathbb{E}_{\mathcal D}[L(f(\mathbf x),y)]=\mathbb E_{\mathbf x,y}[L(f(\mathbf x), y)],
\end{equation}
where $\mathbb E$ is denoted as expectation operator. Under the risk minimization framework, our objective is to learn a optimal classifier, $f^*$, which is a global minimum of $R_L(f)$.

\subsection{Label Noise Model}
The annotation process inevitably introduces label noise, the model of which can be formulated as
\begin{equation}
\label{noisemodel}
    \tilde{y}_n=
    \begin{cases}
     i,\ i\in[k],\ i\not = y_n & \text{with probability }\eta_{\mathbf x_n,i}\\
     y_n & \text{with probability }(1-\eta_{\mathbf x_n})
    \end{cases},
\end{equation}
where $\eta_{\mathbf x_n, i}$ denotes the probability of flipping the true label $y_n$ into $i$ for $\mathbf x_n$, and $\eta_{\mathbf x_n}=\sum_{i\not = y_n}\eta_{\mathbf x_n,i}$ denotes the noise ratio of $\mathbf x_n$. This noise model shows that a realistic corruption probability is dependent on both data features and class labels \cite{2015Learning, goldberger2016training}, and this kind of noise is called \textit{instance-} and \textit{label-dependent} noise \cite{Cheng2020LearningWB}. However, this modeling approach of label noise has not been investigated extensively yet due to its complexity.

Instead, a popular approach for modeling label noise simply assumes that the corruption process is conditionally \textit{independent} of data features when the true label is given \cite{NIPS2013_3871bd64}, \textit{i.e.}, $\eta_{\mathbf x_n}$ and $\eta_{\mathbf x_n,i}$ are only dependent on the class labels, which can be then represented as a label transition matrix. If $\eta_{\mathbf x_n,i}=\eta$, $\forall \mathbf x_n, i$, the noise is called  \textit{symmetric} (or \textit{uniform}) noise, where a true label is flipped into other labels with equal probability. In contrast to symmetric noise, another type of noise is called \textit{asymmetric} if $\forall n,\ \eta_{\mathbf x_n}=\eta$ and $\exists i\not=y_n,\ \forall j\not=y_n\&i,\ \eta_{\mathbf x_n,i}>\eta_{\mathbf x_n,j}$, \textit{i.e.}, a certain class is more likely to be wrongly annotated into a particular label \cite{2020Learning}.

Based on the label noise model, the $L$-risk under noisy case can be formulated as
\vskip-15pt
$$
R_L^\eta(f)=\mathbb E_{\mathcal D}\big[(1-\eta_{\mathbf x})L(f(\mathbf x),y)+\sum_{i\not = y}\eta_{\mathbf x, i} L(f(\mathbf x),i)
\big].
$$
\vskip-8pt
It can be found that, due to the presence of noisy labels, the classifier learning process is influenced by $\sum_{i\not =y}\eta_{\mathbf x,i}L(f(\mathbf x),i)$, \textit{i.e.}, noisy labels would degrade the generalization performance of deep neural networks.
Define $f^*_\eta$ be the global minimum of $R_L^\eta(f)$, then $L$ is noise-tolerant if $f^*_\eta$ is also the global minimum of $R_L(f)$

\subsection{Symmetric Loss Functions}
The most popular family of loss functions in robust learning is symmetric loss \cite{Manwani, symmetric}. A loss is called symmetric if it satisfies
\begin{equation}
\label{symmetric-condition}
    \sum_{i=1}^k L(f(\mathbf x), i)=C,\ \forall x\in\mathcal X, \forall f,
\end{equation}
where $C$ is a constant value. Ghosh \textit{et al.} \cite{symmetric} proved that, for a $k$-classification problem, if the loss $L$ is symmetric and the noise ratio $\eta<\frac{k-1}{k}$, then under symmetric noise $L$ is noise-tolerant. Moreover, if $R_L(f^*)=0$, the loss function is also noise-tolerant under asymmetric noise, where $f^*$ is a global minimizer of $R_L$.

One of the most classic symmetric loss functions is MAE \cite{symmetric}, which is defined as $L(\mathbf u, i)=\|\mathbf e_i-\mathbf u\|_1=2-2u_i$ and obviously satisfies $\sum_i L(\mathbf u,i)=2k-2$. Reverse Cross Entropy (RCE) proposed in \cite{sce} is also belonging to the kind of symmetric loss, which is actually the variant of MAE. Ma \textit{et al.} \cite{ma2020normalized} proposed the normalized loss functions, which can make any loss symmetric by using a simple normalization operation. However, the symmetric condition is too stringent to find a convex loss function \cite{pmlr-v37-plessis15, GHOSH201593, unhingedloss}, leading to difficulties in optimization. Thus, learning with symmetric loss function usually suffers from the underfitting effect. 

In this paper, we propose a new family of loss functions, \textit{asymmetric loss functions}, which includes symmetric loss functions as its special case. More importantly, the proposed asymmetric loss family also guarantees some desirable properties and contain many convex loss functions, which facilitate the subsequent optimization process.

\section{Asymmetric Loss Functions}
In this section, we introduce in details the proposed asymmetric loss functions.  Firstly, we state the \textit{clean-labels-domination assumption}, which serves as the fundamental basic in the subsequent derivation. Then we introduce the proposed asymmetric loss functions, a new class of robust loss function, which achieve robust learning by keep consistency between the Bayes-optimal prediction of the loss and that of the 0-1 loss. Subsequently, we theoretically explore general properties of asymmetric loss functions, including classification calibration, excess risk bound, noise tolerance, and asymmetry ratio. Finally, we show that several commonly-used loss functions can be modified to be asymmetric and thus robust to label noise. The necessary and sufficient conditions are offered for them. The detailed proofs for theorems and corollaries can be found in the supplementary material.

\subsection{Clean-labels-domination Assumption}
For robust learning, it is reasonable to assume that in a training dataset samples have higher probability to be annotated with true semantic labels than any other class labels, which is referred to as \textit{clean-labels-domination assumption}.
In the following, we first provide the formal definition of class-wise clean-labels-domination.
\begin{definition}
Given an underlying clean dataset $\mathcal S$, the corresponding observed noisy dataset is $\tilde{\mathcal S}$. The $i$-th class subset of $\tilde{\mathcal S}$ is formulated as $\tilde{\mathcal S}_i=\{(\mathbf x,\tilde{y}): y=i, (\mathbf x,\tilde{y})\in \tilde{\mathcal S}, (\mathbf x,{y})\in {\mathcal S}\}$, with $i\in[k]$. We define that the class label $i$ is dominant in $\tilde{\mathcal S}_i$ if it satisfies
\begin{equation}
    \sum_{(\mathbf x, \tilde y)\in \tilde{\mathcal S}_i}\mathbb I(\tilde{y}=i)>\max_{j\not = i}\sum_{(\mathbf x, \tilde y)\in \tilde{\mathcal S}_i}\mathbb I(\tilde{y}=j),
\end{equation}
where $\mathbb{I}(\cdot)$ is the identity function. 
\end{definition}

The dataset $\tilde{\mathcal S}$ is claimed to be clean-labels-dominant if in all classes correct labels are dominant. In real-world datasets, the noise ratio of noisy labels is reported to range from 8.0\% to 38.5\% \cite{2020Learning}, which serves as the corroboration that $\tilde{\mathcal S}$ is usually clean-labels-dominant. Based on this empirical observation, we further assume that the label noise model defined in (\ref{noisemodel}) is clean-labels-dominant:
\begin{assumption}
\label{dominate}
The label noise model is clean-labels-dominant, \textit{i.e.}, it satisfies that $\forall\mathbf{x}$, $1-\eta_{\mathbf x}>\max_{j\not=y} \eta_{\mathbf x, j}$.
\end{assumption}
Compared with the symmetric noise assumption behind symmetric losses \cite{symmetric}, \textit{i.e.}, $1-\eta_{\mathbf x}>\frac{1}{k}$, although Assumption \ref{dominate} is more restrictive, it makes sense in general and applicable to most of the real-world applications. Specifically, without the help of any prior knowledge, if there exists an approach that can help to learn a correct classifier on clean-labels-non-dominant cases, then it would fail in learning a correct classifier on clean-labels-dominant cases since the learned classifier tends to classify a sample into a non-dominant class rather than the corresponding dominant class (\textit{i.e}., the true class).

\subsection{Asymmetric Loss Functions}
For a sample $(\mathbf x, y)$ drawn from $\mathcal D$, we have the \textit{conditional $L$-risk}  \cite{2006Convexity}:
\vskip-20pt
$$
L^\eta(f(\mathbf x), y)=(1-\eta_{\mathbf x})L(f(\mathbf x),y)+\sum_{i\not = y}\eta_{\mathbf x, i} L(f(\mathbf x),i).
$$
\vskip-10pt
The exact values of $\{\eta_{\mathbf x,i}\}_{i\not =y}$ are usually unknown, and what we only know is $1-\eta_{\mathbf x}>\max_{i\not=y}\eta_{\mathbf x,i}$ according to Assumption \ref{dominate}. Our purpose is to find a simple and elegant formulation of $L$ such that minimizing the risk leads to a classifier with the same probability of mis-classification as the noise-free case. To this end, in this work, we suggest a new class of loss functions defined as follows:

\begin{definition}
\label{asymmetric-loss-function}
On the given weights $w_1,...,w_k\ge 0$, where $\exists t\in[k]$, s.t., $w_t>\max_{i\not = t}w_i$,
a loss function $L(\mathbf u, i)$ is called asymmetric if $L$ satisfies
\vskip-15pt
\begin{equation}
    \label{asymmetric-condition}
    \mathop{\arg\min}\limits_{\mathbf u}\sum_{i=1}^k w_i L(\mathbf u, i)= \mathop{\arg\min}\limits_{\mathbf u} L(\mathbf u,t),
\end{equation}
\vskip-8pt
where we always have $\mathop{\arg\min}\limits_{\mathbf u} L(\mathbf u,t)=\mathbf e_t$.
\end{definition}
We define that $L$ is \textit{asymmetric} on the label noise model that satisfies Assumption \ref{dominate}, if $L$ is asymmetric on $\{1-\eta_{\mathbf x}\}\cup\{\eta_{\mathbf x,i}\}_{i\not=y}$, $\forall (\mathbf x, y)$ drawn from $\mathcal D$. $L$ is called \textit{completely asymmetric}, if $L$ is asymmetric on any weights $w_1,...,w_k\ge 0$ that contain a unique maximum. And we call $L$ \textit{strictly asymmetric}, if it satisfies $\sum_{i=1}^k w_i L(\mathbf u', i)<\sum_{i=1}^k w_i L(\mathbf u,i)$,  $\forall\ w_1,...,w_k\ge 0$ with a unique maximum $w_t$, and $\forall\ \mathbf u',\mathbf u\in\mathcal C$, $u_t'>u_t$.

The asymmetry is reflected by the fact that minimizing the weighted risk would make the optimization direction shift to the loss term with the maximum weight. This strategy is referred to as \textit{the-largest-takes-all}.

More specifically, according to Definition \ref{asymmetric-loss-function} and Assumption \ref{dominate}, asymmetric loss functions are inherently noise-tolerant, which can eliminate the contribution of noisy labels (\textit{i.e.}, $\sum_{i\not = y}\eta_{\mathbf x, i} L(f(\mathbf x),i)$) in the process of classifier learning. It is desirable since it provides an approach of obtaining the minimum for noise-free case $L(\mathbf u, t)$ from the minimization for noisy case $\sum_{i=1}^k w_i L(\mathbf u, i)$. In other words, the asymmetric loss are tailored to satisfy that the Bayes-optimal prediction under the loss is a point-mass on the highest scoring label, \textit{i.e.}, the loss has Bayes-optimal prediction that matches that of the 0-1 loss.

\subsection{Properties of Asymmetric Loss Functions}
Let $L(\mathbf u,i)$ be asymmetric on the label noise model which is clean-labels-dominant. According to the asymmetric condition (\ref{asymmetric-condition}), it can be derived that $(1-\eta_{\mathbf x})L(\mathbf u,y)+\sum_{i\not = y}\eta_{\mathbf x, i} L(\mathbf u,i)\ge (1-\eta_{\mathbf x})L(\mathbf u^*,y)+\sum_{i\not = y}\eta_{\mathbf x, i} L(\mathbf u^*,i)$, where $\mathbf u^*=\mathbf e_y$, and the equality holds if and only if $\mathbf u=\mathbf u^*$. This inequality reveals a beautiful property for binary classification as follows:
\begin{theorem}[Classification calibration]
\label{classification-calibration}
Completely asymmetric loss functions are classification-calibrated.
\end{theorem}
Classification calibration is known to be a minimal requirement of a loss function for the binary classification task \cite{2003Statistical, 2006Convexity}. We say that $\phi$ is classification-calibrated if driving the excess risk over the Bayes-optimal predictor for $\phi$ to zero also drives the excess risk for 0-1 loss to zero. Actually, the conditional risk minimizer of $L$ is equivalent to the Bayes-optimal classifier $\mathbb{I}(\eta_{\mathbf x}>\frac{1}{2})$ (see more in supplementary materials).

\begin{figure}[ht]
    \centering
    \subfigure[AGCE]{
        \label{fig:agce-cc}
        \includegraphics[width=1.4in]{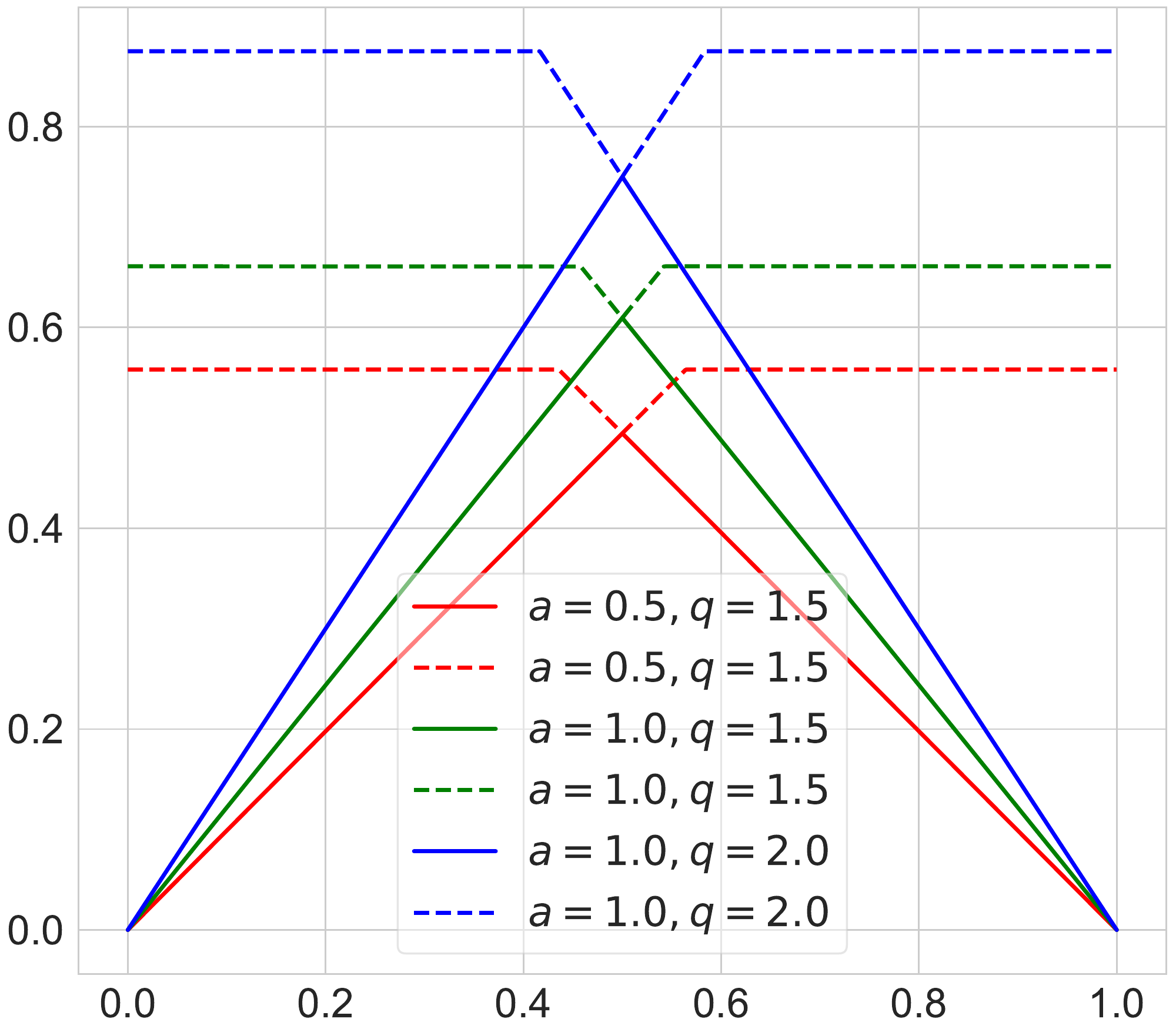}
    }
    \subfigure[AUL]{
        \label{fig:aul-cc}
        \includegraphics[width=1.4in]{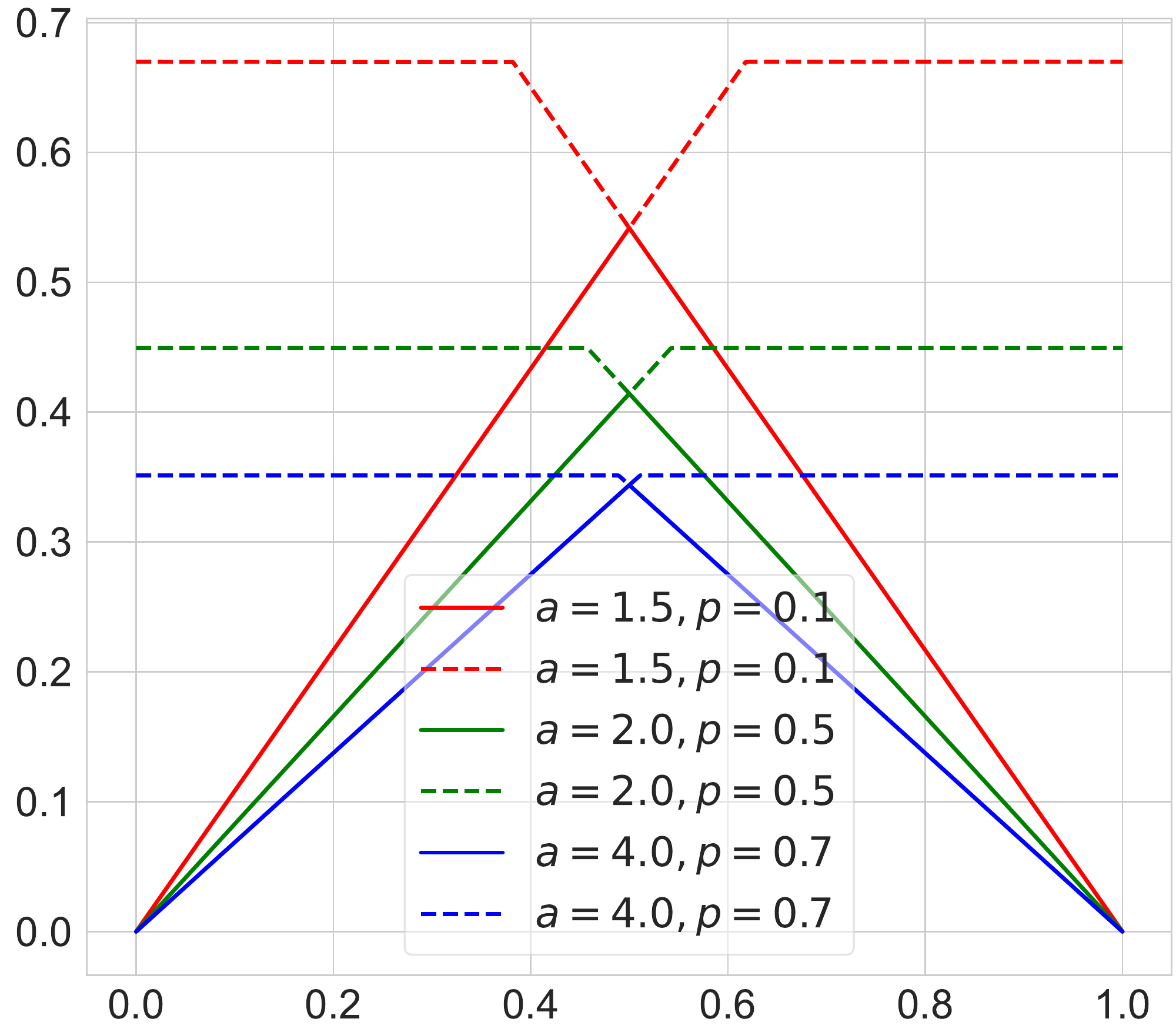}
    }
    \vskip-10pt
    \caption{Verification of classification calibration. Solid and dashed lines denote the curve of $H_{\ell}(\eta)$ and $H_{\ell}^-(\eta)$, respectively. As can be observed, the curve of $H_{\ell}^-$ is always above that of $H_{\ell}$, \textit{i.e.}, the loss functions are classification-calibrated.}
    \label{fig:classificatoin-calibration}
    \vskip-7pt
\end{figure}

Another essential property is \textit{excess risk bound} \cite{2006Convexity}, which provides a relationship between the excess risk of minimizing the mis-classification risk w.r.t the 0-1 loss and the surrogate loss. The following theorem indicates an excess bound for any strictly and completely asymmetric loss functions.

\begin{figure*}[tb]
    \centering
    \subfigure[AGCE ($q<1$)]{
    \label{fig:AGCE-nc-0}
    \includegraphics[width=1.2in]{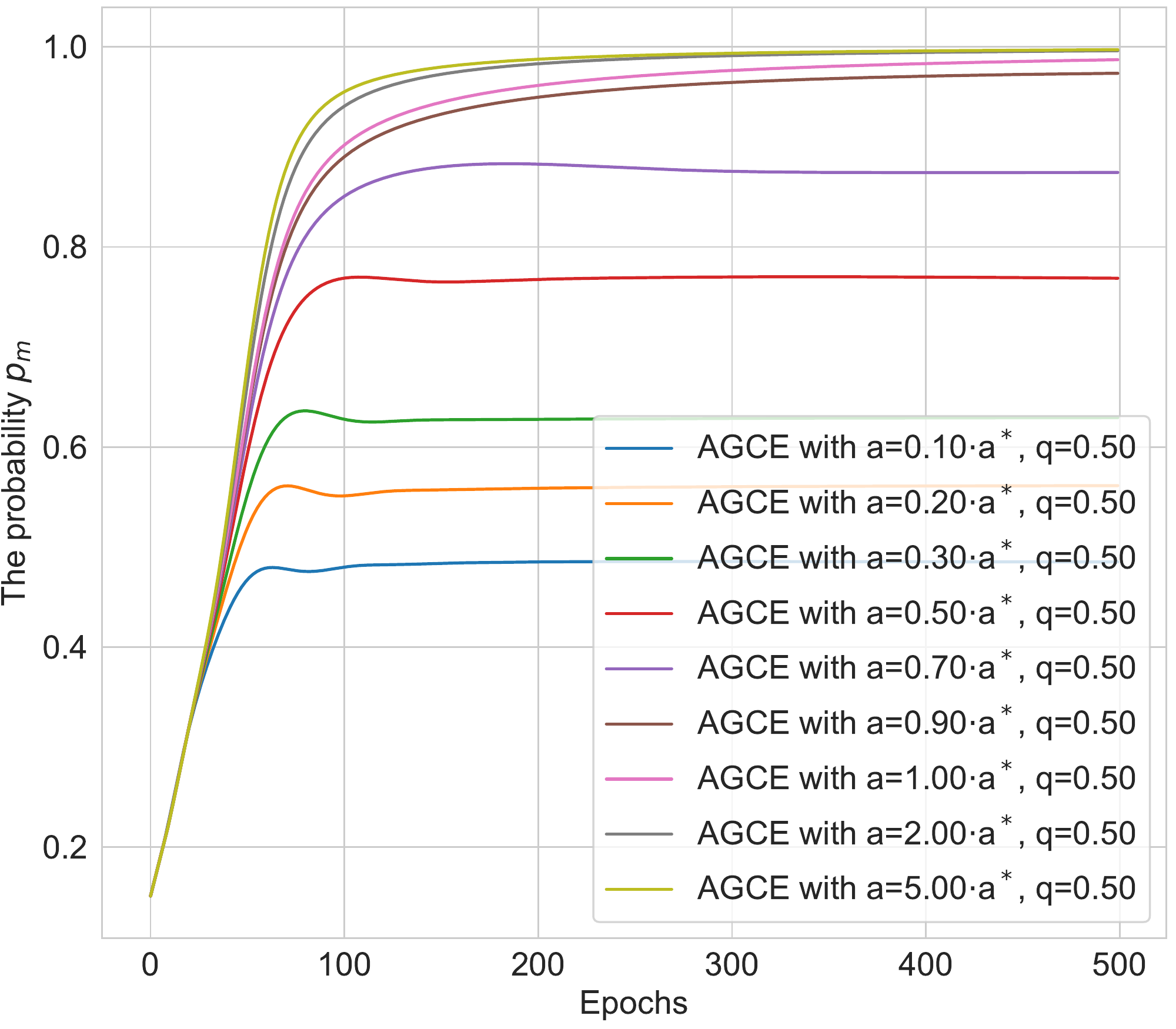}
    }
    \subfigure[AGCE ($q>1$)]{
    \label{fig:AGCE-nc-1}
    \includegraphics[width=1.2in]{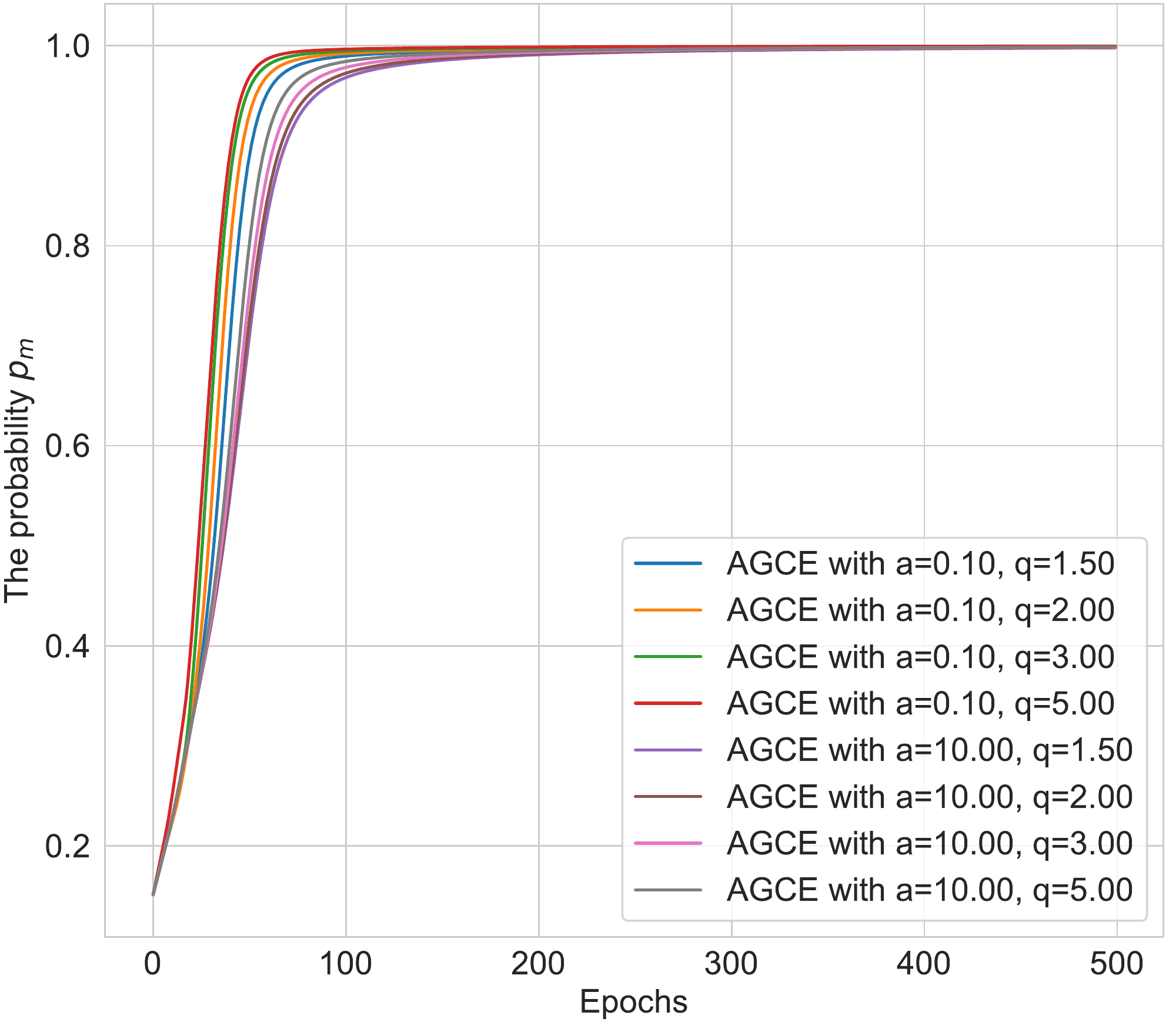}
    }
    \subfigure[AUL ($p<1$)]{
    \label{fig:AUL-nc-0}
    \includegraphics[width=1.2in]{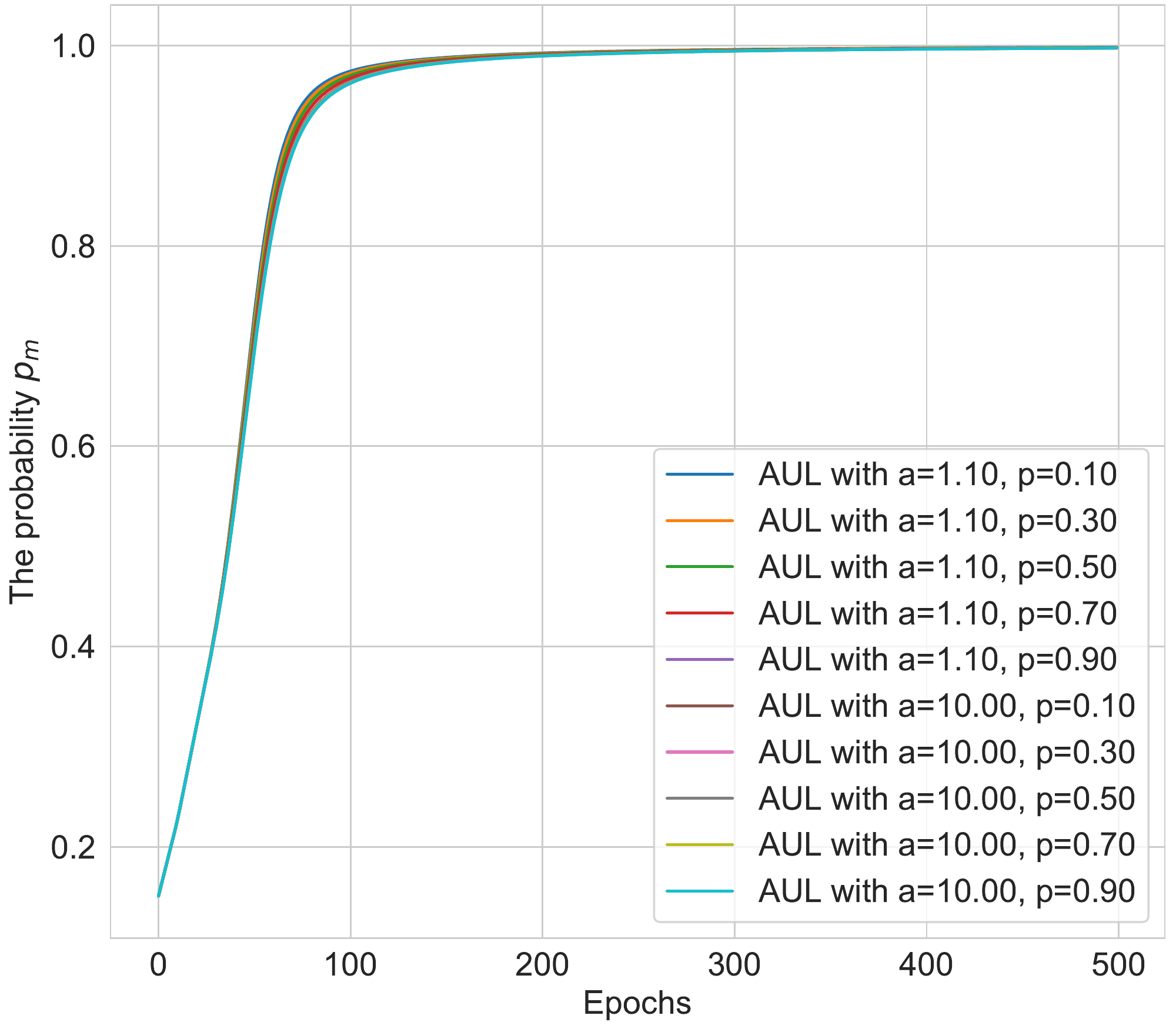}
    }
    \subfigure[AUL ($p>1$)]{
    \label{fig:AUL-nc-1}
    \includegraphics[width=1.2in]{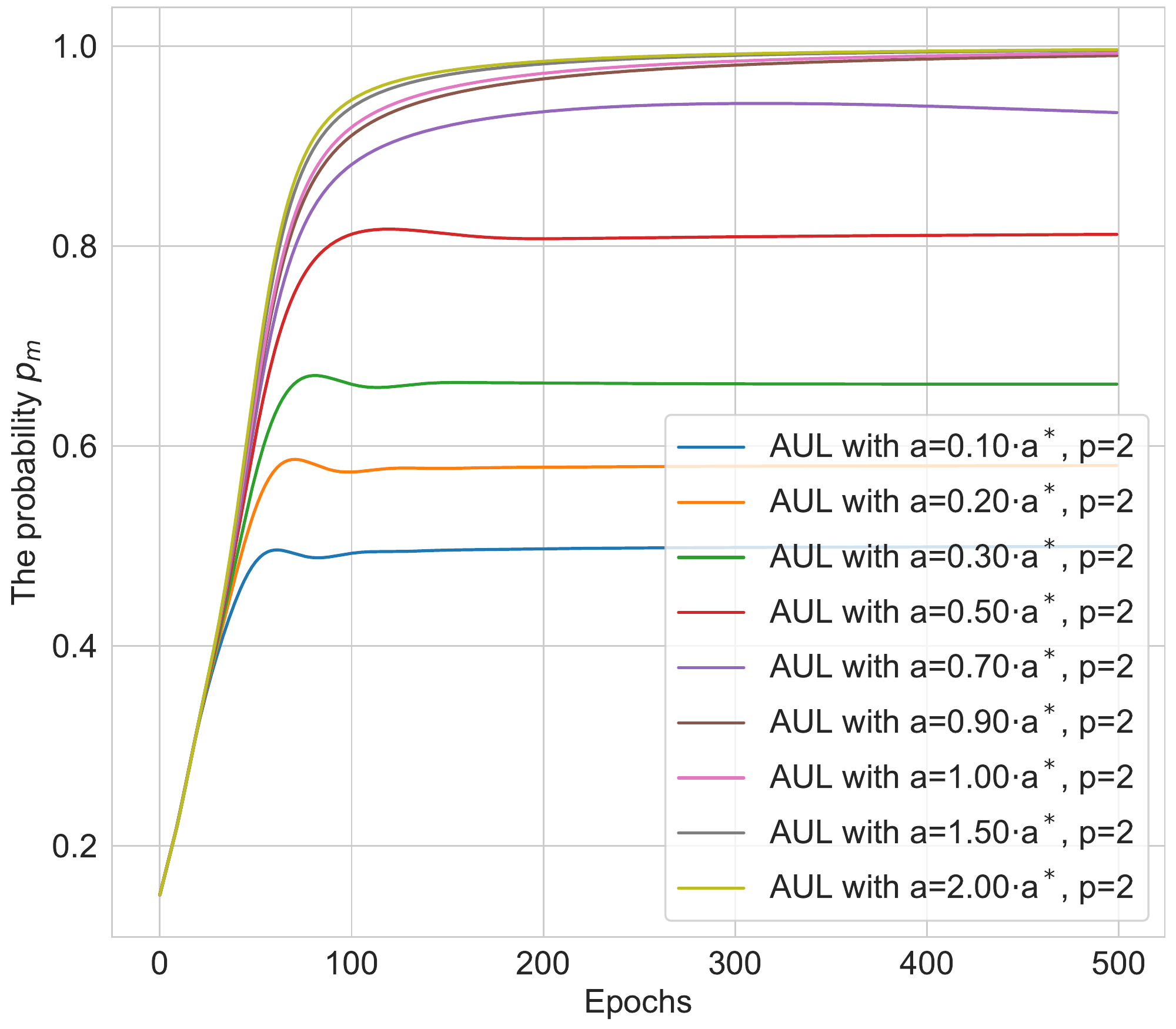}
    }
    \subfigure[AEL]{
    \label{fig:AEL-nc}
    \includegraphics[width=1.2in]{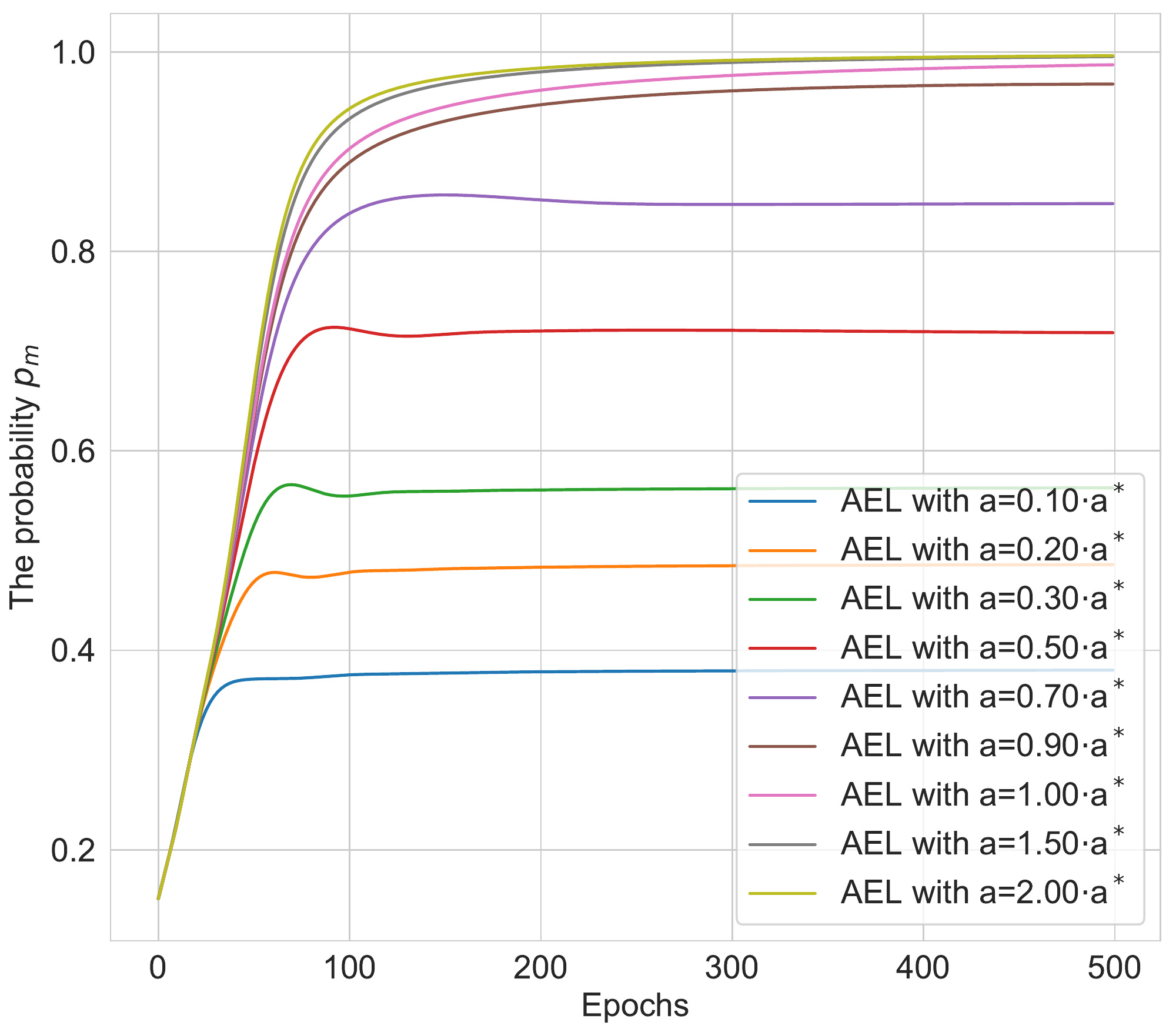}
    }
    \vskip-10pt
    \caption{The validation for necessary and sufficient conditions of AGCE, AUL and AEL, where $m=\arg\max_{i}w_i$, $n=\arg\max_{i\not =m} w_i$, and $a^*$ is the value such that $\frac{w_m}{w_n}\cdot r(\ell)= 1$ for different loss functions.}
    \label{fig:NC}
    \vskip-10pt
\end{figure*}

\begin{theorem}[Excess risk bound]
An excess risk bound of a strictly and completely asymmetric loss function $L(\mathbf u, i)=\ell(u_i)$ can be expressed as
\begin{equation}
    R_{\ell_{0-1}}(f)-R_{\ell_{0-1}}^{*}\le \frac{2(R_{\ell}(f)-R_{\ell}^{*})}{\ell(0)-\ell(1)},
\end{equation}
where $R_{\ell_{0-1}}^{*}=\inf_{g} R_{\ell_{0-1}}(g)$ and $R_{\ell}^{*}=\inf_{g} R_{\ell}(g)$.
\end{theorem}
The result suggests that the excess risk bound of any strictly and completely asymmetric loss function is controlled only by the difference of $\ell(0)-\ell(1)$. 
Intuitively, the excess risk bound shows that if the hypothesis $f$ minimizes the surrogate risk $R_\ell(f)=R^{*}_\ell$, then $f$ must also minimize the mis-classification risk $R_{\ell_{0-1}}(f)=R_{\ell_{0-1}}^{*}$.

As aforementioned, symmetric loss functions are well-studied  with general properties
\cite{Manwani, symmetric, BERAUC}. Here we reveal the relationship between symmetric loss functions and asymmetric loss functions.

\begin{theorem}
\label{sy-is-asy}
Symmetric loss functions are completely asymmetric.
\end{theorem}
An important condition for symmetric loss functions to be noise-tolerant under asymmetric noise is $R_L(f^*)=0$, i.e., there exists a hypothesis can fit the distribution $\mathcal D$ perfectly. Here we use deep networks as the hypothesis class to obtain enough fitting ability
\cite{2017Understanding, NEURIPS2019_6a61d423}.
\begin{assumption}
\label{universal}
Given the loss function $L$ and a separable distribution $\mathcal D$, we assume that there exists a hypothesis $f:\mathcal X\rightarrow \mathcal C$, $f\in\mathcal H_{net}$, $\forall (\mathbf x, y)$ drawn from $\mathcal D$, such that $f$ minimizes $L(f(\mathbf x), y)$.
\end{assumption}

To satisfy this assumption, the hypothesis class $\mathcal H_{net}$ should be as universal as possible to approximate complex functions. According to the universal approximation theorem \cite{cybenko1989approximation, Martin}, if a certain deep network model is employed, $\mathcal H_{net}$ will be a universal hypothesis class and thus contains the optimal function.

\begin{theorem}[Noise tolerance]
\label{robust}
In a multi-classification problem, given an appropriate neural network class $\mathcal H$ which satisfies Assumption \ref{universal}, the loss function $L$ is noise-tolerant if $L$ is asymmetric on the label noise model.
\end{theorem}

This theorem shows that noise tolerance can be obtained without knowing the exact noise rates when the loss is asymmetric on the label noise model. This conclusion does not depend on the data distribution. We just require that the label noise model is clean-labels-dominant and there is a neural network which is as universal as possible. Therefore, the key question becomes how to design a loss function being asymmetric on the label noise model. Moreover, if a loss is completely asymmetric, then it is robust to any label noise model. In the next subsection, we will provide a comprehensive analysis.

Inspired by the benefit of symmetric \cite{sce} or complementary learning\cite{kim2019nlnl}, the Active Passive Loss (APL) framework was proposed \cite{ma2020normalized} for both robust and sufficient learning. The following theorem indicates that the asymmetric loss functions are also suitable for the APL framework. In our experiments, we also employ the framework to achieve better or at least comparable performance.
\begin{theorem}
\label{linearity}
$\forall \alpha,\ \beta>0$, if $L_1$ and $L_2$ are asymmetric, then $\alpha L_1+\beta L_2$ is asymmetric.
\end{theorem}

We know that all symmetric loss functions are also asymmetric according to Theorem \ref{sy-is-asy}. Is there a new asymmetric loss function? However, Definition \ref{asymmetric-loss-function} is too abstract to find a new specific form. In the following, we will provide a comprehensive theoretical analysis about designing asymmetric loss functions and propose several specific ones.

\begin{figure*}[!t]
    \centering
    \subfigure[AGCE ($q=0.5$)]{
        \label{fig:AGCE-CIFAR10-a}
        \includegraphics[width=1.2in]{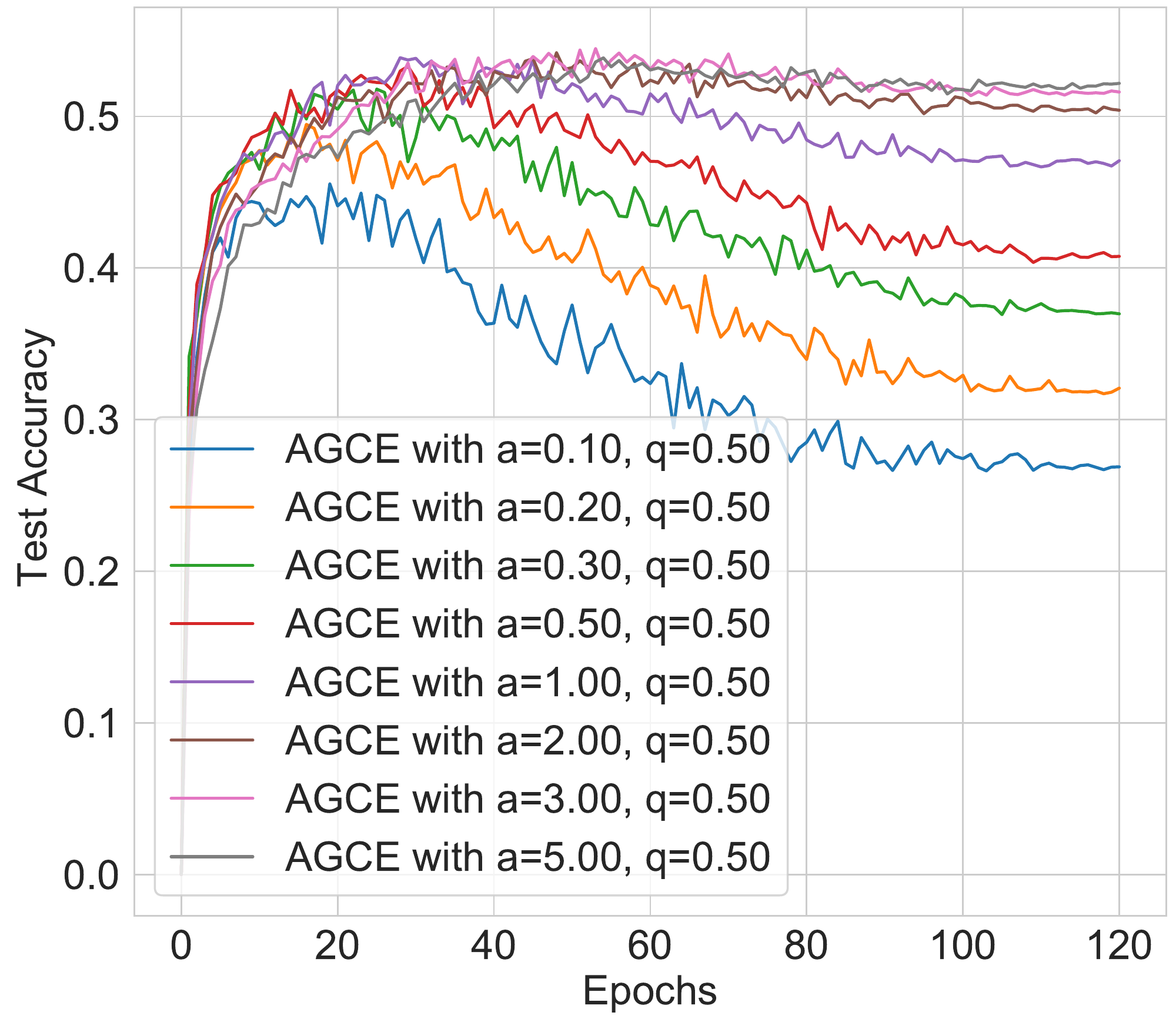}
    }
    \subfigure[AGCE ($q\ge 1$)]{
        \label{fig:AGCE-CIFAR10-b}
        \includegraphics[width=1.2in]{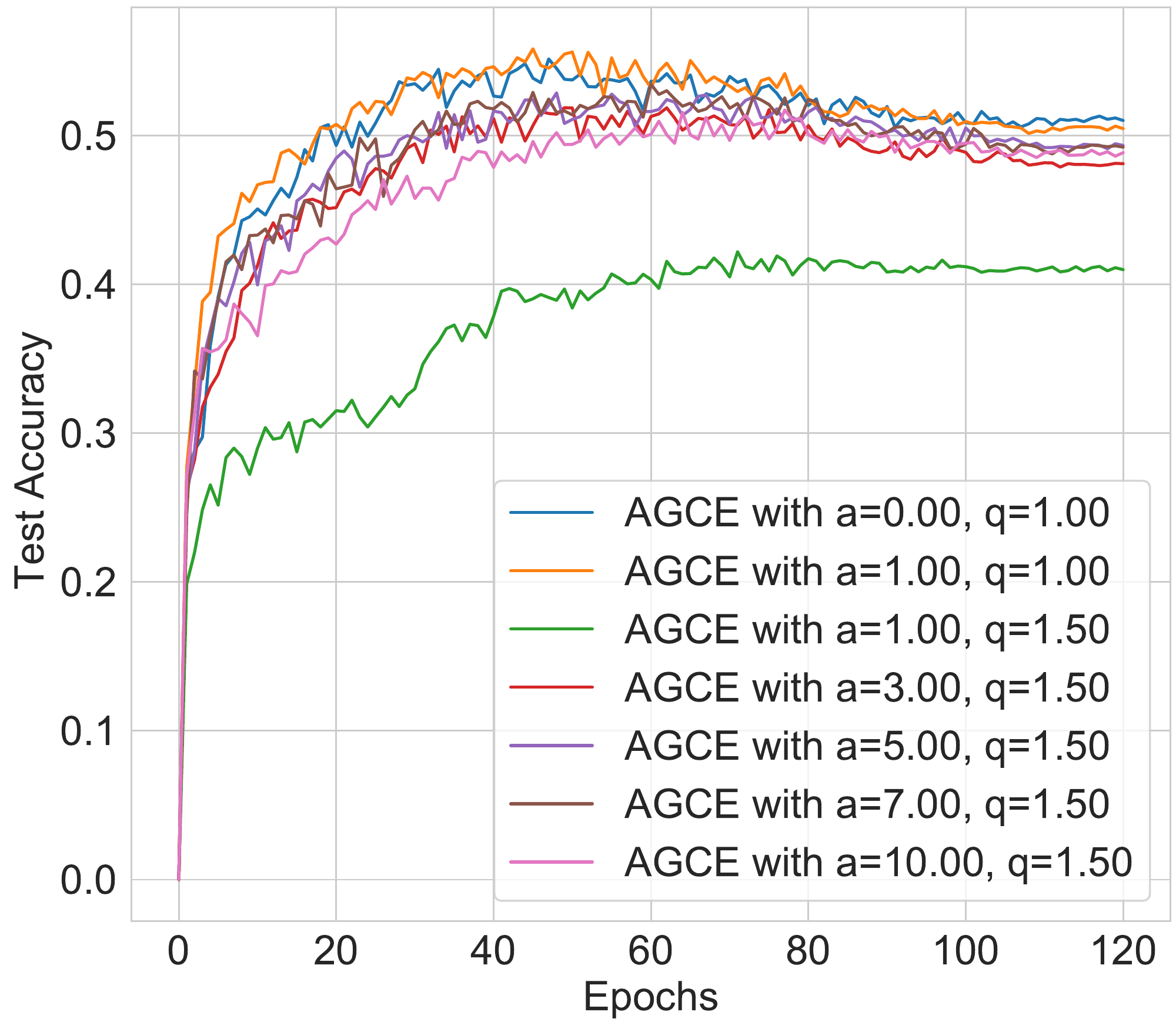}
    }
    \subfigure[AUL ($p<1$)]{
        \label{fig:AUL-CIFAR10-a}
        \includegraphics[width=1.2in]{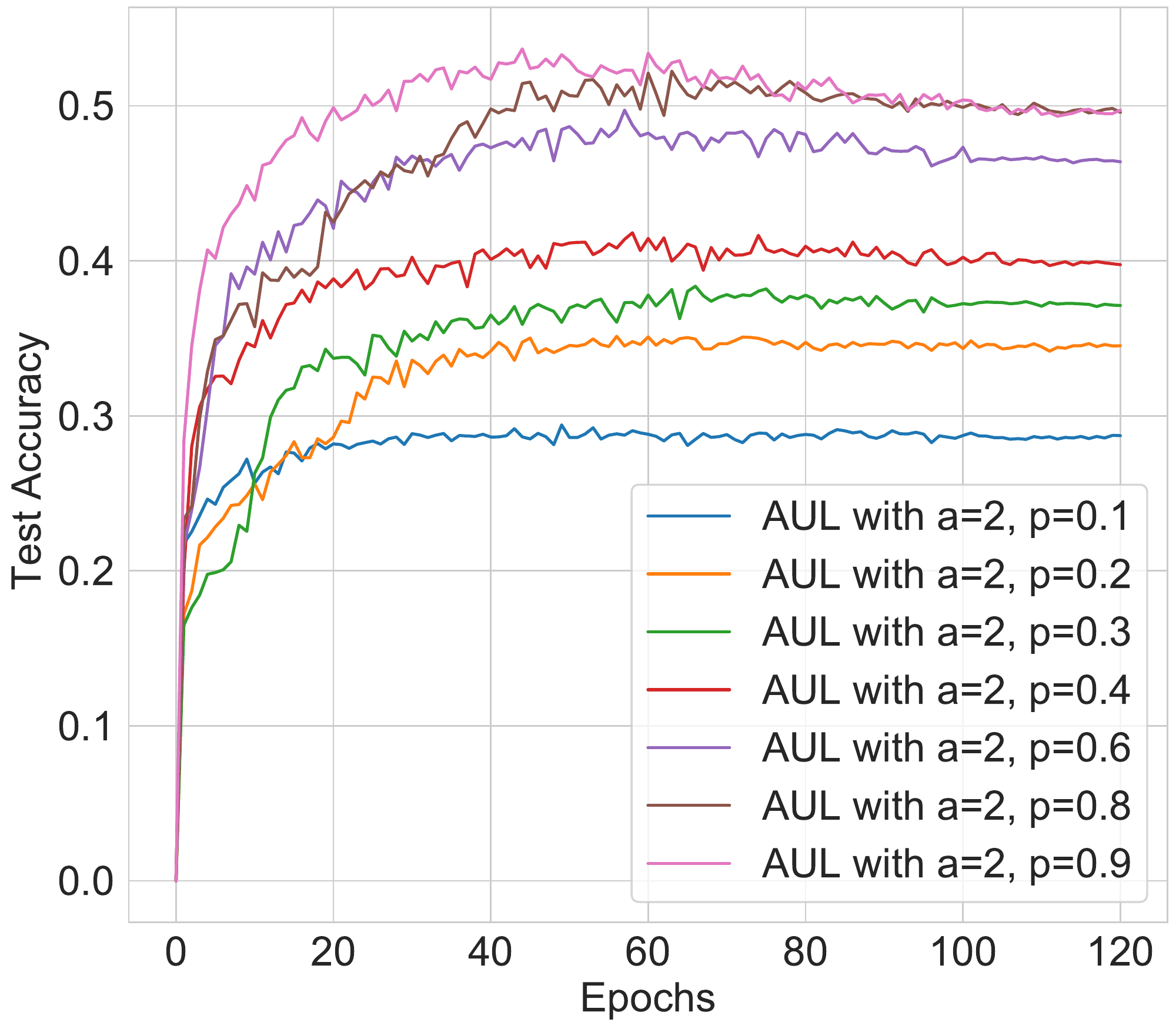}
    }
    \subfigure[AUL ($p=2$)]{
        \label{fig:AUL-CIFAR10-b}
        \includegraphics[width=1.2in]{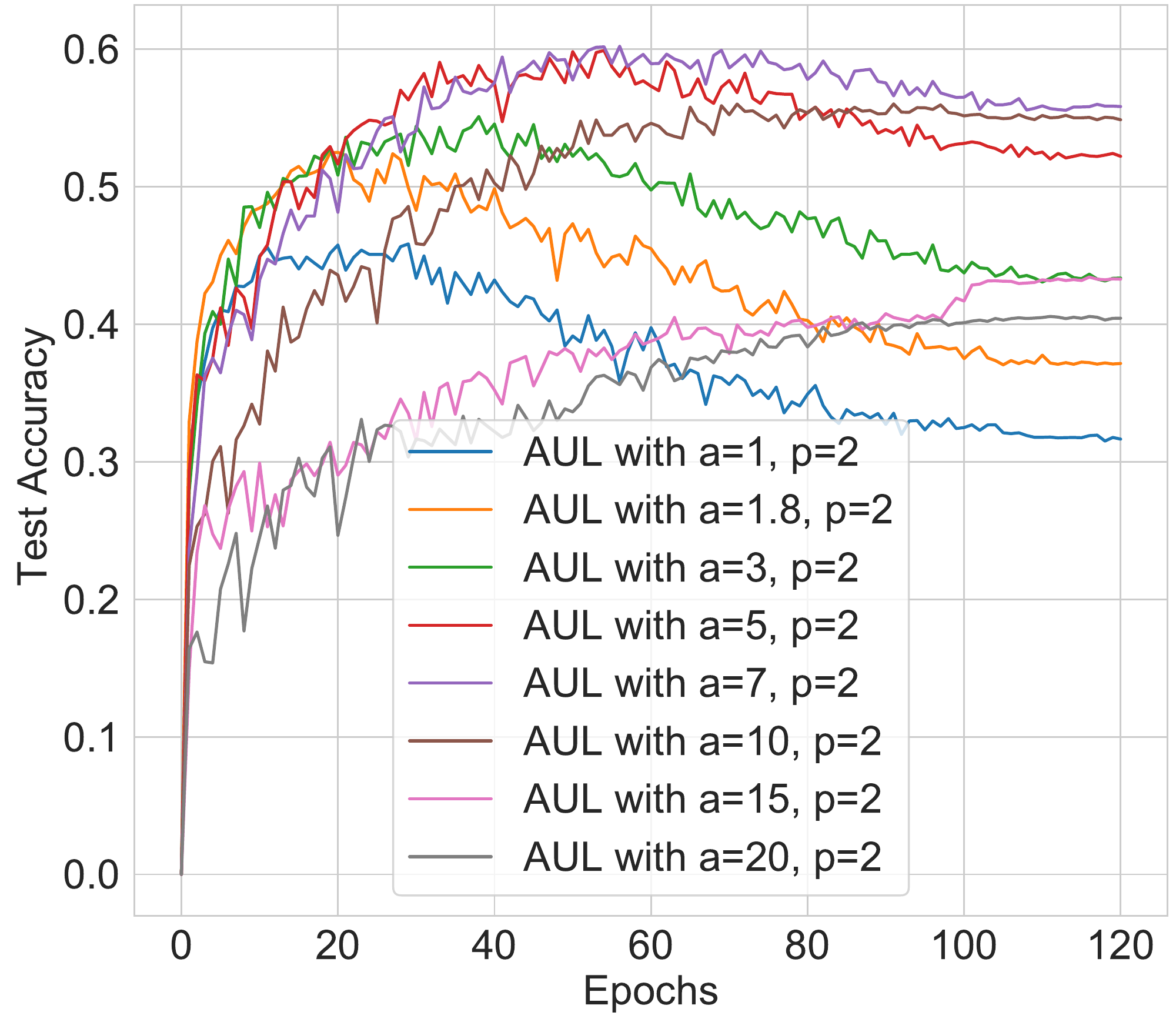}
    }
    \subfigure[AEL ($p=2$)]{
        \label{fig:AEL-CIFAR10}
        \includegraphics[width=1.2in]{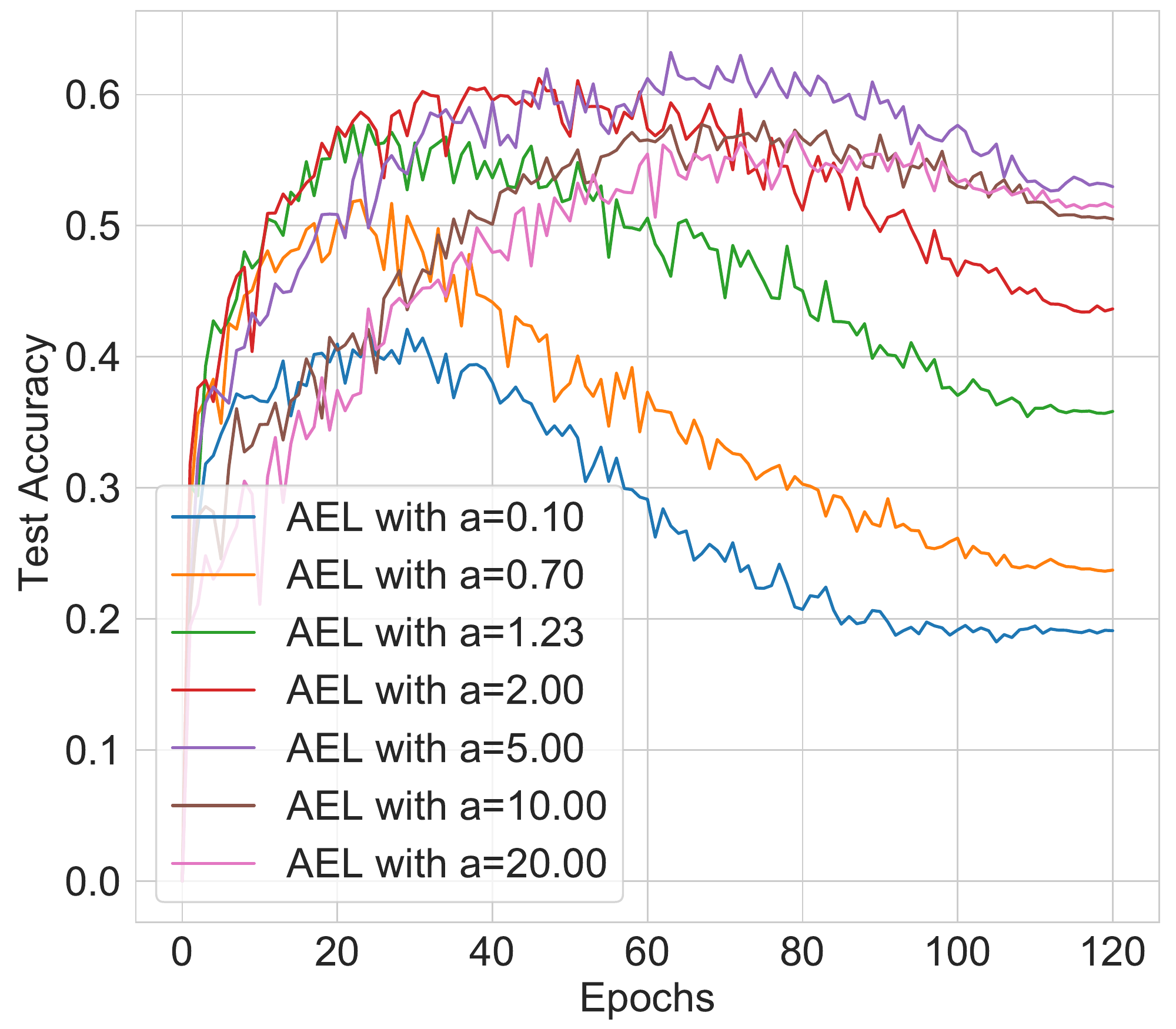}
    }
    \vskip-10pt
    \caption{Test accuracies of AGCE, AUL and AEL with different parameters on CIFAR-10 under 0.8 symmetric noise.}
    \vskip-10pt
\end{figure*}

\subsection{Asymmetry Ratio}
As we can see, the asymmetry or direction of minimization is dependent on which one is the maximum weight, but how to measure the asymmetry of the function and select an asymmetric enough loss? We give the following definition:

\begin{definition}
Consider a loss function $L(\mathbf u,i)=\ell(u_i)$, we define the asymmetry ratio $r(\ell)$ as
\vskip-10pt
\begin{equation}
r(\ell)= \inf_{\substack{0\le u_1,u_2\le 1\\ u_1+u_2\le 1\\ 0\le \Delta u\le u_2}} \frac{\ell(u_1)-\ell(u_1+\Delta u)}{\ell(u_2-\Delta u)-\ell(u_2)}.
\end{equation}
\end{definition}
\vskip-5pt
The asymmetry ratio $r$ denotes the infimum ratio of change in the loss function $\ell$ when we increase the value of $u_1$ to $u_1+\Delta u$, and correspondingly decrease the value of $u_2$ to $u_2-\Delta u$. For example, the asymmetry ratio of MAE is 1, and the asymmetric ratio of GCE is 0 ($q<1$). Based on the definition, we obtain the sufficient condition that $L$ is asymmetric on some weights.
\begin{theorem}[Sufficiency]
\label{suf}
On the given weights $w_1,..,w_k$, where $w_m > w_n$ and $w_n=\max_{i\not = m}w_i$, the loss function $L(\mathbf u,i)=\ell(u_i)$ is asymmetric if $\frac{w_m}{w_n}\cdot r(\ell)\ge 1$. 
\end{theorem}

\textbf{Remark.} Let $c=\min_{(\mathbf x,y), i\not=y}\frac{1-\eta_{\mathbf x}}{\eta_{\mathbf x,i }}$, which can be regarded as a measure of the clean level for the label noise mode in (\ref{noisemodel}). The larger the $c$, the cleaner the labels. 
Moreover, we usually have $c\ge 1$ in accordance with Assumption \ref{dominate}, then $\ell$ will be completely asymmetric if $r(\ell)\ge 1$. On the other hand, we can estimate $c$ to design a loss that satisfies $r(\ell)\ge \frac{1}{c}$, \textit{i.e.}, being asymmetric on the label noise model, which leads to noise tolerance in accordance with Theorem \ref{robust}. In fact, a loss satisfying $r(\ell)\ge \frac{1}{1.5}$ can be verified to be asymmetric to handle all synthetic noises regardless of on MNIST or CIFAR-10/-100. For a real-world dataset, the clean level is usually higher than the synthetic case.
In a sense, Theorem \ref{suf} associates the clean level and the asymmetry ratio with noise tolerance. In Section \ref{exp}, we empirically show that larger $c\cdot r(\ell)$ would provide more noise tolerance.
In the following, we shows that if $c\cdot r(\ell) \ge 1$, asymmetric losses will produce at least a positive weighted optimization rather than negative effects for any hypothesis class.

\begin{theorem}
In a binary classification problem, we assume that $L$ is strictly asymmetric on the label noise model which is clean-labels-dominant, for any hypothesis class $\mathcal H$, let $f^*=\arg\min_{f\in\mathcal H}R_L^{\eta}(f)$. If $\forall \mathbf x$, we have $\frac{1-\eta_{\mathbf x}}{\eta_{\mathbf x}}\cdot r(L)>1$, then $f^*$ also minimizes a positive weighted $L$-risk $R_{w, L}(f)=\mathbb E[w(\mathbf x, y)\cdot L(f(\mathbf x), y)]$.
\end{theorem}

According to the definition of the asymmetric ratio, we can easily obtain an upper bound of $r(\ell)$ when modifying the half-space constraint $u_1+u_2\le 1$ to the hyperplane $u_1+u_2=1$ and setting $\Delta u=u_2$, i.e.,
\vskip-15pt
\begin{equation}
    r(\ell)\le\inf_{\substack{0\le u_1,u_2\le 1\\ u_1+u_2= 1}} \frac{\ell(u_1)-\ell(1)}{\ell(0)-\ell(u_2)}=r_u(\ell).
\end{equation}
\vskip-8pt
In some cases, the equality will hold, for example, both $r$ and $r_u$ of MAE are equal to 1. Actually, a completely asymmetric loss $\ell$ satisfies $r_u(\ell)\ge 1$.

\begin{theorem}[Necessity]
\label{nec}
On the given weights $w_1,..,w_k$, where $w_m > w_n$ and $w_n=\max_{i\not = m}w_i$, the loss function $L(\mathbf u,i)=\ell(u_i)$ is asymmetric only if $\frac{w_m}{w_n}\cdot r_{u}(\ell)\ge 1$. 
\end{theorem}

According to Theorem \ref{suf} and Theorem \ref{nec}, when $r(\ell)=r_u(\ell)$, $\frac{w_m}{w_n}\cdot r(\ell)\ge 1$ will become the necessary and sufficient condition for $L(\mathbf u, i)=\ell(u_i)$ to be asymmetric. The following corollaries are straightforward from this.
\begin{corollary}
\label{AGCE}
On the given weights $w_1,..,w_k$, where $w_m > w_n$ and $w_n=\max_{i\not = m}w_i$, the loss function $L_q(\mathbf u,i)=[(a+1)^q-(a + u_i)^q]/q$ (where $q>0$, $a> 0$) is asymmetric $\Leftrightarrow$ $\frac{w_m}{w_n}\ge (\frac{a+1}{a})^{1-q}\cdot\mathbb{I}(q\le 1)+\mathbb{I}(q>1)$.
\end{corollary}

\begin{figure}[ht]
    \centering
    \subfigure[AGCEs]{
    \label{fig:AGCE}
    \includegraphics[width=1.4in]{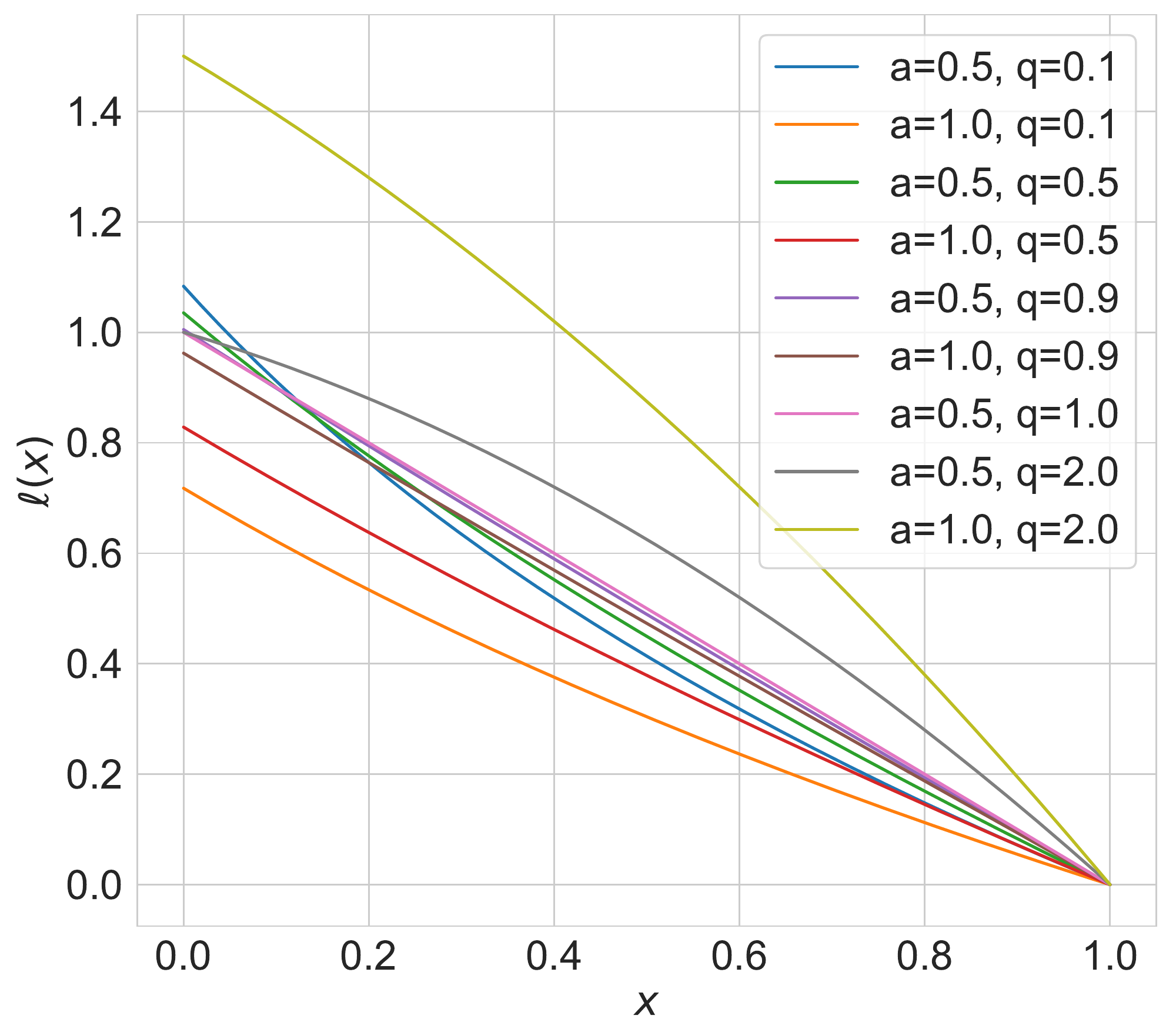}
    }
    \subfigure[AULs]{
    \label{fig:AUL}
    \includegraphics[width=1.4in]{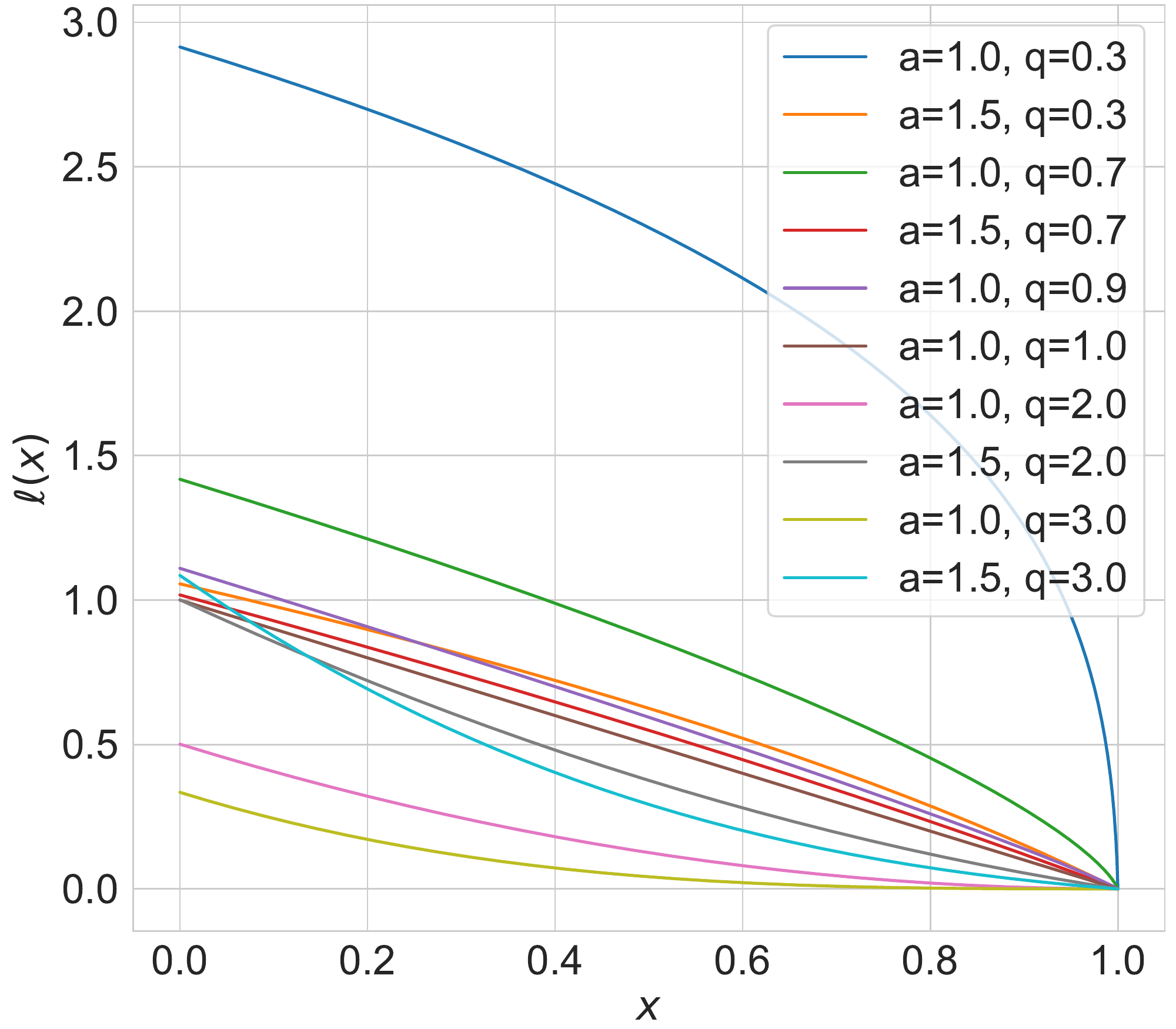}
    }
    \vskip-10pt
    \caption{Illustration of asymmetric loss functions.}
    \label{fig:ALFs}
    \vskip-5pt
\end{figure}

Mathematically, the loss function $L_q$, shown in Figure \ref{fig:AGCE}, is the negative shifted Box-Cox transformation, which we name as the Asymmetric Generalized Cross Entropy (AGCE) because when $0<q\le 1$ and $a=0$, the loss function is called GCE \cite{GCE} which can be seen as a generalized mixture of CCE (when $q\rightarrow 0$) and MAE (when $q=1$). Like MAE, both the $r$ and $r_u$ of AGCE are equal. More specifically, $r$ is equal to $(\frac{a}{a+1})^{1-q}$ when $q\le 1$, and 1 when $q\ge1$. As a consequence,  AGCE is completely asymmetric when $q\ge 1$. Corollary \ref{AGCE} shows that if $q>1$, the loss function beyond the range of $q$ in GCE is asymmetric, or if $q\le 1$ and $\frac{w_m}{w_n}\ge (\frac{a+1}{a})^{1-q}$, the convex loss function is also asymmetric, but when $q<1$ and $a=0$, the conventional GCE is not asymmetric.

\begin{corollary}
\label{AUL}
On the given weights $w_1,..,w_k$, where $w_m > w_n$ and $w_n=\max_{i\not = m}w_i$, the loss function $L_p(\mathbf u,i)=[(a-u_i)^p-(a-1)^p]/p$ (where $p>0$ and $a> 1$) is asymmetric $\Leftrightarrow$ $\frac{w_m}{w_n}\ge (\frac{a}{a-1})^{p-1}\cdot\mathbb{I}(p> 1)+\mathbb{I}(p\le1)$.
\end{corollary}

We call the loss function above the Asymmetric Unhinged Loss (AUL) shown in Figure \ref{fig:AUL}
, because it is derived from the unhinged loss ($a=1$ and $p=1$). Both the $r$ and $r_u$ of AUL are also equal, more specifically, the value of $r$ is $(\frac{a-1}{a})^{p-1}$ when $p\ge 1$, and $1$ when $p<1$. Similar to AGCE, AUL is completely asymmetric when $p\le 1$.
\vskip-5pt

\begin{corollary}
\label{AEL}
On the given weights $w_1,..,w_k$, where $w_m > w_n$ and $w_n=\max_{i\not = m}w_i$, the exponential loss function $L_{a}(\mathbf u, i)=\exp(-u_i/a)$ (where $a>0$) is asymmetric $\Leftrightarrow$ $\frac{w_m}{w_n}\ge  \exp(1/a)$.
\end{corollary}

We call the convex loss function above the Asymmetric Exponential Loss (AEL). According to the Corollary \ref{AEL}, we know that both $r$ and $r_u$ of AEL are equal to $\exp(-1/a)\le 1$, so AELs will not be completely asymmetric.

\section{Experiments}
\label{exp}
In this section, we empirically investigate asymmetric loss functions on benchmark datasets, including MNIST \cite{MNIST}, CIFAR-10/-100 \cite{CIFAR}, and a real-world noisy dataset WebVision \cite{li2017webvision}.

\begin{table*}[!t]
\small
\centering
\caption{Test accuracies (\%) of different methods on benchmark datasets with clean or symmetric label noise ($\eta\in[0.2, 0.4, 0.6, 0.8]$). The results (mean$\pm$std) are reported over 3 random runs and the top 3 best results are \textbf{boldfaced}.}
\label{symmetric-noise}
\begin{tabular}{c|c|c|cccc}
    \hline
     \multirow{2}*{Datasets} & \multirow{2}*{Methods} & \multirow{2}*{Clean ($\eta=0.0$)} & \multicolumn{4}{c}{Symmetric Noise Rate ($\eta$)}  \\
     ~ & ~ & ~ & 0.2 & 0.4 & 0.6 & 0.8\\
     \hline\hline
     \multirow{10}*{MNIST} & CE & 99.15 $\pm$ 0.05 & 91.62 $\pm$ 0.39 & 73.98 $\pm$ 0.27 & 49.36 $\pm$ 0.43 & 22.66 $\pm$ 0.61\\
    ~ & FL &99.13 $\pm$ 0.09 &91.68 $\pm$ 0.14 &74.54 $\pm$ 0.06 &50.39 $\pm$ 0.28 &22.65 $\pm$ 0.26 \\
    ~ & GCE &99.27 $\pm$ 0.05 &98.86 $\pm$ 0.07 &97.16 $\pm$ 0.03 &81.53 $\pm$ 0.58 &33.95 $\pm$ 0.82\\
    ~ & NLNL &98.61 $\pm$ 0.13 &98.02 $\pm$ 0.14 &97.17 $\pm$ 0.09 &95.42 $\pm$ 0.30 &86.34 $\pm$ 1.43\\
    ~ & SCE &99.23 $\pm$ 0.10 &98.92 $\pm$ 0.12 &97.38 $\pm$ 0.15 &88.83 $\pm$ 0.55 &48.75 $\pm$ 1.54\\
    ~ & NCE &98.60 $\pm$ 0.06 &98.57 $\pm$ 0.01 &98.29 $\pm$ 0.05 &97.65 $\pm$ 0.08 &93.78 $\pm$ 0.41\\
    ~ & NCE+RCE &{99.36 $\pm$ 0.05} &\textbf{99.14 $\pm$ 0.03} &98.51 $\pm$ 0.06 &95.60 $\pm$ 0.21 &74.00 $\pm$ 1.68\\
    \cline{2-7}
      ~ & \textbf{AUL} & {99.14 $\pm$ 0.05} &\textbf{99.05 $\pm$ 0.09} &\textbf{98.90 $\pm$ 0.09} &\textbf{98.67 $\pm$ 0.04} &\textbf{96.73 $\pm$ 0.20}\\
    ~ & \textbf{AGCE} &99.05 $\pm$ 0.11 &\textbf{98.96 $\pm$ 0.10} &\textbf{98.83 $\pm$ 0.06} &\textbf{98.57 $\pm$ 0.12} &\textbf{96.59 $\pm$ 0.12}\\
    ~ & \textbf{AEL} & 99.03 $\pm$ 0.05 &98.93 $\pm$ 0.06 &\textbf{98.78 $\pm$ 0.13} &\textbf{98.51 $\pm$ 0.06} &\textbf{96.40 $\pm$ 0.11}\\
     \hline\hline
     \multirow{10}*{CIFAR10} & CE & 90.48 $\pm$ 0.11 & 74.68 $\pm$ 0.25 & 58.26 $\pm$ 0.21 & 38.70 $\pm$ 0.53 & 19.55 $\pm$ 0.49\\
     ~ & FL & 89.82 $\pm$ 0.20 & 73.72 $\pm$ 0.08 & 57.90 $\pm$ 0.45 & 38.86 $\pm$ 0.07 & 19.13 $\pm$ 0.28\\
     ~ & GCE & 89.59 $\pm$ 0.26 & 87.03 $\pm$ 0.35 & 82.66 $\pm$ 0.17 & 67.70 $\pm$ 0.45 & 26.67 $\pm$ 0.59\\
     ~ & SCE & 91.61 $\pm$ 0.19 & 87.10 $\pm$ 0.25 & 79.67 $\pm$ 0.37 & 61.35 $\pm$ 0.56 & 28.66 $\pm$ 0.27\\
     ~ & NLNL & 90.73 $\pm$ 0.20 & 73.70 $\pm$ 0.05 & 63.90 $\pm$ 0.44 & 50.68 $\pm$ 0.47 & 29.53 $\pm$ 1.55\\
     ~ & NCE & 75.65 $\pm$ 0.26 & 72.89 $\pm$ 0.25 & 69.49 $\pm$ 0.39 & 62.64 $\pm$ 0.18 & 41.49 $\pm$ 0.66\\
     ~ & NCE+RCE &{90.87 $\pm$ 0.37} &\textbf{89.25 $\pm$ 0.42} &\textbf{85.81 $\pm$ 0.08} &\textbf{79.72 $\pm$ 0.20} &\textbf{55.74 $\pm$ 0.95}\\
     \cline{2-7}
     ~ & \textbf{AUL} & 91.27 $\pm$ 0.12 & 89.21 $\pm$ 0.09 & 85.64 $\pm$ 0.19 & 78.86 $\pm$ 0.66 & \textbf{52.92 $\pm$ 1.20}\\
     ~ & \textbf{AGCE} & 88.95 $\pm$ 0.22 & 86.98 $\pm$ 0.12 & 83.39 $\pm$ 0.17 & 76.49 $\pm$ 0.53 & 44.42 $\pm$ 0.74\\
     ~ & \textbf{AEL} & 86.38 $\pm$ 0.19 & 84.27 $\pm$ 0.12 & 81.12 $\pm$ 0.20 & 74.86 $\pm$ 0.22 & 51.41 $\pm$ 0.32\\
     ~ & \textbf{NCE+AUL} & 91.10 $\pm$ 0.13 & \textbf{89.31 $\pm$ 0.20} & \textbf{86.23 $\pm$ 0.18} & \textbf{79.70 $\pm$ 0.08} & \textbf{59.44 $\pm$ 1.14}\\
     ~ & \textbf{NCE+AGCE} & {90.94 $\pm$ 0.12} & \textbf{89.21 $\pm$ 0.08} & \textbf{86.19 $\pm$ 0.15} & \textbf{80.13 $\pm$ 0.18} & {50.82 $\pm$ 1.46}\\
     ~ & \textbf{NCE+AEL} & {90.71 $\pm$ 0.04} & {88.57 $\pm$ 0.14} & {85.01 $\pm$ 0.38} & {77.33 $\pm$ 0.18} & 47.90 $\pm$ 1.21\\
     \hline\hline
     \multirow{10}*{CIFAR100} & CE &71.33 $\pm$ 0.43	&56.51 $\pm$ 0.39	&39.92 $\pm$ 0.10	&21.39 $\pm$ 1.17	&\ \  7.59 $\pm$ 0.20\\
     ~ & FL &70.06 $\pm$ 0.70	&55.78 $\pm$ 1.55	&39.83 $\pm$ 0.43	&21.91 $\pm$ 0.89	&\ \ 7.51 $\pm$ 0.09\\
     ~ & GCE &63.09 $\pm$ 1.39	&61.57 $\pm$ 1.06	&56.11 $\pm$ 1.35	&45.28 $\pm$ 0.61	&17.42 $\pm$ 0.06\\
     ~ & SCE &69.62 $\pm$ 0.42	&52.25 $\pm$ 0.14	&36.00 $\pm$ 0.69	&20.14 $\pm$ 0.60	&\ \ 7.67 $\pm$ 0.63\\
     ~ & NLNL & 68.72 $\pm$ 0.60	&46.99 $\pm$ 0.91	& 30.29 $\pm$ 1.64	&16.60 $\pm$ 0.90	&11.01 $\pm$ 2.48\\
     ~ & NCE &29.96 $\pm$ 0.73	&25.27 $\pm$ 0.32	&19.54 $\pm$ 0.52	&13.51 $\pm$ 0.65	&\ \ 8.55 $\pm$ 0.37\\
    ~ & NCE+RCE &68.65 $\pm$ 0.40	&64.97 $\pm$ 0.49	&58.54 $\pm$ 0.13	&45.80 $\pm$ 1.02	&\textbf{25.41 $\pm$ 0.98}\\
     \cline{2-7}
     ~ & \textbf{NCE+AUL} & 68.96 $\pm$ 0.16 & \textbf{65.36 $\pm$ 0.20} & \textbf{59.25 $\pm$ 0.23} & \textbf{46.34 $\pm$ 0.21} & 23.03 $\pm$ 0.64\\
     ~ & \textbf{NCE+AGCE} & 69.03 $\pm$ 0.37 & \textbf{65.66 $\pm$ 0.46} & \textbf{59.47 $\pm$ 0.36} & \textbf{48.02 $\pm$ 0.58} & \textbf{24.72 $\pm$ 0.60}\\
     ~ & \textbf{NCE+AEL} & 68.70 $\pm$ 0.20 & \textbf{65.36 $\pm$ 0.14} & \textbf{59.51 $\pm$ 0.03} & \textbf{46.94 $\pm$ 0.07} & \textbf{24.48 $\pm$ 0.24}\\
     \hline
\end{tabular}
\vskip-12pt
\end{table*}

\subsection{The Robustness of Asymmetric Loss Functions}
\textbf{Validation of Classification Calibration.} We first conduct an experiment to validate the classification calibration in Theorem \ref{classification-calibration}. As a corroboration, we plot the curves of $H_{\ell}(\eta)$ and $H_{\ell}^-(\eta)$ (the definitions can be found in the supplementary material) for the proposed losses AGCE with $q>1$ and AUL with $p<1$, which are completely asymmetric according to Corollaries \ref{AGCE} and \ref{AUL}. As shown in Fig. \ref{fig:classificatoin-calibration}, under the same parameter setting, the curve of $H_{\ell}^-(\eta)$ (dashed ones) is always above the corresponding of $H_{\ell}(\eta)$ (solid ones) when $\eta\neq 1/2$, \textit{i.e.}, the loss functions are classification-calibrated.

\textbf{Validation of Corollaries.} We also design a simple experiment to validate the necessary and sufficient conditions in Corollaries \ref{AGCE}, \ref{AUL} and \ref{AEL}, where we randomly generate a positive weight vector $\mathbf w\in\mathbb R_+^k$ ($k$ is set to $10$), and initialize a random variable $\mathbf z\in\mathbb{R}^k$. Our goal is to optimize $\mathbf z$ by minimizing  $\sum_{i=1}^k w_i L(\sigma(\mathbf z),i)$, where $\sigma(\cdot)$ denotes the softmax function. As aforementioned, asymmetric loss functions will optimize $\mathbf p=\sigma(\mathbf z)$ as a one-hot vector. The experimental results are shown in Figure \ref{fig:NC}. Let $m=\arg\max_i w_i$, we can see that when $\frac{w_m}{w_n}\cdot r(\ell)\ge 1$, $p_m$ is very close to $1$, and has an obvious gap from 1 when $\frac{w_m}{w_n}\cdot r(\ell)< 1$. The consequence holds regardless of AGCE, AUL, or AEL, and an important phenomenon is that the curve is more and more asymmetric as $a$ or $r(\ell)$ gets bigger and bigger.

\begin{figure*}[htb]
    \centering
    \subfigure[GCE with $\eta=0.0$]{
    \label{fig:GCE0.0}
    \includegraphics[width=1.2in]{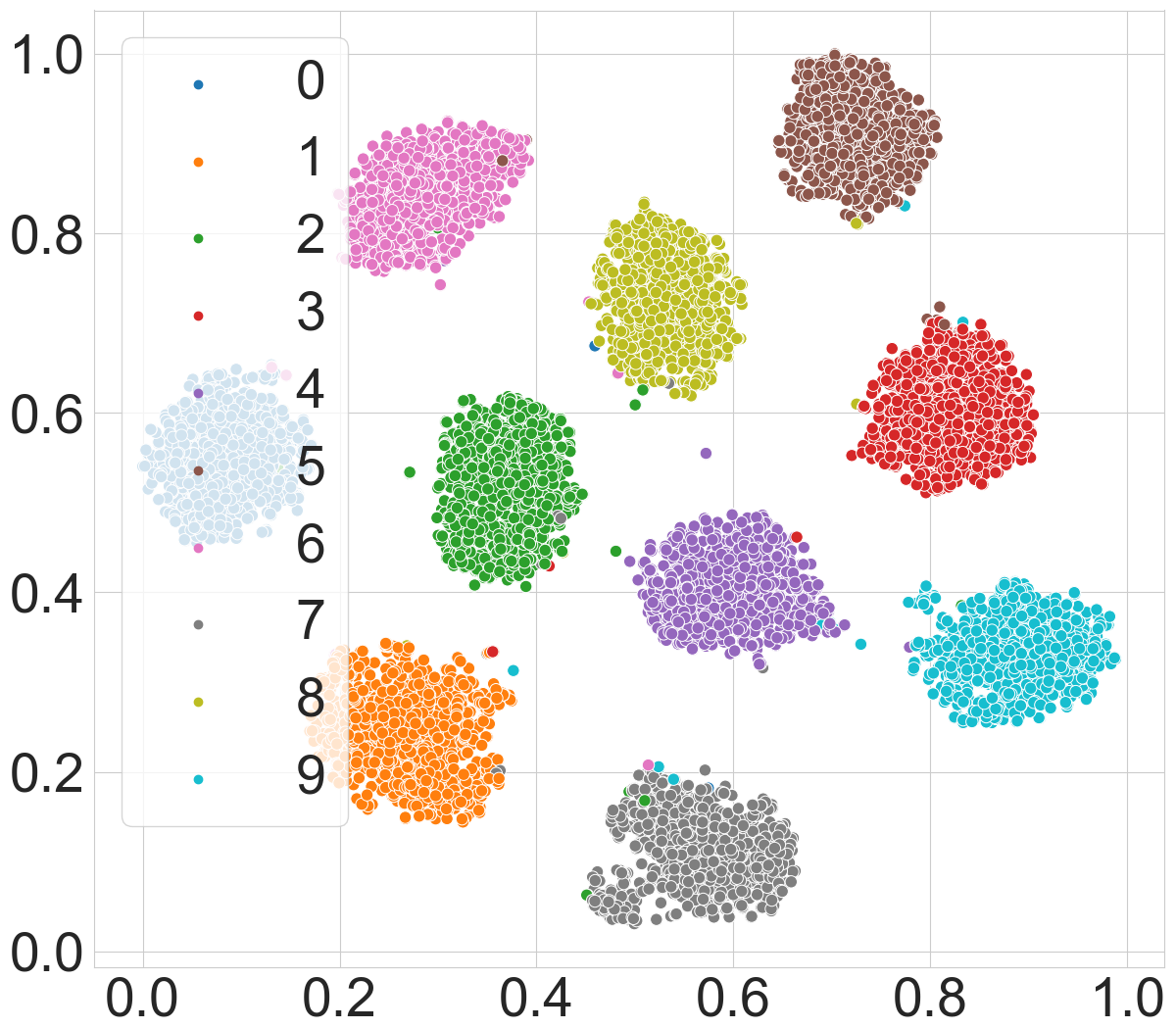}
    }
    \subfigure[GCE with $\eta=0.2$]{
    \label{fig:GCE0.2}
    \includegraphics[width=1.2in]{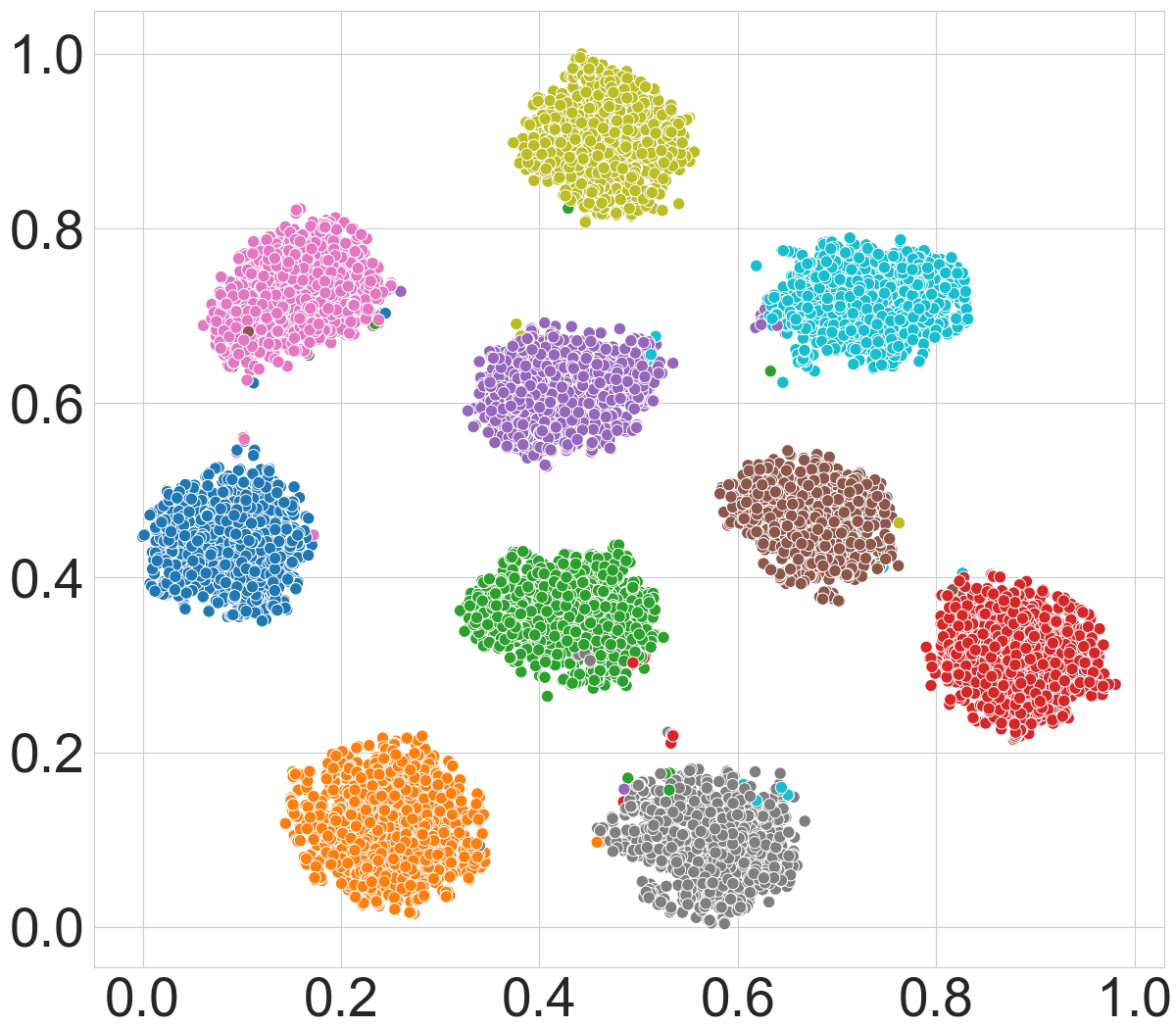}
    }
    \subfigure[GCE with $\eta=0.4$]{
    \label{fig:GCE0.4}
    \includegraphics[width=1.2in]{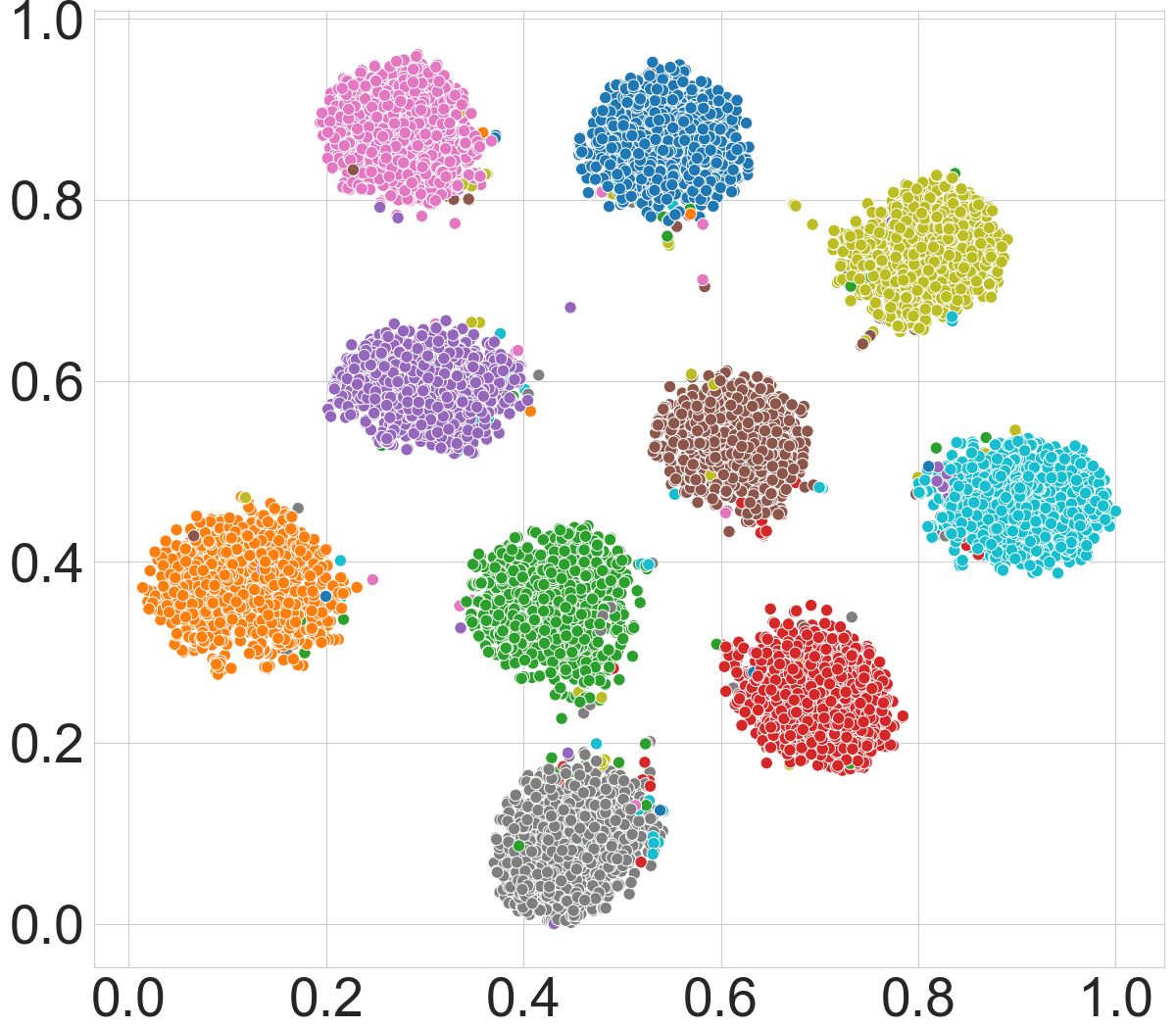}
    }
    \subfigure[GCE with $\eta=0.6$]{
    \label{fig:GCE0.6}
    \includegraphics[width=1.2in]{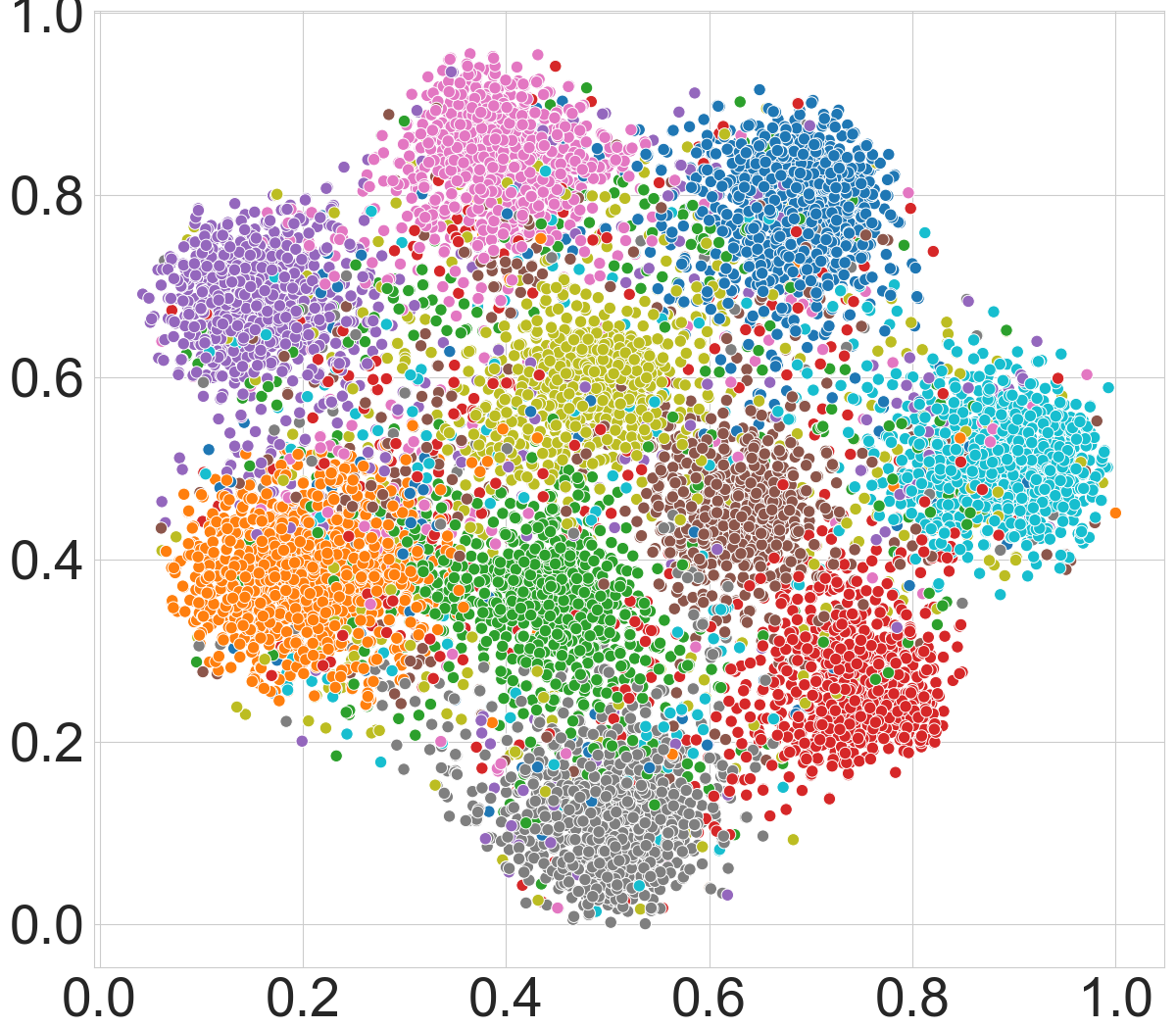}
    }
    \subfigure[GCE with $\eta=0.8$]{
    \label{fig:GCE0.8}
    \includegraphics[width=1.2in]{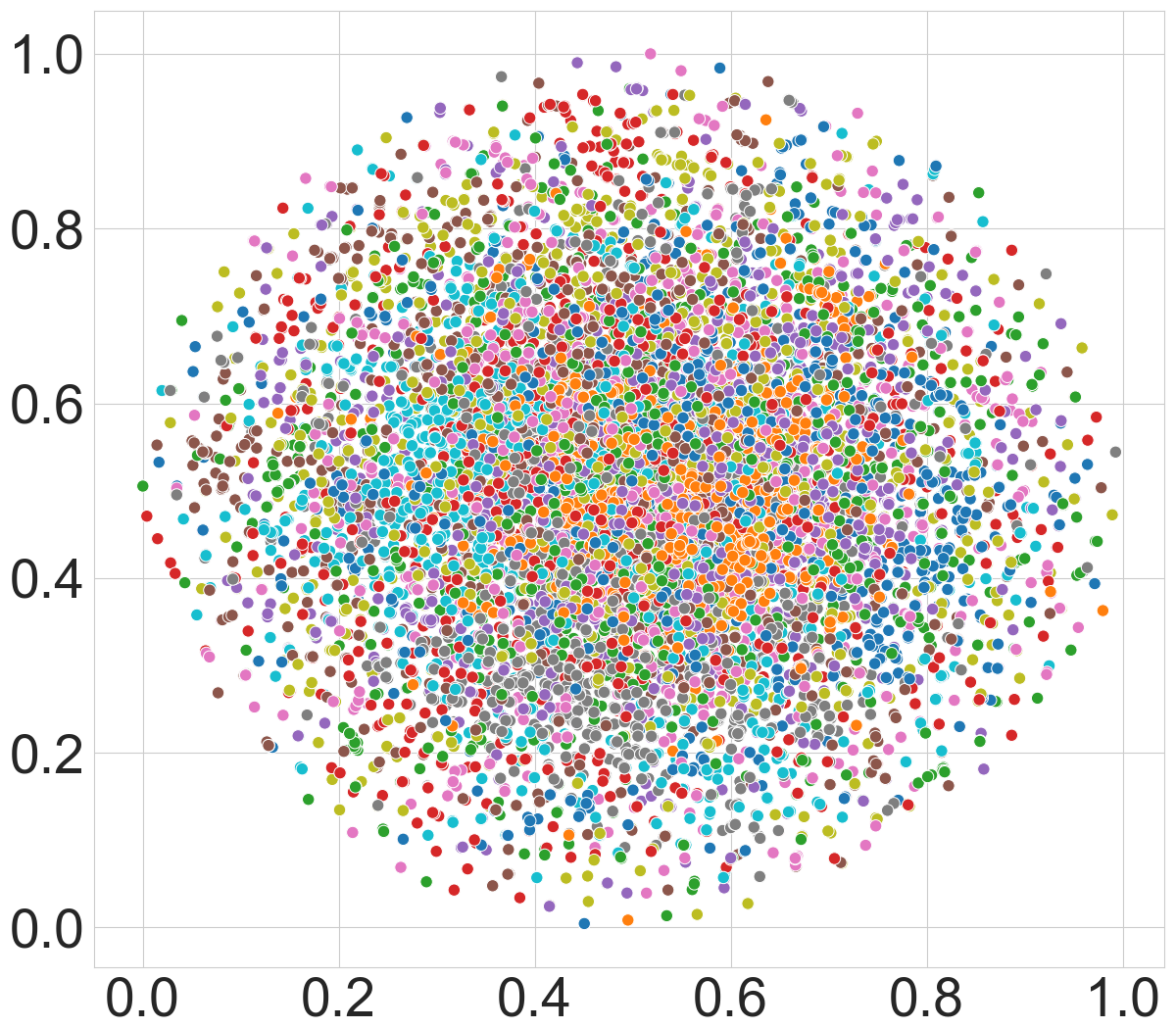}
    }
    \subfigure[AGCE with $\eta=0.0$]{
    \label{fig:AGCE0.0}
    \includegraphics[width=1.2in]{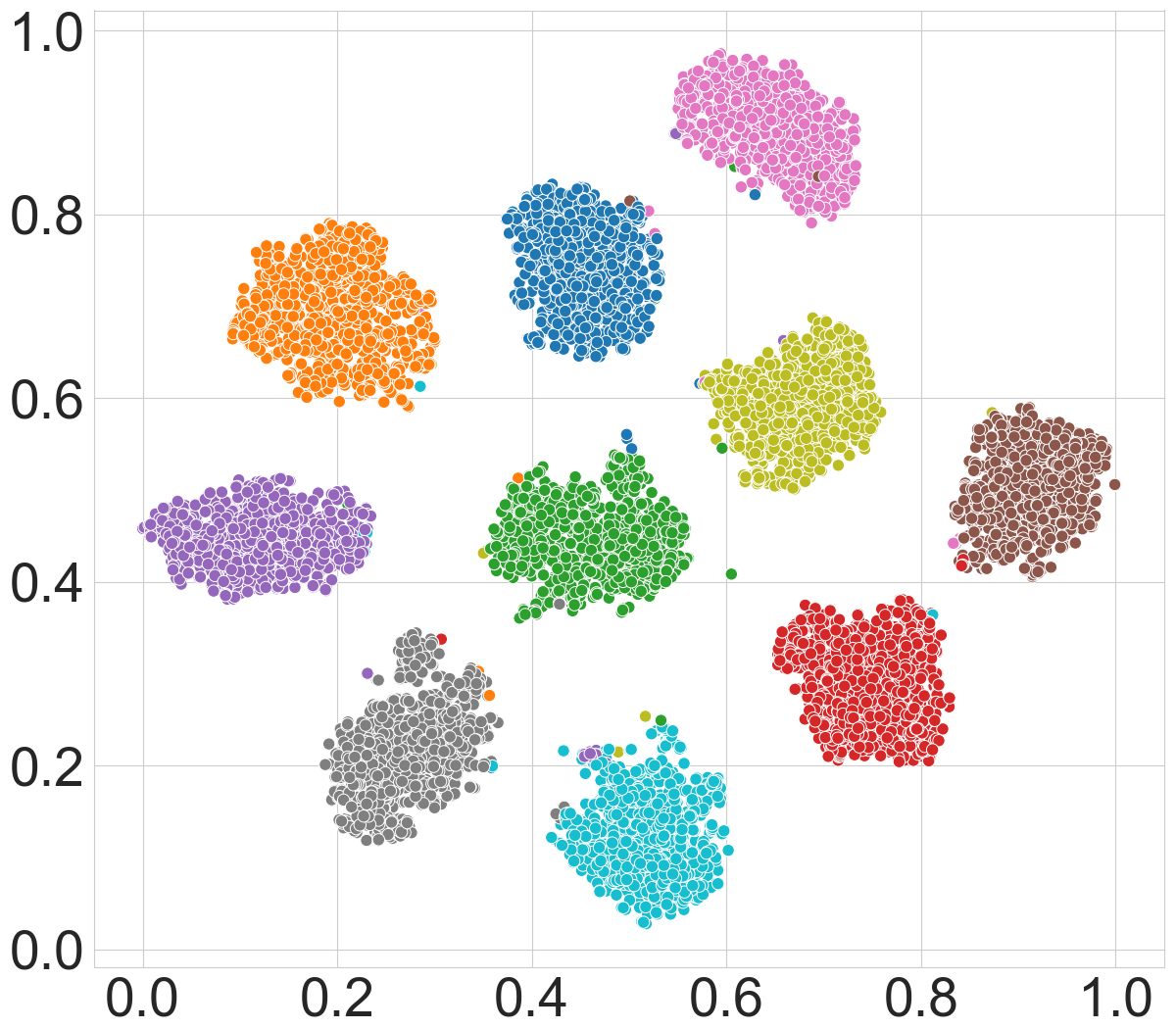}
    }
    \subfigure[AGCE with $\eta=0.2$]{
    \label{fig:AGCE0.2}
    \includegraphics[width=1.2in]{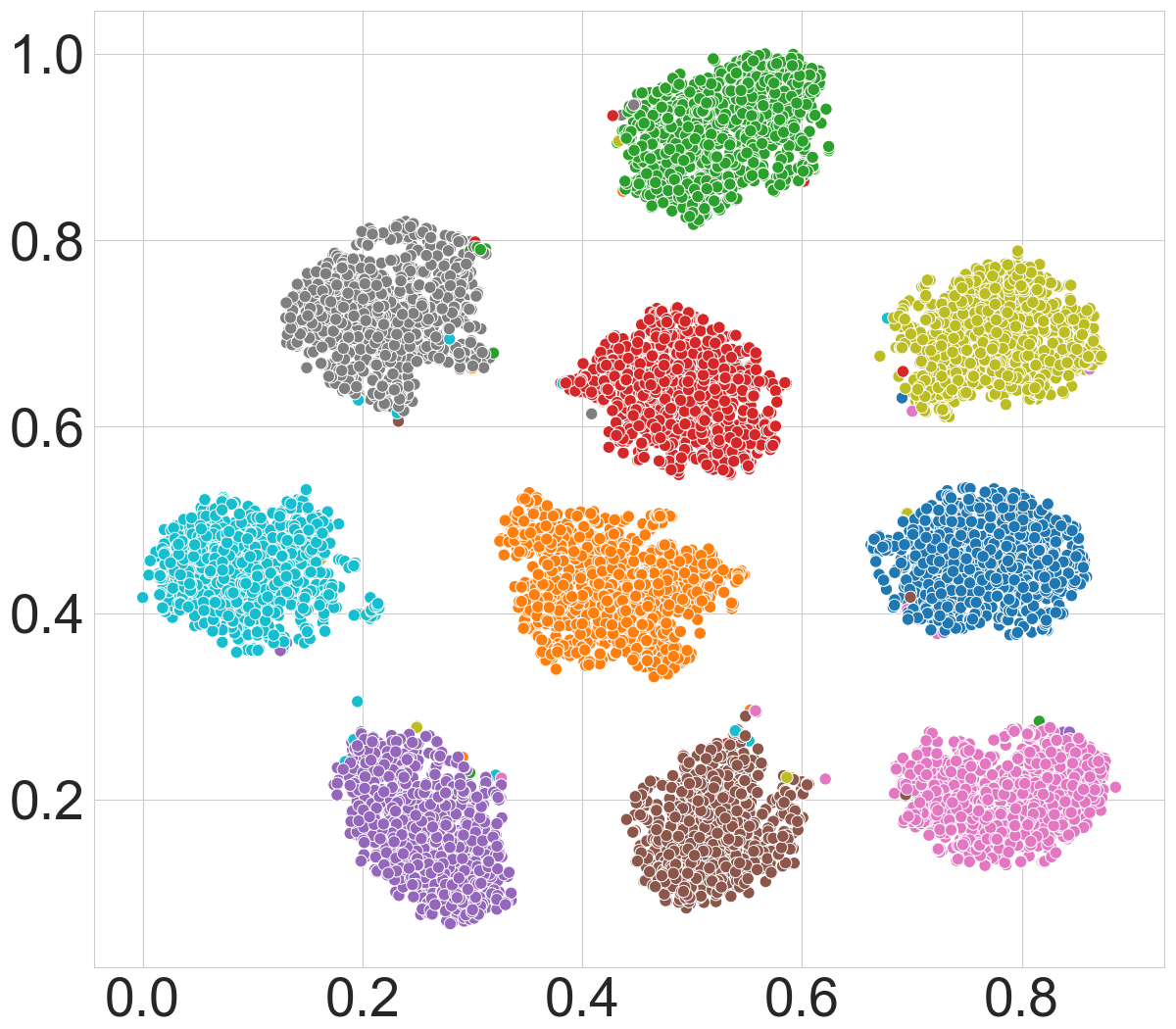}
    }
    \subfigure[AGCE with $\eta=0.4$]{
    \label{fig:AGCE0.4}
    \includegraphics[width=1.2in]{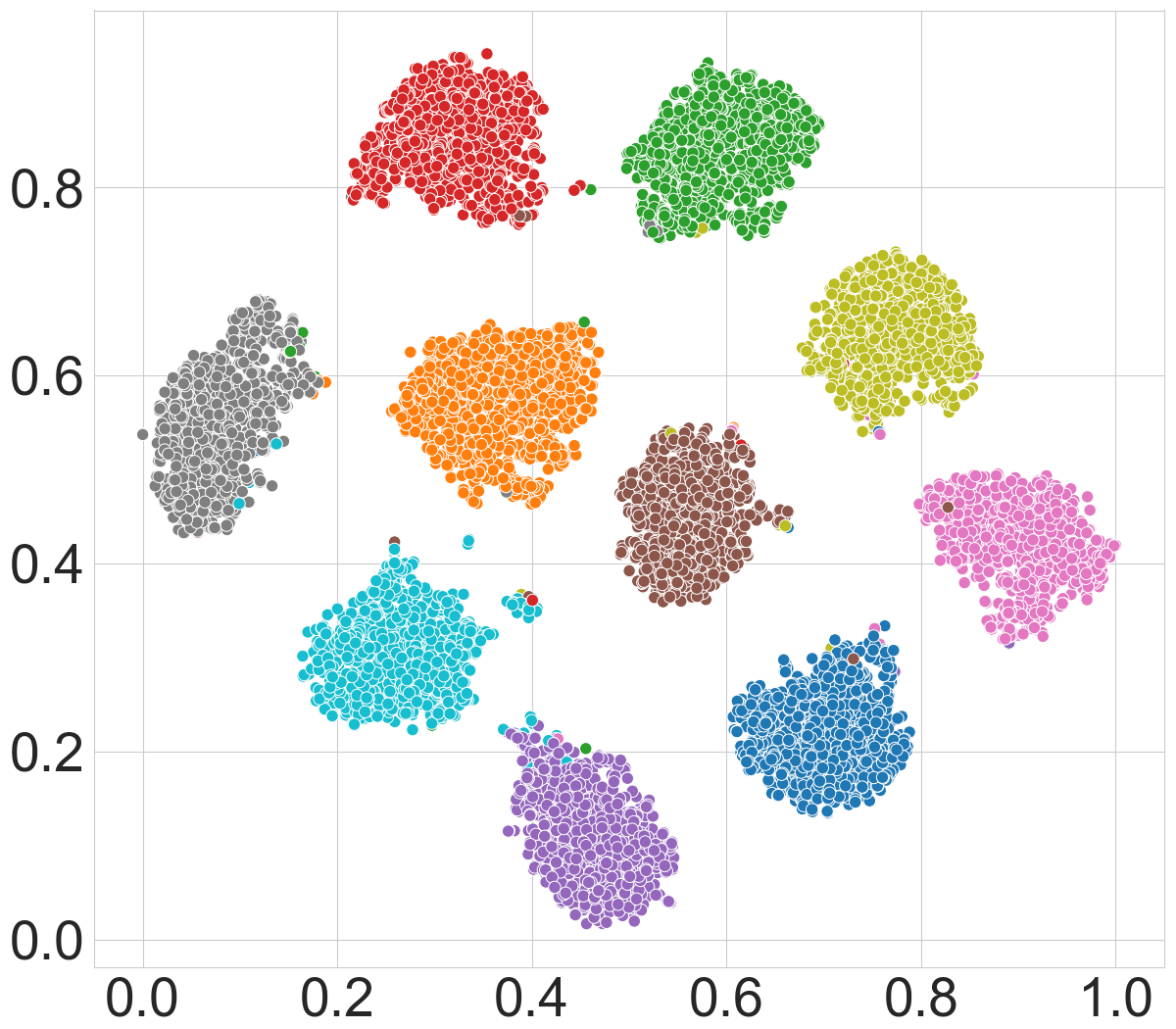}
    }
    \subfigure[AGCE with $\eta=0.6$]{
    \label{fig:AGCE0.6}
    \includegraphics[width=1.2in]{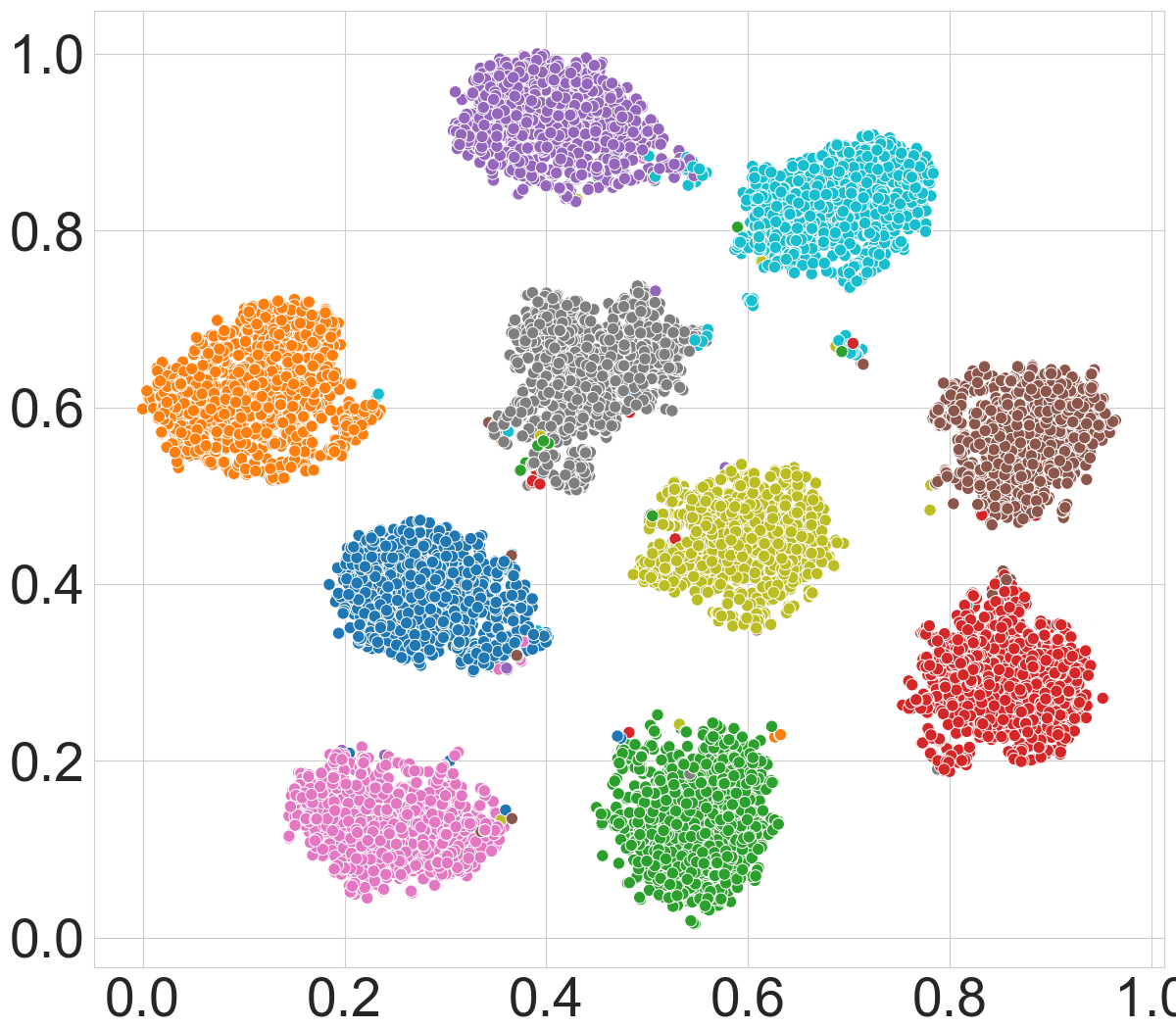}
    }
    \subfigure[AGCE with $\eta=0.8$]{
    \label{fig:AGCE0.8}
    \includegraphics[width=1.2in]{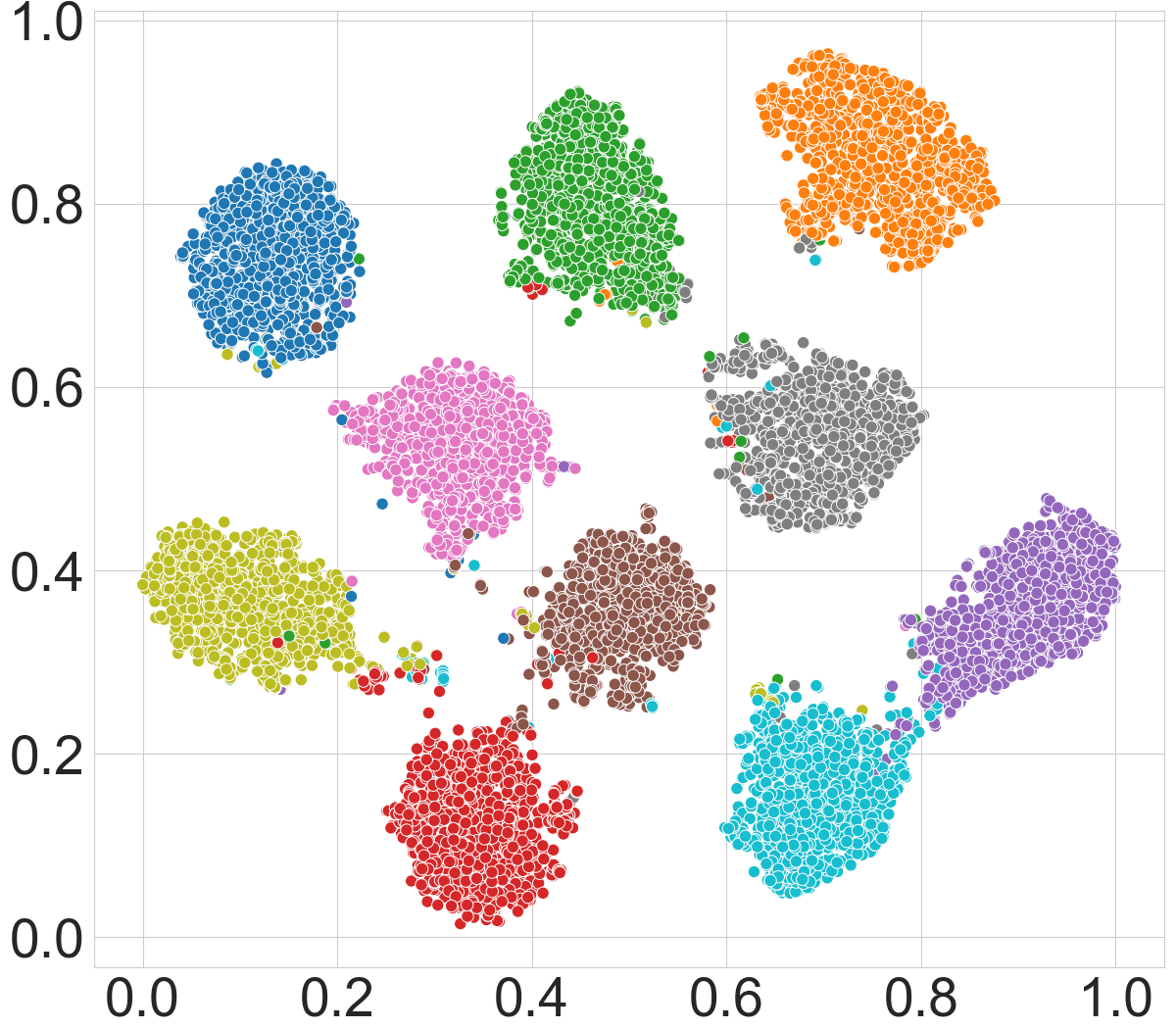}
    }
    \vskip-10pt
    \caption{Visualization for GCE (top) and AGCE (bottom) on MNIST with different symmetric noise ($\eta\in[0.0, 0.2, 0.4, 0.6, 0.8]$) by t-SNE \cite{tsne} 2D embeddings of deep features.}
    \label{fig:tsne-gce-agce}
    \vskip-10pt
\end{figure*}
\textbf{About Hyper-parameters.} We then run a set of experiments on CIFAR-10 to verify the robustness of asymmetric loss functions AGCE, AUL, and AEL with different hyper-parameter settings. The label noise is set to be symmetric and the noise rate is set to $0.8$. We use an 8-layer CNN as the model to be learned.

One of the advantages of asymmetric losses is that we do not need to know the exact values of noise rates, especially for completely asymmetric losses. For the symmetric noise with rate 0.8, the clean level $c=\min_{(\mathbf x,y)\in\mathcal S, i\not=y}\frac{1-\eta_{\mathbf x}}{\eta_{\mathbf x,i }}=\frac{9}{4}$. To make the AGCE asymmetric with $q=0.5$, we need to guarantee $\frac{9}{4}\cdot (\frac{a}{a+1})^{1-0.5}\ge 1$, \textit{i.e.}, $a\ge \frac{16}{65}$. To make the AUL asymmetric with $p=2$, we need to guarantee $\frac{9}{4}\cdot (\frac{a-1}{a})^{2-1}\ge 1$, \textit{i.e.}, $a\ge \frac{9}{5}$. To make the AEL asymmetric, we need to guarantee $\frac{9}{4}\ge \exp(1/a)$, \textit{i.e.}, $a\ge 1/\ln \frac{9}{4}$.

As shown in Figures \ref{fig:AGCE-CIFAR10-a} and \ref{fig:AGCE-CIFAR10-b}, when $q=1.5$, all the curves remain robust, and when $q=0.5$, the AGCE whose asymmetry ratio is smaller than $\frac{16}{65}$ exhibits significant overfitting after epoch 20. In Figure \ref{fig:AGCE-CIFAR10-a},  although the curve is not robust on $a=0.3>\frac{16}{65}$, it will be more and more robust as $a$ increases gradually. Our understanding is that the data is not ideal enough such that the optimization is a trade-off between sample separability and the asymmetry of loss. Similar experimental phenomena have occurred on AUL and AEL. According to Figures \ref{fig:AUL-CIFAR10-a} and \ref{fig:AUL-CIFAR10-b}, when $p<1$, AUL always remains robust, and when $p=2$, AUL becomes more and more robust as the asymmetric ratio $r(\ell)=\frac{a-1}{a}$ gets larger, which is similar to AEL.

\textbf{Remark.} An important experimental conclusion is that as $a$ or the asymmetric ratio $r$ increases, whether it is for AGCE ($q<1$), AUL ($p>1$), or AEL,  $c\cdot r(L)$ is becoming larger, and the training process shows more robust results, but may lead to less fitting ability. Therefore, we roughly follow a principled approach for hyper-parameter tuning: for simple datasets, we prefer the hyperparameters with a higher asymmetry ratio to obtain robustness, while for complicated datasets we tend to use hyper-parameters with a lower asymmetry ratio to obtain better fitting ability.

\subsection{Evaluation on Benchmark Datasets}
\textbf{Baselines.} We consider several state-of-the-art methods: Generalized Cross Entropy (GCE) \cite{GCE}, Negative Learning for Noisy Labels (NLNL) \cite{kim2019nlnl}, Symmetric Cross Entropy (SCE) \cite{sce}, Normalized Cross Entropy (NCE), the weighting of NCE and Reverse Cross Entropy (RCE), as well as our proposed AUL, AGCE and AEL. Inspired by the Active Passive Loss \cite{ma2020normalized}, we combine the proposed AGCE, AUL, and AEL with NCE, then we obtain NCE+ALFs, \textit{i.e.}, NCE+AGCE, NCE+AUL and NCE+AEL. We also train networks using the commonly-used losses Cross Entropy and Focal Loss \cite{lin2017focal}.

\textbf{Experimental Details.} The noise generation, networks, training details, hyper-parameter settings and more experimental results can be found in the supplementary material.

\textbf{Results.} Tables \ref{symmetric-noise} and \ref{asymmetric-noise} report the test accuracy results of each loss function on the benchmark datasets with symmetric label noise and asymmetric label noise, respectively. As we can see, our proposed AGCE, AUL and AEL have a significant improvement in most label noise settings for MNIST and CIFAR-10. For example, compared with GCE, SCE, NLNL and NCE, AUL achieves better test accuracy on MNIST and CIFAR-10 for symmetric noise with any noise rate and asymmetric noise with noise rate $\eta\in\{0.1, 0.2, 0.3\}$. However, in our limited parameter tuning, ALFs suffer from underfitting with asymmetric label noise with $\eta=0.4$. According to Theorems \ref{sy-is-asy} and \ref{linearity}, the proposed asymmetric loss functions can be applied to the APL framework \cite{ma2020normalized}. And our NCE+ALFs, especially NCE+AGCE and NCE+AUL, achieve the top three best results in most test scenarios across all datasets. In several cases, our method are better than all baseline methods. The results demonstrate that asymmetric loss functions can be robust enough to get the outstanding performance for both symmetric and asymmetric label noise.

\begin{table*}[!t]
\small
\centering
\caption{Test accuracies (\%) of different methods on benchmark datasets with asymmetric label noise ($\eta\in[0.1, 0.2, 0.3, 0.4]$). The results (mean$\pm$std) are reported over 3 random runs and the top 3 best results are \textbf{boldfaced}.}
\label{asymmetric-noise}
\begin{tabular}{c|c|cccc}
    \hline
     \multirow{2}*{Datasets} & \multirow{2}*{Methods} & \multicolumn{4}{c}{Asymmetric Noise Rate ($\eta$)}  \\
     ~ & ~ & 0.1 & 0.2 & 0.3 & 0.4\\
     \hline\hline
     \multirow{10}*{MNIST} & CE & 97.57 $\pm$ 0.22 & 94.56 $\pm$ 0.22 & 88.81 $\pm$ 0.10 & 82.27 $\pm$ 0.40\\
    ~ & FL &97.58 $\pm$ 0.09 &94.25 $\pm$ 0.15 &89.09 $\pm$ 0.25 &82.13 $\pm$ 0.49\\
    ~ & GCE &99.01 $\pm$ 0.04 &96.69 $\pm$ 0.12 &89.12 $\pm$ 0.24 &81.51 $\pm$ 0.19\\
    ~ & NLNL &98.63 $\pm$ 0.06 &98.35 $\pm$ 0.01 &97.51 $\pm$ 0.15 &95.84 $\pm$ 0.26\\
    ~ & SCE &\textbf{99.14 $\pm$ 0.04} &98.03 $\pm$ 0.05 &93.68 $\pm$ 0.43 &85.36 $\pm$ 0.17\\
    ~ & NCE &98.49 $\pm$ 0.06 &98.18 $\pm$ 0.12 &96.99 $\pm$ 0.17 &94.16 $\pm$ 0.19\\
    ~ & NCE+RCE &\textbf{99.35 $\pm$ 0.03} &98.99 $\pm$ 0.22 &97.23 $\pm$ 0.20 &90.49 $\pm$ 4.04\\
    \cline{2-6}
     ~ & \textbf{AUL} & \textbf{99.15 $\pm$ 0.09} &\textbf{99.15 $\pm$ 0.02} &\textbf{98.98 $\pm$ 0.05} &\textbf{98.62 $\pm$ 0.09}\\
    ~ & \textbf{AGCE} & 99.10 $\pm$ 0.02 &\textbf{99.07 $\pm$ 0.09} &\textbf{98.95 $\pm$ 0.03} &\textbf{98.44 $\pm$ 0.11}\\
    ~ & \textbf{AEL} & 98.99 $\pm$ 0.05 &\textbf{99.06 $\pm$ 0.07} &\textbf{98.90 $\pm$ 0.15} &\textbf{98.34 $\pm$ 0.08}\\
     \hline\hline
     \multirow{10}*{CIFAR10} & CE & 87.55 $\pm$ 0.14 & 83.32 $\pm$ 0.12 & 79.32 $\pm$ 0.59 & 74.67 $\pm$ 0.38\\
     ~ & FL & 86.43 $\pm$ 0.30 & 83.37 $\pm$ 0.07 & 79.33 $\pm$ 0.08 & 74.28 $\pm$ 0.44\\
     ~ & GCE & 88.33 $\pm$ 0.05 & 85.93 $\pm$ 0.23 & 80.88 $\pm$ 0.38 & 74.29 $\pm$ 0.43\\
     ~ & SCE & 89.77 $\pm$ 0.11 & 86.20 $\pm$ 0.37 & 81.38 $\pm$ 0.35 & 75.16 $\pm$ 0.39\\
     ~ & NLNL & 88.54 $\pm$ 0.25 & 84.74 $\pm$ 0.08 & 81.26$\pm$ 0.43 & 76.97 $\pm$ 0.52\\
     ~ & NCE & 74.06 $\pm$ 0.27 & 72.46 $\pm$ 0.32 & 69.86 $\pm$ 0.51 & 65.66 $\pm$ 0.42\\
    ~ & NCE+RCE & \textbf{90.06 $\pm$ 0.13} & \textbf{88.45 $\pm$ 0.16} & \textbf{85.42 $\pm$ 0.09} & \textbf{79.33 $\pm$ 0.15}\\
     \cline{2-6}
     ~ & \textbf{AUL} & \textbf{90.19 $\pm$ 0.16} & 88.17 $\pm$ 0.11 & 84.87 $\pm$ 0.04 & 56.33 $\pm$ 0.07 \\
     ~ & \textbf{AGCE} & 88.08 $\pm$ 0.06 & 86.67 $\pm$ 0.14 & 83.59 $\pm$ 0.15 & 60.91 $\pm$ 0.20\\
     ~ & \textbf{AEL} & 85.22 $\pm$ 0.15 & 83.82 $\pm$ 0.15 & 82.43 $\pm$ 0.16 & 58.81 $\pm$ 3.62\\
     ~ & \textbf{NCE+AUL} & {90.05 $\pm$ 0.20} & \textbf{88.72 $\pm$ 0.26} & \textbf{85.48 $\pm$ 0.18} & \textbf{79.26 $\pm$ 0.05}\\
     ~ & \textbf{NCE+AGCE} & \textbf{90.35 $\pm$ 0.15} & \textbf{88.48 $\pm$ 0.16} & \textbf{85.96 $\pm$ 0.24} & \textbf{80.00 $\pm$ 0.44}\\
     ~ & \textbf{NCE+AEL} & {89.95 $\pm$ 0.04} & {87.93 $\pm$ 0.06} & {84.81 $\pm$ 0.26} & {77.27 $\pm$ 0.11}\\
     \hline\hline
     \multirow{10}*{CIFAR100} & CE &64.85 $\pm$ 0.37	&58.11 $\pm$ 0.32	&50.68 $\pm$ 0.55	&40.17 $\pm$ 1.31\\
     ~ & FL &64.78 $\pm$ 0.50	&58.05 $\pm$ 0.42	&51.15 $\pm$ 0.84	&\textbf{41.18 $\pm$ 0.68}\\
     ~ & GCE &63.01 $\pm$ 1.01	&59.35 $\pm$ 1.10	&53.83 $\pm$ 0.64	&40.91 $\pm$ 0.57\\
     ~ & SCE &61.63 $\pm$ 0.84	&53.81 $\pm$ 0.42	&45.63 $\pm$ 0.07	&36.43 $\pm$ 0.20\\
     ~ & NLNL & 59.55 $\pm$ 1.22 & 50.19 $\pm$ 0.56 & 42.81 $\pm$ 1.13 & 35.10 $\pm$ 0.20\\
     ~ & NCE &27.59 $\pm$ 0.54	&25.75 $\pm$ 0.50	&24.28 $\pm$ 0.80	&20.64 $\pm$ 0.40\\
     ~ & NCE+RCE &66.38 $\pm$ 0.16	&\textbf{62.97 $\pm$ 0.24}	&\textbf{55.38 $\pm$ 0.49}	&\textbf{41.68 $\pm$ 0.56}\\
     \cline{2-6}
     ~ & \textbf{NCE+AUL} & \textbf{66.62 $\pm$ 0.09} & \textbf{63.86 $\pm$ 0.18} & 50.38 $\pm$ 0.32 & 38.59 $\pm$ 0.48\\
     ~ & \textbf{NCE+AGCE}& \textbf{67.22 $\pm$ 0.12} & \textbf{63.69 $\pm$ 0.19} & \textbf{55.93 $\pm$ 0.38} & \textbf{43.76 $\pm$ 0.70}\\
     ~ & \textbf{NCE+AEL} & \textbf{66.92 $\pm$ 0.22} & 62.50 $\pm$ 0.23 & \textbf{52.42 $\pm$ 0.98} & 39.99 $\pm$ 0.12
     \\
     \hline
\end{tabular}
\vskip-8pt
\end{table*}

\textbf{Visualization.} We further investigate the feature representations learned by AGCE compared to that learned by GCE. We first extract the high-dimensional features at the second last layer, then project all features of test samples in to 2D embeddings by t-SNE \cite{tsne}. The projected representations on MNIST with different symmetric label noise are illustrated in Fig. \ref{fig:tsne-gce-agce}. As can be observed, GCE encounters obvious overfitting with label noise, and the embeddings look completely mixed together when $\eta=0.8$. On the contrary, AGCE learns good representations with more separated and clearly bounded clusters in all noisy cases.

\subsection{Evaluation on Real-world Noisy Label}
To evaluate the effectiveness of asymmetric loss functions, we test on the real-world noisy dataset WebVision \cite{li2017webvision}, where we follow the "Mini" setting in \cite{jiang2018mentornet, ma2020normalized} that only takes the first 50 concepts of the Google resized image subset as the training dataset and further evaluate the trained ResNet-50 \cite{he2016deep} on the same 50 concepts of the corresponding validation set. 

\begin{table}[h]
\vskip-14pt
\setlength{\tabcolsep}{0.8mm}
\small
\centering
\caption{Top-1 validation accuracies (\%) on WebVision validation set using different loss functions.}
\label{webvision}
\begin{tabular}{c|ccccccc}
    \hline
    Loss & CE & GCE & SCE & NCE+RCE & \textbf{NCE+AGCE} & \textbf{AGCE}\\
    \hline
    Acc & 66.96 & 61.76 & 66.92 & 66.32 & \textbf{67.12} & \textbf{69.40}\\ 
    \hline
\end{tabular}
\end{table}
\vskip-7pt

The top-1 validation accuracies under different loss functions on the clean WebVision validation set are reported in Table \ref{webvision}. More experimental details and results can be found in supplementary materials. As shown in Table \ref{webvision}, the proposed loss functions AGCE and NCE+AGCE outperform the existing loss functions GCE, SCE, and NCE+RCE. The results demonstrate that asymmetric loss functions can help the trained model against real-world label noise.

\section{Conclusion}
This paper introduces asymmetric loss functions, which allow training a noise-tolerant classifier with noisy labels as long as clean labels dominate. We then prove that completely asymmetric losses are classification-calibrated, and have an excess risk bound when the asymmetry is strict. Furthermore, we introduce the asymmetric ratio to measure the asymmetry. The empirical results demonstrate that the larger ratio will provide better robustness. We also prove asymmetric loss functions will provide a global clean weighted-risk when minimizing the noisy risk for any hypothesis class. The experiments on benchmark datasets show the advantage of using the modified loss functions.

\section*{Acknowledgement}
This work was supported by National Natural Science Foundation of China under Grants 61922027, 61827804 and 61932022, and by National Key R\&D Program of China under Grant 2018AAA0102801 and 2019YFE0109600.


\newtheorem{thm}{Theorem}
\newenvironment{thmbis}[1]
  {\renewcommand{\thethm}{#1}%
   \addtocounter{thm}{-1}%
   \begin{thm}}
  {\end{thm}}

\newtheorem{coro}{Corollary}
\newenvironment{corobis}[1]
  {\renewcommand{\thecoro}{#1}%
   \addtocounter{coro}{-1}%
   \begin{coro}}
  {\end{coro}}

\newtheorem{lem}{Lemma}
\newenvironment{lembis}[1]
  {\renewcommand{\thelem}{#1}%
   \addtocounter{lem}{-1}%
   \begin{lem}}
  {\end{lem}}

\onecolumn
\icmltitle{Asymmetric Loss Functions for Learning with Noisy Labels:\\ Supplementary Materials}

\appendix
\section{More Analysis about Clean Labels Domination Assumption}
For robust training, we assume that samples in the training dataset have a larger probability of keeping their true semantic label than the wrong class labels, which is referred to as \textit{clean labels domination assumption}. In the following, we provide more intuitive analysis about this assumption to show its reasonability.

\begin{figure}[htb]
    \centering
    \subfigure[Class ``cat" dominates]{
    \label{clda}
    \includegraphics[width=3in]{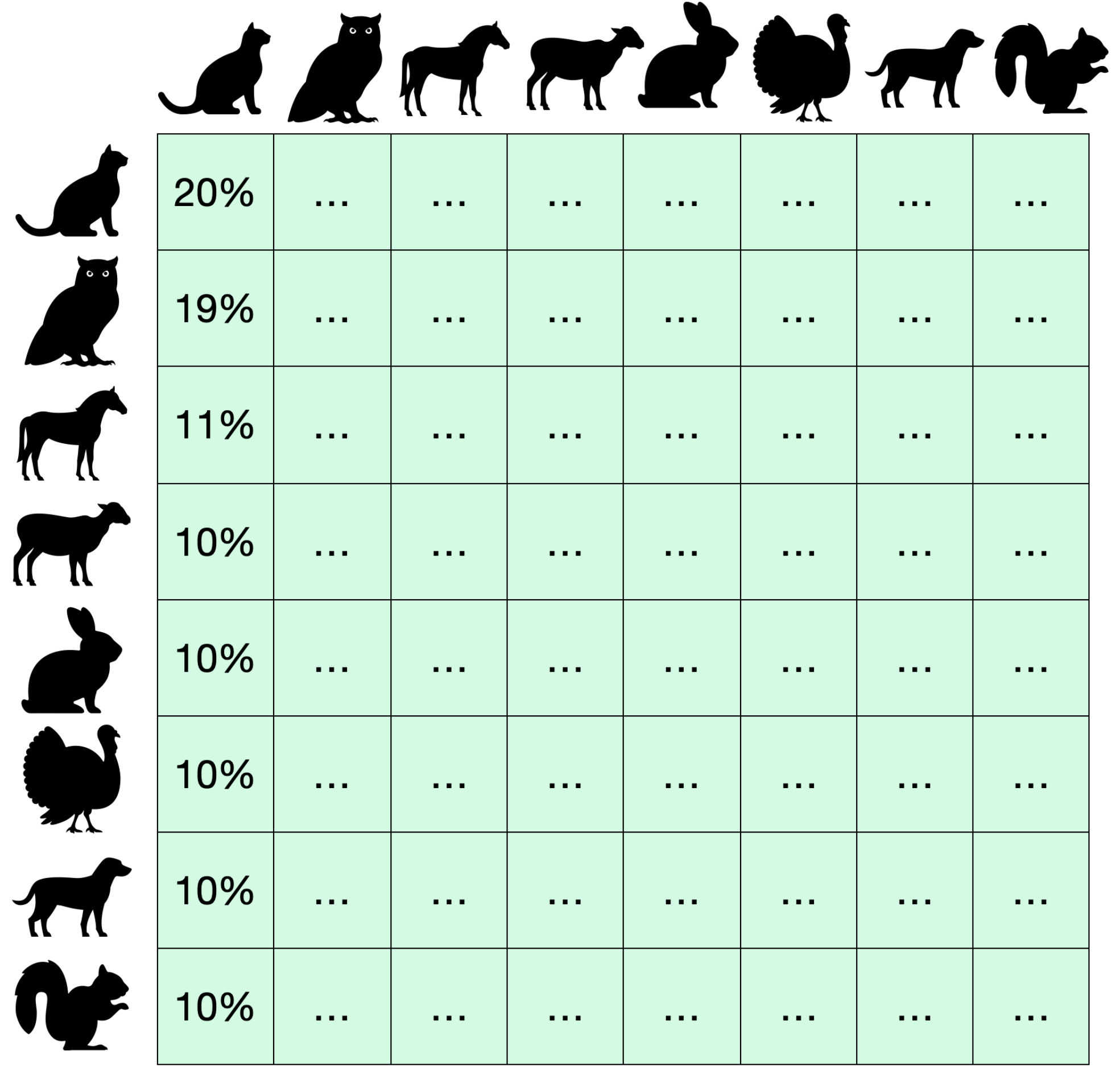}
    }
    \quad
    \subfigure[Class ``cat" does not dominate]{
    \label{cldb}
    \includegraphics[width=3in]{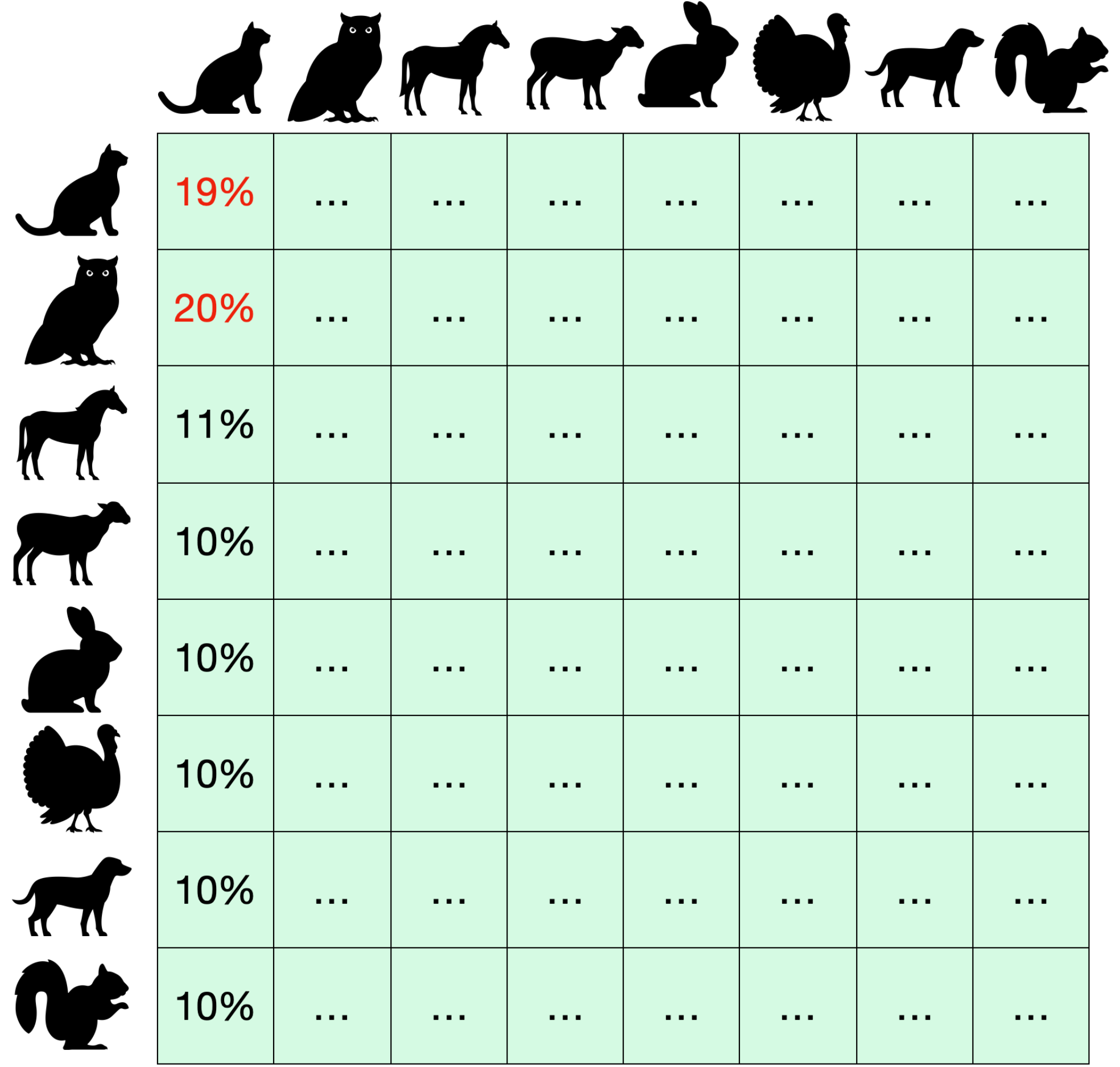}
    }
    \caption{Illustration of label noise model under clean-labels-dominate and -non-dominate settings.}
    \label{cld}
\end{figure}

In Figure 1, label noise models under clean-labels-dominate and -non-dominate setting are shown, from which an intuitive understanding about the clean labels domination assumption can be derived. Figure \ref{clda} and \ref{cldb} exhibit noise transmission matrices, which denote the probability of flipping the class of columns to the class of rows. In Figure \ref{clda}, \textit{cats} have a 20\% probability of keeping the true label, while having smaller probability of wrongly flipping to labels of any other classes. For example, they have a 19\% probability to be annotated as \textit{owls}. In this case, we call \textit{cats} clean-labels-dominant, since images with true cats labels dominate in the cats class, a classifier can be learned to correctly separate \textit{cats} from other classes by classifying a sample to the dominant class . In Figure \ref{cldb}, the situation is reversed, where \textit{cats} have bigger probability of flipping to \textit{owls} than keeping the true label, which is denoted as the case of clean-labels-non-dominate. It means that \textit{owls} account for the largest proportion in \textit{cats} class, which sounds ridiculous. On the other hand, without the help of prior knowledge, even if there exists a learned classifier that works well in a clean-labels-non-dominant dataset, it would produce wrong results on a clean-labels-dominant dataset since it tends to classify a sample into a non-dominant class rather than the corresponding dominant class (\textit{i.e}., the true class).

\section{Classification calibration and Excess Risk Bound}
In the binary classification problem with label set $\{0, 1\}$ which is different from $\{-1, 1\}$, we need to slightly modify the definition of classification calibration in \cite{2003Statistical, 2006Convexity}.

Let $f(\mathbf x)$ denote the predictive result of $p(y=1|\mathbf x)$, and $R_{\ell}(f)$ denote the risk of a classifier $f$ based on a loss function $\ell$ or the $\ell$-risk, i.e., $R_{\ell}(f)=\mathbb{E} \ell(f(\mathbf x), y)$. And the risk of a global minimizer is $R_{\ell}^*=\inf_{f} R_{\ell}(f)$.

For the zero-one loss $\ell_{0-1}$, we have $R_{\ell_{0-1}}(f)=\mathbb{E}[\mathbb I(\mathop{\text{sign}}(f(\mathbf x)-1/2)\not = \mathop{\text{sign}}(y-1/2))].$ $R^*$ denote the Bayes risk, i.e.,
\begin{equation}
    R_{\ell_{0-1}}^*(f)=\inf_{f} R_{\ell_{0-1}}(f)
\end{equation}
Given a loss function $\ell(t)$ (eg., exponential loss, cross entropy loss, or unhinged loss), where $t=yf(\mathbf x)+(1-y)(1-f(\mathbf x))$ is the predictive probability of data point $(x, y)$, the \textit{conditional $\ell-$risk} is defined as
\begin{equation}
    C_{\eta_{\mathbf x}}(f(\mathbf x), \ell)=\mathbb E_{y|\mathbf x}[\ell(y f(\mathbf x)+(1-y)(1-f(\mathbf x))]=\eta_{\mathbf x}\ell(f(\mathbf x))+(1-\eta_{\mathbf x})\ell(1-f(\mathbf x))
\end{equation}
where $\eta_{\mathbf x}=p(y=1|\mathbf x)$. Similarly, we define the "optimal $\ell$-risk" as
\begin{equation}
    R_\ell^{*}=\inf_f R_\ell(f)=\inf_f \mathbb E[\eta_{\mathbf x}\ell(f(\mathbf x))+(1-\eta_{\mathbf x})\ell(1-f(\mathbf x))]
\end{equation}
When $\ell$ is the \textit{zero-one} loss, we obtain the Bayes-optimal classifier $\mathbb{I}(\eta_{\mathbf x}>\frac{1}{2})$.

The \textit{excess risk} for a classifier $f$ is given by $R_{\ell_{0-1}}(f)-R^{*}_{\ell_{0-1}}$, and the "excess $\ell$-risk" is $R_\ell(f)-R_\ell^*$.

For a fixed value of $\mathbf x$, the minimum of the expectation is given by
\begin{equation}
\label{H}
    H_\ell(\eta)=\inf_{\alpha\in[0, 1]} (\eta\ell(\alpha)+(1-\eta)\ell(1-\alpha)),
\end{equation}
so we write
\begin{equation}
    R_\ell^{*}=\mathbb E[H_\ell(\eta_\mathbf x)].
\end{equation}
For a good classifier, we want $\text{sign}(f(\mathbf x)-\frac{1}{2})=\text{sign}(\alpha-\frac{1}{2})=\text{sign}(\eta-\frac{1}{2})$, i.e., $(\alpha-\frac{1}{2})(\eta-\frac{1}{2})\ge 0$. So we define a quantity similar to Eq. \ref{H} but optimized only where $\alpha$ is not a good classifier:
\begin{equation}
    \label{H-}
    H_\ell^-(\eta)=\inf_{\{\alpha: (\alpha-\frac{1}{2} )(\eta-\frac{1}{2})\le 0,\alpha\in[0,1]\}}(\eta \ell(\alpha)+(1-\eta)\ell(1-\alpha)).
\end{equation}
We define a loss function $\ell$ to be "classification-calibrated" if $H^-_\ell(\eta)>H_\ell(\eta)$ for all $\eta\not = \frac{1}{2}$. Intuitively, this means that the loss function strictly penalizes a classifier $f$ for not classifying in accordance with $\eta_{\mathbf x}$.
\subsection{Classification-calibration}
\begin{thmbis}{1}
Completely asymmetric loss functions are classification-calibrated.
\end{thmbis}
\begin{proof}
For any weights $w_1, w_2$ and $w_1\not = w_2$, we define a completely asymmetric loss function $\ell$  as follows
\begin{equation}
    \mathop{\arg\min}\limits_{u\in[0,1]}w_1\ell(u)+w_2\ell(1-u)=\mathbb I[w_1> w_2],
\end{equation}
i.e., $w_1\ell(u)+w_2\ell(1-u)\ge \mathbb I[w_1> w_2]\cdot[w_1\ell(1)+w_2\ell(0)]+\mathbb I[w_1<w_2]\cdot[w_1\ell(0)+w_2\ell(1)]$, and the equality holds if and only if $u=\mathbb I(w_1> w_2)$. In other words, the \textbf{conditional risk minimizer} of $\ell$ can be expressed as $\mathbb{I}(\eta_{\mathbf x}>1-\eta_{\mathbf x})$, which is equivalent to the Bayes-optimal classifier $\mathbb{I}(\eta_{\mathbf x}>\frac{1}{2})$.

Then if $\ell$ is asymmetric on $\eta, 1-\eta$, where $\eta\not =\frac{1}{2}$, we have
\begin{equation}
    H_\ell(\eta) =\inf_{\alpha\in[0,1]}(\eta \ell(\alpha)+(1-\eta)\ell (1-\alpha)=\begin{cases}
    \eta\ell(1)+(1-\eta)\ell(0), & \eta>\frac{1}{2}\\
    \eta\ell(0)+(1-\eta)\ell(1), & \eta<\frac{1}{2}
    \end{cases}
\end{equation}
and
\begin{equation}
    H_\ell^-(\eta)
    =\begin{cases}
    \inf_{0\le\alpha\le \frac{1}{2}}(\eta \ell(\alpha)+(1-\eta)\ell(1-\alpha)), & \eta> \frac{1}{2}\\
    \inf_{\frac{1}{2}\le \alpha \le 1}(\eta \ell(\alpha)+(1-\eta)\ell(1-\alpha)), & \eta< \frac{1}{2}\\
    \end{cases}.
\end{equation}
Because $\ell$ is asymmetric on $\eta$, $1-\eta$, then for all  $\eta>\frac{1}{2}$, we have
\begin{equation}
    \inf_{0\le\alpha\le \frac{1}{2}}(\eta \ell(\alpha)+(1-\eta)\ell(1-\alpha))>\eta \ell(1)+(1-\eta)\ell(0),
\end{equation}
and for all $\eta<\frac{1}{2}$,
\begin{equation}
    \inf_{\frac{1}{2}\le \alpha \le 1}(\eta \ell(\alpha)+(1-\eta)\ell(1-\alpha)) > \eta\ell(0)+(1-\eta)\ell(1).
\end{equation}

so it follows that $H_\ell^-(\eta)>H_\ell(\eta)$ for all $\eta\not=\frac{1}{2}$, so asymmetric loss functions are classification-calibrated.
\end{proof}

\subsection{Excess Risk Bound}

\begin{thmbis}{2}
An excess risk bound of a strictly and completely asymmetric loss function $L(\mathbf u, i)=\ell(u_i)$ can be expressed as
\begin{equation}
    R_{\ell_{0-1}}(f)-R_{\ell_{0-1}}^{*}\le \frac{2(R_{\ell}(f)-R_{\ell}^{*})}{\ell(0)-\ell(1)},
\end{equation}
where $R_{\ell_{0-1}}^{*}=\inf_{g} R_{\ell_{0-1}}(g)$ and $R_{\ell}^{*}=\inf_{g} R_{\ell}(g)$.
\end{thmbis}
\begin{proof}
Consider a loss function $\ell$, the transform $\tilde{\psi}:[-1,1]\rightarrow R_+$ from \cite{2006Convexity} is defined as
\begin{equation}
   \begin{aligned}
        \tilde{\psi}(\theta)&=H^-_{\ell}\left(\frac{1+\theta}{2}\right)-H_{\ell}\left(\frac{1+\theta}{2}\right)\\
   \end{aligned}
\end{equation}

For $\theta\in(0,1]$, we have
\begin{equation}
   \begin{aligned}
        \tilde{\psi}(\theta)&=H^-_{\ell}\left(\frac{1+\theta}{2}\right)-H_{\ell}\left(\frac{1+\theta}{2}\right)\\
        &= \inf_{0\le\alpha\le \frac{1}{2}}\left[\frac{1+\theta}{2}\ell(\alpha)+\frac{1-\theta}{2}\ell(1-\alpha)\right]-\left[\frac{1+\theta}{2}\ell(1)+\frac{1-\theta}{2}\ell(0)\right]\\
        &=\frac{1}{2}[2\ell(1/2)-\ell(0)-\ell(1)]+\frac{\theta}{2}[\ell(0)-\ell(1)].
   \end{aligned}
\end{equation}
where $\frac{1+\theta}{2}\ell(\alpha)+\frac{1-\theta}{2}\ell(1-\alpha)\ge \frac{1+\theta}{2}\ell(1/2)+\frac{1-\theta}{2}\ell(1/2)$, for $\alpha\in[0,1/2]$, since $\ell$ is strictly asymmetric.

For $\theta\in[-1, 0)$, we have
\begin{equation}
    \begin{aligned}
        \tilde{\psi}(\theta)&=H^-_{\ell}\left(\frac{1+\theta}{2}\right)-H_{\ell}\left(\frac{1+\theta}{2}\right)\\
        &= \inf_{\frac{1}{2}\le\alpha\le1}\left[\frac{1+\theta}{2}\ell(\alpha)+\frac{1-\theta}{2}\ell(1-\alpha)\right]-\left[\frac{1+\theta}{2}\ell(0)+\frac{1-\theta}{2}\ell(1)\right]\\
        &=\frac{1}{2}[2\ell(1/2)-\ell(0)-\ell(1)]-\frac{\theta}{2}[\ell(0)-\ell(1)].
   \end{aligned}
\end{equation}
where $\frac{1+\theta}{2}\ell(\alpha)+\frac{1-\theta}{2}\ell(1-\alpha)\ge \frac{1+\theta}{2}\ell(1/2)+\frac{1-\theta}{2}\ell(1/2)$, for $\alpha\in[1/2,1]$, since $\ell$ is strictly asymmetric.

We can see that $\tilde{\psi}$ is symmetric about 0, i.e., $\tilde{\psi}(-t)=\tilde{\psi}(t)$, and $\tilde{\psi}(0)=\frac{1}{2}[2\ell(1/2)-\ell(0)-\ell(1)]\ge 0$. Therefore,  $\tilde{\psi}(\theta)$ is convex. For simplicity, let $\sigma(t)=\mathbb(t-1/2)$. Then, according to Jensen's inequality, we have
$$
    \begin{aligned}
        &\tilde{\psi}(R_{\ell_{0-1}}(f)-R^{*}_{\ell_{0-1}})\\
        & = \tilde{\psi}\left(\mathbb{E}[\mathbb I(\sigma(f(\mathbf x))\not = \sigma(\eta_{\mathbf x})|2\eta_{\mathbf x}-1|]\right)\\
        &\le \mathbb{E}[\tilde{\psi}\left(\mathbb I(\sigma(f(\mathbf x))\not = \sigma(\eta_{\mathbf x})|2\eta_{\mathbf x}-1|\right)]\\
        & = \mathbb{E}\left[\mathbb{I}(\sigma(f(\mathbf x))=\sigma( \eta_{\mathbf x}))\cdot\tilde{\psi}(0)\right] + \mathbb{E}\left[\mathbb{I}(\sigma(f(\mathbf x))\not=\sigma( \eta_{\mathbf x}))\cdot\tilde{\psi}(|2\eta_{\mathbf x}-1|)\right]\\
        & \le \tilde{\psi}(0) + \mathbb{E}\left[\mathbb{I}(\sigma(f(\mathbf x))\not=\sigma( \eta_{\mathbf x}))\cdot\left(H^-_{\ell}(\eta_{\mathbf x}) - H_{\ell}(\eta_{\mathbf x})\right)\right]\\
        & = \tilde{\psi}(0) + \mathbb{E}\left[\mathbb{I}(\sigma(f(\mathbf x))\not=\sigma( \eta_{\mathbf x}))\cdot\left(\inf_{\{\alpha: (\alpha-\frac{1}{2} )(\eta_{\mathbf x}-\frac{1}{2})\le 0,\alpha\in[0,1]\}}C_{\eta_{\mathbf x}}(\alpha, \ell) - H_{\ell}(\eta_{\mathbf x})\right)\right]\\
        &\le \tilde{\psi}(0) + \mathbb{E}\left[\mathbb{I}(\sigma(f(\mathbf x))\not=\sigma( \eta_{\mathbf x}))\cdot (C_{\eta_{\mathbf x}}(f(\mathbf x), \ell)-H_\ell(\eta_{\mathbf x}))\right]\\
        &\le \tilde{\psi}(0) + \mathbb{E}\left[\mathbb{I}(\sigma(f(\mathbf x))\not=\sigma( \eta_{\mathbf x}))\cdot (C_{\eta_{\mathbf x}}(f(\mathbf x), \ell)-H_\ell(\eta_{\mathbf x}))\right] + \mathbb{E}\left[\mathbb{I}(\sigma(f(\mathbf x))=\sigma( \eta_{\mathbf x}))\cdot (C_{\eta_{\mathbf x}}(f(\mathbf x), \ell)-H_\ell(\eta_{\mathbf x}))\right]\\
        & = \tilde{\psi}(0) + \mathbb{E}\left[C_{\eta_{\mathbf x}}(f(\mathbf x), \ell)-H_\ell(\eta_{\mathbf x})\right]\\
        & = \tilde{\psi}(0) + R_{\ell}(f)-R^{*}_{\ell},
    \end{aligned}
$$
where we have used the fact that for any $\mathbf x$, and in particular when $\text{sign}(f(\mathbf x)-1/2)=\text{sign}(\eta_{\mathbf x}-1/2)$, $C_{\eta_{\mathbf x}}(f(\mathbf x), \ell)\ge H_\ell(\eta_{\mathbf x})$. On the other hand, since $\tilde{\psi}(\theta)= \tilde{\psi}(0) +  \frac{|\theta|}{2}[\ell(0)-\ell(1)]$, we have
\begin{equation}
   \tilde{\psi}(0) + \frac{R_{\ell_{0-1}}(f)-R^{*}_{\ell_{0-1}}}{2}[\ell(0)-\ell(1)]= \tilde{\psi}(R_{\ell_{0-1}}(f)-R^{*}_{\ell_{0-1}})\le \tilde{\psi}(0) +  R_{\ell}(f)-R^{*}_{\ell},
\end{equation}
i.e.,  we obtain the excess risk bound as follows
\begin{equation}
    R_{\ell_{0-1}}(f)-R_{\ell_{0-1}}^{*}\le \frac{2(R_{\ell}(f)-R_{\ell}^{*})}{\ell(0)-\ell(1)}.
\end{equation}
The result suggests that the excess risk bound of any completely asymmetric loss function is controlled only by the difference of $\ell(0)-\ell(1)$. 
Intuitively, the excess risk bound suggests that if the prediction function $f$ minimizes the surrogate risk $R_\ell(f)=R^{*}_\ell$, then the prediction function $f$ must also minimize the misclassification risk $R_{\ell_{0-1}}(f)=R_{\ell_{0-1}}^{*}$.
\end{proof}
 
\section{Proof of Theorems and Corollaries}

\begin{thmbis}{3}
Symmetric loss functions are completely asymmetric.
\end{thmbis}
\begin{proof}
    For any weights $w_1,...,w_k$, $\exists t$, s.t., $w_t>\max_{i\not = t}w_i$, i.e., $w_i-w_t<0$. Let $L$ be a symmetric loss function, then
\begin{equation}
    \begin{aligned}
        \sum_{i=1}^k w_i L(\mathbf u,i)&= w_t L(\mathbf u,t)+\sum_{i\not = t}w_i L(\mathbf u,i)\\
            &=w_t C+\sum_{i\not =t}(w_i-w_t)L(\mathbf u,i)\\
        &\ge w_t C + \min_{\mathbf u\in U'} \sum_{i\not =t}(w_i-w_t)L(\mathbf u,i)
\end{aligned}
\end{equation}
where $U'=\{\mathbf u:\sum_{i\not=t}L(\mathbf u,i)=C-\min_\mathbf u L(\mathbf u, t)\}=\{\mathop{\arg\min}\limits_{\mathbf u} L(\mathbf u,t)\}$. Therefore, $\mathop{\arg\min}\limits_{\mathbf u}\sum_{i=1}^k w_i L(\mathbf u, i)= \mathop{\arg\min}\limits_{\mathbf u} L(\mathbf u,t)$, i.e., $L$ is a completely asymmetric loss function.
\end{proof}

\subsection{Proof for theorems}
\begin{thmbis}{4}[Noise-Tolerance]
In a multi-classification problem, given an appropriate neural network class $\mathcal H$  which satisfies Assumption 7, then the loss function $L$ is noise-tolerant if $L$ is asymmetric on the label noise model.
\end{thmbis}

\begin{proof}
     Let $f^*=\arg\min_{f\in\mathcal H} R_L^\eta(f)$, when we regard the conditional risk $L^{\eta}(\mathbf x, y)$ as a new loss function, then $f^*$ minimizes $L^\eta(f(\mathbf x), y)$ for each $(\mathbf x,y)$. Because $L$ is an asymmetric loss and $1-\eta_\mathbf x$ is bigger than $\eta_{\mathbf x,i}$, $f^*$ also minimizes $L(f(\mathbf x), y)$. Therefore, we have 
     $$
     R_L(f)=\mathbb E_{\mathbf x, y} L(f(\mathbf x),y)\ge \mathbb E_{\mathbf x,y}L(f^*(\mathbf x),y)=R_L(f^*),
     $$
     so $f^*$ minimizes $R_L(f)$.
\end{proof}

\begin{thmbis}{5}
$\forall \alpha,\ \beta>0$, if $L_1$ and $L_2$ are asymmetric, then $\alpha L_1+\beta L_2$ is asymmetric.
\end{thmbis}
\begin{proof}
Given weights $w_1,...,w_k$, $w_t>\max_{i\not = t}w_i$, because $L_1$ and $L_2$ are asymmetric, let $\mathbf u^*=\mathbf e_t= \arg\min_{\mathbf u} L_1(\mathbf u,t)=\arg\min_{\mathbf u} L_2(\mathbf u,t)$, i.e.,
\begin{equation}
\begin{aligned}
    &\sum_{i=1}^kw_i L_1(\mathbf u,i)\ge \sum_{i=1}^kw_i L_1(\mathbf u^*,i)\quad \text{and}\\
    &\sum_{i=1}^kw_i L_2(\mathbf u,i)\ge \sum_{i=1}^kw_i L_2(\mathbf u^*,i)
\end{aligned}
\end{equation}
Then we have $\sum_{i=1}^kw_i [\alpha L_1(\mathbf u,i)+\beta L_2(\mathbf u,i)]\ge \sum_{i=1}^kw_i (\alpha L_1(\mathbf u^*,i)+\beta L_2(\mathbf u^*,i)]$, and the equality holds if and only if $\mathbf u=\mathbf u^*$,  so $\alpha L_1+\beta L_2$ is asymmetric.
\end{proof}

\begin{lembis}{1}
\label{lemma-strict}
Consider a loss function $L(\mathbf u,i)=\ell (u_i)$, for any $w_1>w_2\ge0$, $\mathbf u\in \mathcal C$, if $\ell$ satisfies $w_1 \ell(u_1)+w_2\ell(u_2)\ge w_1\ell(u_1+u_2)+w_2\ell(0)$, and the equality holds only if $u_2=0$,  then $L$ is completely asymmetric.
\end{lembis}
\begin{proof}
Given any weights $w_1,...,w_k$, $w_t>\max_{i\not = t}w_i$, the optimal solution is $\mathbf u^*=\mathbf e_t$, then
\begin{equation}
    \begin{aligned}
        \sum_{i=1}^k w_i L(\mathbf u,i)&=w_t \ell(u_t)+\sum_{i\not = t}w_i\ell(u_i)\\
        &\ge w_t \ell(u_t+\sum_{i\not =t}u_i) + \sum_{i\not = t}w_i \ell(0)\\
        &=\sum_{i=1}^k w_i L(\mathbf u^*, i)
\end{aligned}
\end{equation}
The equality holds if and only if $u_i=0$, for $i\not = t$, i.e., $\mathbf u^*$ is the only one minimizes $ \sum_{i=1}^k w_i L(\mathbf u,i)$, so $L$ is completely asymmetric.
\end{proof}

\begin{thmbis}{6}[Sufficiency]
On the given weights $w_1,..,w_k$, where $w_m > w_n$ and $w_n=\max_{i\not = m}w_i$, the loss function $L(\mathbf u,i)=\ell(u_i)$ is asymmetric if $\frac{w_m}{w_n}\cdot r(\ell)\ge 1$. 
\end{thmbis}
\begin{proof}
If $\frac{w_m}{w_n}\cdot r(\ell)\ge 1$, then for any $i\not =m$, we have
\begin{equation}
    \begin{aligned}
    \frac{w_m}{w_i}&\ge\frac{1}{r(\ell)}\ge\sup_{\substack{0\le u_m,u_i\le 1\\ u_m+u_i\le 1}} \frac{\ell(0)-\ell(u_i)}{\ell(u_m)-\ell(u_m+u_i)}\ge \frac{\ell(0)-\ell(u_i)}{\ell(u_m)-\ell(u_m+u_i)}
    \end{aligned}
\end{equation}
i.e., $w_m \ell(u_n)+w_i\ell(u_i)\ge w_m \ell(u_m+u_i)+w_i \ell(0)$, so $L$ is asymmetric according to Theorem \ref{lemma-strict}.
\end{proof}

\begin{thmbis}{7}
In a binary classification problem, we assume that $L$ is strictly asymmetric on the label noise model which keeps dominant, for any $\mathcal H$, let $f^*=\arg\min_{f\in\mathcal H}R_L^{\eta}(f)$. If $\forall \mathbf x$, $\frac{1-\eta_{\mathbf x}}{\eta_{\mathbf x}}\cdot r(L)>1$ hold, then $f^*$ also minimizes a positive weighted $L$-risk $R_{w, L}(h)=\mathbb E w(\mathbf x, y)L(f(\mathbf x), y)$.
\end{thmbis}
\begin{proof}
Without loss of generality, let the label set be \{0,1\}, and $f^*=\arg\min_{f\in\mathcal H} R_L^{\eta}(f)$, then we have
\begin{equation}
\begin{aligned}
&R_L^{\eta}(f^*)-R_L^{\eta}(f)\\
=&\mathbb E_{\mathbf x,y}\Big[(1-\eta_{\mathbf x})\big[L(f^*(\mathbf x),y)-L(f(\mathbf x),y)\big]+\eta_{\mathbf x}\big[L(f^*(\mathbf x), 1-y)-L(f(\mathbf x), 1-y)\big]\Big]\\
=&\mathbb E_{\mathbf x,y}\mathbb I(f^*(\mathbf x)_y< f(\mathbf x)_y)\Big[(1-\eta_{\mathbf x})\big[L(f^*(\mathbf x),y)-L(f(\mathbf x),y)\big]+\eta_{\mathbf x}\big[L(f^*(\mathbf x), 1-y)-L(f(\mathbf x), 1-y)\big]\Big]+\\
&\mathbb E_{\mathbf x,y}\mathbb I(f^*(\mathbf x)_y> f(\mathbf x)_y)\Big[(1-\eta_{\mathbf x})\big[L(f^*(\mathbf x),y)-L(f(\mathbf x),y)\big]+\eta_{\mathbf x}\big[L(f^*(\mathbf x), 1-y)-L(f(\mathbf x), 1-y)\big]\Big]\\
\ge& \mathbb E_{\mathbf x,y}\mathbb I(f^*(\mathbf x)_y< f(\mathbf x)_y)\Big[(1-\eta_{\mathbf x})\big[L(f^*(\mathbf x),y)-L(f(\mathbf x),y)\big]-\frac{\eta_{\mathbf x}}{r(L)}\big[L(f^*(\mathbf x), y)-L(f(\mathbf x), y)\big]\Big]+\\
&\mathbb E_{\mathbf x,y}\mathbb I(f^*(\mathbf x)_y> f(\mathbf x)_y)\Big[(1-\eta_{\mathbf x})\big[L(f^*(\mathbf x),y)-L(f(\mathbf x),y)\big]+\frac{\eta_{\mathbf x}}{r(L)}\big[L(f^*(\mathbf x),y)-L(f(\mathbf x),y)\big]\Big]\\
=&\mathbb E_{\mathbf x,y}w(\mathbf x, y)L(f^*(\mathbf x),y)-\mathbb E_{\mathbf x,y}w(\mathbf x, y)L(f(\mathbf x),y)
\end{aligned}
\end{equation}
where we have
\begin{equation}
    L(f^*(\mathbf x), 1-y)-L(f(\mathbf x), 1-y) \ge\begin{cases}
    -\frac{1}{r(L)}\big[L(f^*(\mathbf x),y)-L(f(\mathbf x),y)\big], & f^*(\mathbf x)_y<f(\mathbf x)_y\\
    \frac{1}{r(L)}\big[L(f^*(\mathbf x),y)-L(f(\mathbf x),y)\big],  & f^*(\mathbf x)_y>f(\mathbf x)_y
    \end{cases}
\end{equation}
and 
\begin{equation}
0< w(\mathbf x, y)=\begin{cases}
(1-\eta_{\mathbf x}-\frac{\eta_{\mathbf x}}{r(L)}), & f^*(\mathbf x)_y<f(\mathbf x)_y\\
1-\eta_{\mathbf x}, &f^*(\mathbf x)_y=f(\mathbf x)_y\\
(1-\eta_{\mathbf x}+\frac{\eta_{\mathbf x}}{r(L)}), & f^*(\mathbf x)_y<f(\mathbf x)_y\\
\end{cases}
\end{equation}

Otherwise, $R_L^{\eta}(f^*)-R_L^{\eta}(f)\le 0$, so we obtain
\begin{equation}
    \mathbb E_{\mathbf x,y}w(\mathbf x, y)L(f^*(\mathbf x),y)\le\mathbb E_{\mathbf x,y}w(\mathbf x, y)L(h(\mathbf x),y),
\end{equation}
i.e., $f^*$ also minimizes the positive weighted $L$-risk $\mathbb E_{\mathbf x,y}w(\mathbf x, y)L(f^*(\mathbf x),y)$.

\end{proof}

\begin{thmbis}{8}[Necessity]
On the given weights $w_1,..,w_k$, where $w_m > w_n$ and $w_n=\max_{i\not = m}w_i$, the loss function $L(\mathbf u,i)=\ell(u_i)$ is asymmetric only if $\frac{w_m}{w_n}\cdot r_{u}(\ell)\ge 1$. 
\end{thmbis}
\begin{proof}
If the loss function $L_q(\mathbf u,i)=\ell(u_i)$ is asymmetric, then for $w_m>w_n$, let $u_i=0$, $i\not=m,n$, then $w_m \ell(u_m) + w_n \ell(u_n)\ge w_m \ell(1) + w_n \ell(0)$ always holds, i.e.,
\begin{equation}
    \frac{w_m}{w_n}\cdot\inf_{\substack{0\le u_m,u_n\le 1\\ u_m+u_n= 1}}\frac{\ell(u_m)-\ell(1)}{\ell(0)-\ell(u_n)}\ge 1,
\end{equation}
so $\frac{w_m}{w_n}\cdot r_u(\ell)\ge 1$.
\end{proof}

\subsection{Proof of corollaries}
\begin{corobis}{1}
On the given weights $w_1,..,w_k$, where $w_m > w_n$ and $w_n=\max_{i\not = m}w_i$, the loss function $L_q(\mathbf u,i)=[(a+1)^q-(a + u_i)^q]/q$ (where $q>0$, $a\ge 0$) is asymmetric if and only if $\frac{w_m}{w_n}\ge (\frac{a+1}{a})^{1-q}\cdot\mathbb{I}(q\le 1)+\mathbb{I}(q>1)$.
\end{corobis}
\begin{proof}
$\Rightarrow$ If the loss function $L_q(\mathbf u,i)=\ell(u_i)$ is asymmetric, then for $w_m>w_n$, let $u_i=0$, $i\not=m,n$, then $w_m\ell(u_m) + w_n \ell(u_n)\ge w_m \ell(1) + w_n \ell(0)$ always holds, i.e.,
\begin{equation}
    w_m[(a+1)^q-(a+u_m)^q]\ge w_n[(a+u_n)^q-a^q].
\end{equation}

\begin{equation}
    \frac{(a+u_1+\Delta u)^q-(a+u_1)^q}{(a+u_2)^q-(a+u_2-\Delta u)^q}
\end{equation}

so we have 
$$
    \frac{w_m}{w_n}\ge \sup_{0\le u\le 1}\frac{(a+1-u)^q-a^q}{(a+1)^q-(a+u)^q}.
$$
$\text{RHS}$ equals to $(\frac{a+1}{a})^{1-q}$ if $q\le 1$, and equals to $1$ when $q>1$.

$\Leftarrow$ According to Theorem \ref{lemma-strict}, $L$ is asymmetric
$$
\begin{aligned}
    &\Leftarrow w_m \ell(u_m)+w_i \ell(u_i) \ge w_m \ell(u_m+u_i) + w_i \ell(0)\\
    &\Leftrightarrow \frac{w_m}{w_i}\ge \sup_{\substack{u_i,u_m\ge0\\ u_i+u_m\le1}}\frac{\ell(0)-\ell(u_i)}{\ell(u_m)-\ell(u_m+u_i)}\\
    &\Leftrightarrow \frac{w_m}{w_i}\ge \sup_{\substack{u_i,u_m\ge0\\ u_i+u_m\le1}}\frac{(a+u_i)^q-a^q}{(a+u_i+u_m)^q-(a+u_m)^q}\\
    &\Leftrightarrow \frac{w_m}{w_i}\ge \mathbb{I}(q\le 1)\cdot \sup_{0\le u_m\le 1}\left(\frac{a+u_m}{a}\right)^{1-q}+\mathbb{I}(q>1)\\
    &\Leftrightarrow \frac{w_m}{w_i}\ge \left(\frac{a+1}{a}\right)^{1-q}\cdot\mathbb{I}(q\le 1)+\mathbb{I}(q>1).
\end{aligned}
$$
On the other hand, if $\frac{w_m}{w_n}\ge (\frac{a+1}{a})^{1-q}\cdot\mathbb{I}(q\le 1)+\mathbb{I}(q>1)$. Then for any $i\not =m$, we have $\frac{w_m}{w_i}\ge (\frac{a+1}{a})^{1-q}\cdot\mathbb{I}(q\le 1)+\mathbb{I}(q>1)$.
\end{proof}

\begin{corobis}{2}
On the given weights $w_1,..,w_k$, where $w_m > w_n$ and $w_n=\max_{i\not = m}w_i$. The loss function $L_p(\mathbf u,i)=[(a-u_i)^p-(a-1)^p]/p$ (where $p>0$ and $a\ge 1$) is asymmetric if and only if $\frac{w_m}{w_n}\ge (\frac{a}{a-1})^{p-1}\cdot\mathbb{I}(p> 1)+\mathbb{I}(p\le1)$.
\end{corobis}
\begin{proof}
$\Rightarrow$ If $L_p(\mathbf u,i)=\ell(u_i)$ is asymmetric, then for $w_m>w_n\ge0$, let $u_i=0$, $i\not=m,n$, then $w_m \ell(u_m)+w_n \ell(u_n)\ge w_m \ell(1)+w_n \ell(0)$ always holds, i.e.,
$$
w_m[(a-u_m)^p-(a-1)^p)]\ge w_n [a^p-(a-u_n)^p],
$$
so we have
$$
    \frac{w_m}{w_n}\ge \sup_{0\le u\le 1}\frac{a^p-(a-1+u)^p}{(a-u)^p-(a-1)^p}.
$$
RHS equals to $(\frac{a}{a-1})^{p-1}$ if $p>1$, and equals to $1$ when $p\le 1$.

$\Leftarrow$ According to Theorem \ref{lemma-strict}, $L$ is asymmetric
$$
\begin{aligned}
    &\Leftarrow w_m \ell(u_m)+w_i \ell(u_i) \ge w_m \ell(u_m+u_i) + w_i \ell(0)\\
    &\Leftrightarrow \frac{w_n}{w_i}\ge \sup_{\substack{u_i,u_m\ge0\\ u_i+u_m\le1}}\frac{\ell(0)-\ell(u_i)}{\ell(u_m)-\ell(u_n+u_i)}\\
    &\Leftrightarrow \frac{w_m}{w_i}\ge \sup_{\substack{u_i,u_m\ge0\\ u_i+u_m\le1}}\frac{a^p-(a-u_i)^p}{(a-u_m)^p-(a-u_i-u_m)^p}\\
    &\Leftrightarrow \frac{w_m}{w_i}\ge \mathbb{I}(p> 1)\cdot \sup_{0\le u_m\le 1}\left(\frac{a}{a-u_m}\right)^{p-1}+\mathbb{I}(p\le 1)\\
    &\Leftrightarrow \frac{w_m}{w_i}\ge \left(\frac{a}{a-1}\right)^{p-1}\cdot\mathbb{I}(p> 1)+\mathbb{I}(p\le1).
\end{aligned}
$$
On the other hand, if $\frac{w_m}{w_n}\ge (\frac{a+1}{a})^{1-q}\cdot\mathbb{I}(q\le 1)+\mathbb{I}(q>1)$. Then for any $i\not =m$, we have $\frac{w_m}{w_i}\ge (\frac{a}{a-1})^{p-1}\cdot\mathbb{I}(p> 1)+\mathbb{I}(p\le1)$.
\end{proof}

\begin{corobis}{3}
\label{a-AEL}
On the given weights $w_1,..,w_k$, where $w_m > w_n$ and $w_n=\max_{i\not = m}w_i$. The exponential loss function $L_{a}(\mathbf u, i)=\exp(-u_i/a)$ (where $a>0$) is asymmetric if and only if $\frac{w_m}{w_n}\ge  \exp(1/a)$.
\end{corobis}
\begin{proof}
$\Rightarrow$ If $L_a(\mathbf u,i)=\ell(u_i)$ is asymmetric, then for $w_m>w_n\ge0$, let $u_i=0$, $i\not=m,n$, then $w_m \ell(u_m)+w_n \ell(u_n)\ge w_m \ell(u_m+u_n)+ w_n \ell(0)$ always holds, i.e.,
$$
w_m[\exp(\frac{-u_m}{a})-\exp(\frac{-u_m-u_n}{a})]\ge w_n [1-\exp(\frac{-u_n}{a})],
$$
so we have
$$
    \frac{w_m}{w_n}\ge \exp\left(\frac{u_m}{a}\right)\Rightarrow a\ge \frac{1}{\ln w_m-\ln w_n}.
$$

$\Leftarrow$ According to Theorem \ref{lemma-strict}, $L_a$ is asymmetric
$$
\begin{aligned}
    &\Leftarrow w_m \ell(u_m)+w_i \ell(u_i) \ge w_m \ell(u_m+u_i) + w_i \ell(0)\\
    &\Leftrightarrow \frac{w_m}{w_i}\ge \exp\left(\frac{u_m}{a}\right).
\end{aligned}
$$
On the other hand, when $a\ge \frac{1}{\ln w_m-\ln w_n}$, then for any $i\not = m$, we have $\frac{w_m}{w_i}\ge \exp(1/a)$.
\end{proof}

\section{Experiments}
\subsection{Evaluation on Benchmark Datasets}
\textbf{Noise generation.} The noisy labels are generated following standard approaches in previous works \cite{ma2020normalized, 8099723}. For symmetric noise, we corrupt the training labels by flipping labels in each class randomly to incorrect labels to other classes with flip probability $\eta\in\{0.2, 0.3, 0.6, 0.8\}$. For asymmetric noise, we flip the labels within a specific set of classes. For MNIST, flipping 7 $\rightarrow$ 1, 2 $\rightarrow$ 7, 5 $\leftrightarrow$ 6, 3 $\rightarrow$ 8. For CIFAR-10, flipping TRUCK $\rightarrow$ AUTOMOBILE, BIRD $\rightarrow$ AIRPLANE, DEER $\rightarrow$ HORSE, CAR $\leftrightarrow$ DOG. For CIFAR-100, the 100 classes are grouped into 20 super-classes with each having 5 sub-classes, and each class are flipped  within the same super-class into the next in a circular fashion.

\textbf{Networks and training.} We follow the experimental settings in \cite{ma2020normalized}: 4-layer CNN for MNIST, an 8-layer CNN for CIFAR-10 and a ResNet-34 \cite{he2016deep} for CIFAR-100. The networks are trained for 50, 120, 200 epochs for MNIST, CIFAR-10, CIFAR-100, respectively. For all the training, we use SGD optimizer with momentum 0.9 and cosine learning rate annealing. Weight decay is set to $1\times 10^{-3}$, $1\times 10^{-4}$ and $1\times 10^{-5}$ for MNIST, CIFAR-10 and CIFAR-100, respectively. The initial learning rate is set to 0.01 for MNIST/CIFAR-10 and 0.1 for CIFAR-100. Batch size is set to 128. Typical data augmentations including random width/height shift and horizontal flip are applied.

\textbf{Parameter settings}.We set the parameter settings which match their original papers for all baseline methods. The details can be seen in Table \ref{ps}.

\begin{table}[htb]
    \centering
    \caption{Parameters settings for different methods.}
    \label{ps}
    \begin{tabular}{c|cccc}
    \hline
    Method & MNIST & CIFAR-10 & CIFAR-100 & WebVision\\
    \hline
    GCE\& NGCE ($q$) & (0.7) & (0.7) & (0.7) & (0.7)\\
    SCE ($A$, $\alpha$, $\beta$) & (-4, 0.01, 1.0) & (-4, 0.1, 1.0) & (-4, 6.0, 1.0) & (-4, 10.0, 1.0)\\
    FL\& NFL ($\gamma$) & (0.5) & (0.5) & (0.5) & -\\
    AGCE ($a$, $q$)& (4, 0.2) & (0.6, 0.6) & - & (1e-5,0.5)\\
    AUL ($a$, $p$)& (3, 0.1) & (5.5, 3) & - & -\\
    AEL ($a$) & (3.5) & (2.5) & - & -\\
    NFL+RCE ($A$, $\alpha$, $\beta$) & (-4, 1.0, 100.0) & (-4, 1.0, 1.0) & (-4, 10.0, 1.0) & -\\
    NCE+MAE ($\alpha$, $\beta$) & (1.0, 100.0) & (1.0, 1.0) & (10.0, 1.0) & -\\
    NCE+RCE ($\alpha$, $\beta$) & (1.0, 100.0) & (1.0, 1.0) & (10.0, 1.0) & (50.0, 0.1)\\
    NCE+AGCE ($a$, $q$, $\alpha$, $\beta$) & (4, 0.2, 0, 1) & (6, 1.5, 1, 4) & (1.8, 3, 10, 0.1) & (2.5, 3, 50, 0.1)\\
    NCE+AUL ($a$, $p$, $\alpha$, $\beta$)& (3, 0.1, 0, 1) & (6.3, 1.5, 1, 4) & (6, 3, 10, 0.015) & -\\
    NCE+AEL ($a$, $\alpha$, $\beta$) & (3.5, 0, 1) & (5, 1, 4) & (1.5, 10, 0.1) & -\\
    \hline
    \end{tabular}
\end{table}

\begin{figure}[htb]
    \centering
    \subfigure[GCE with $\eta=0.0$]{
    \label{ffig:GCE-mnist-s-0.0}
    \includegraphics[width=1.2in]{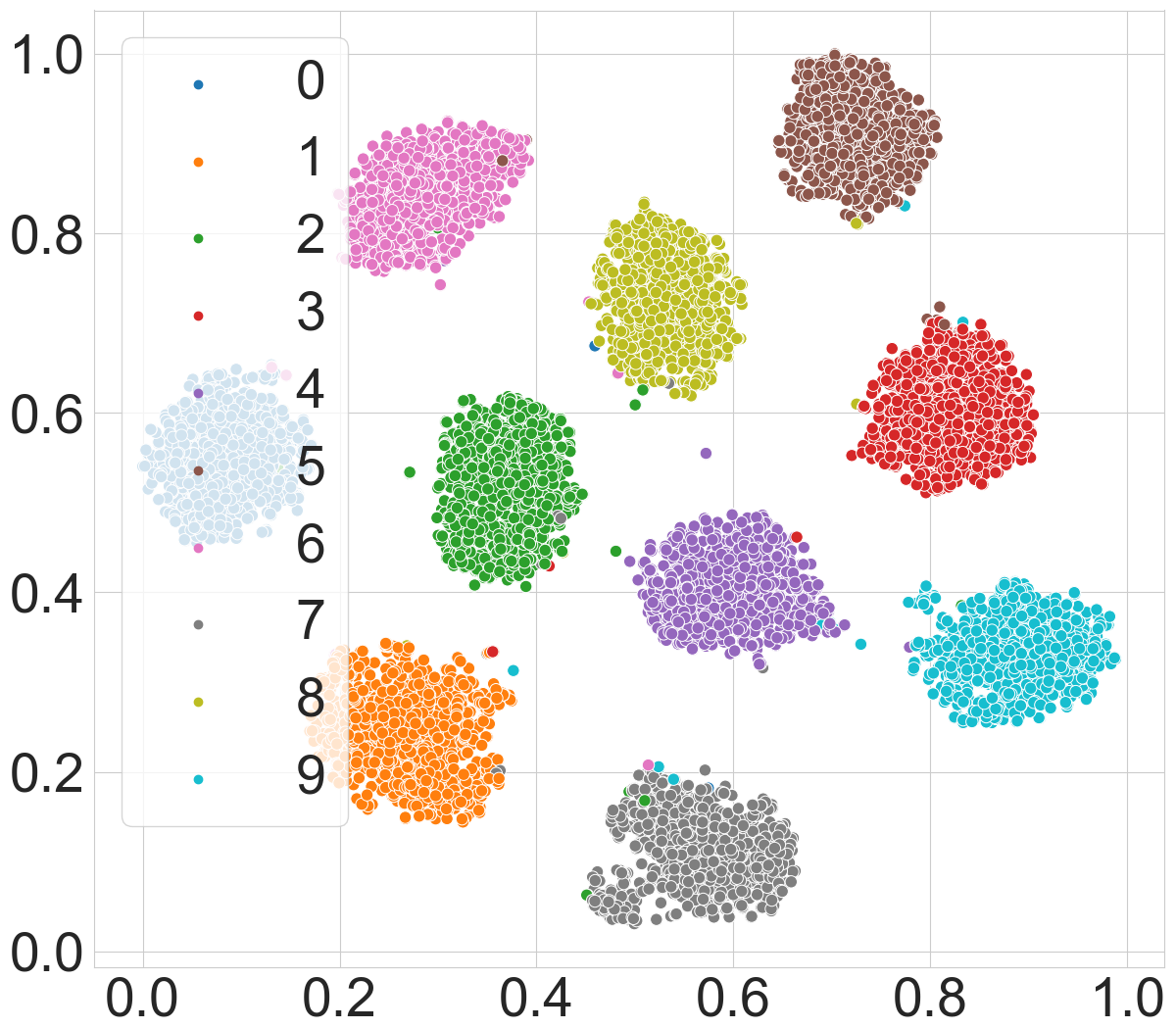}
    }
    \subfigure[GCE with $\eta=0.2$]{
    \label{ffig:GCE-mnist-s-0.2}
    \includegraphics[width=1.2in]{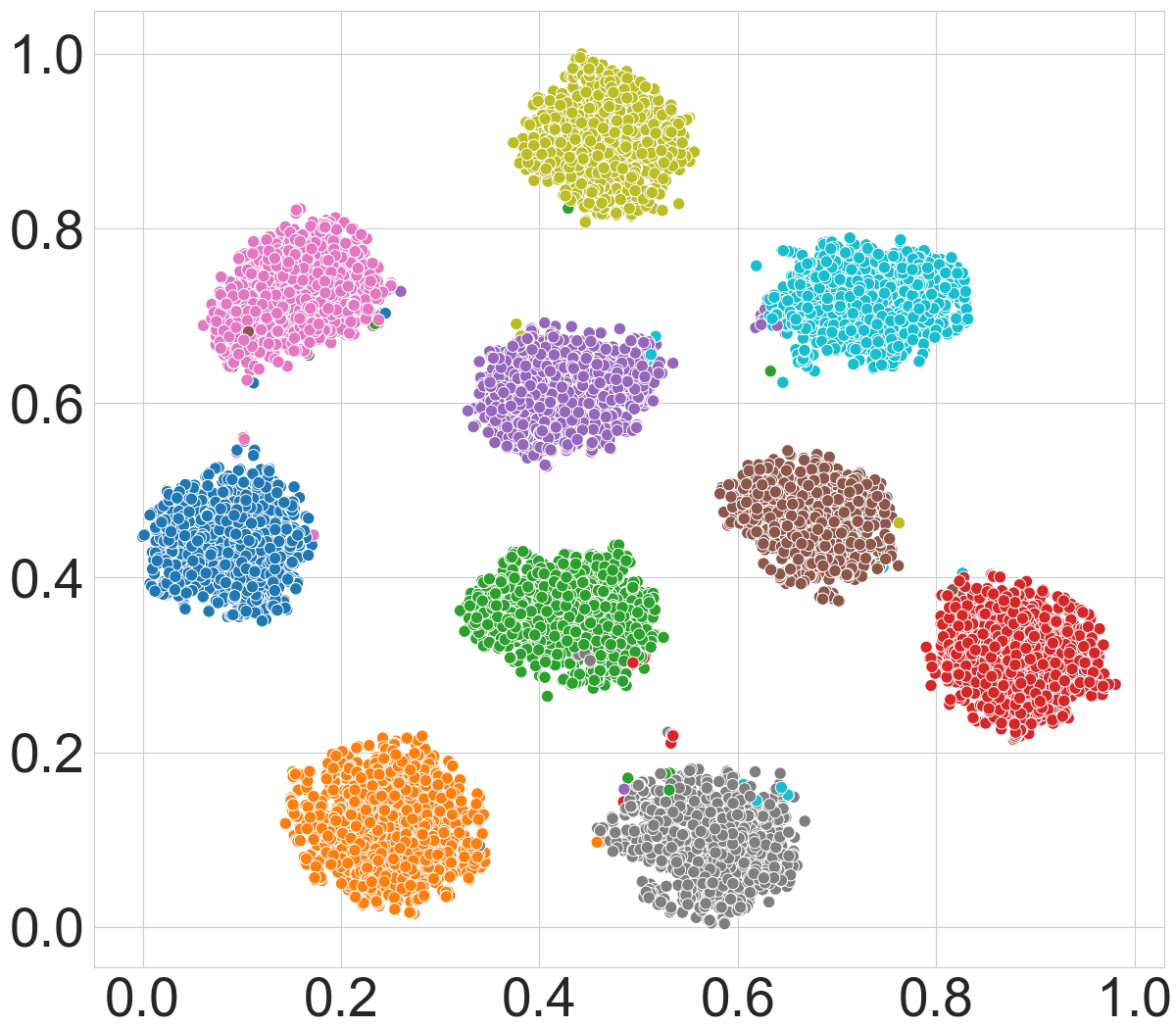}
    }
    \subfigure[GCE with $\eta=0.4$]{
    \label{ffig:GCE-mnist-s-0.4}
    \includegraphics[width=1.2in]{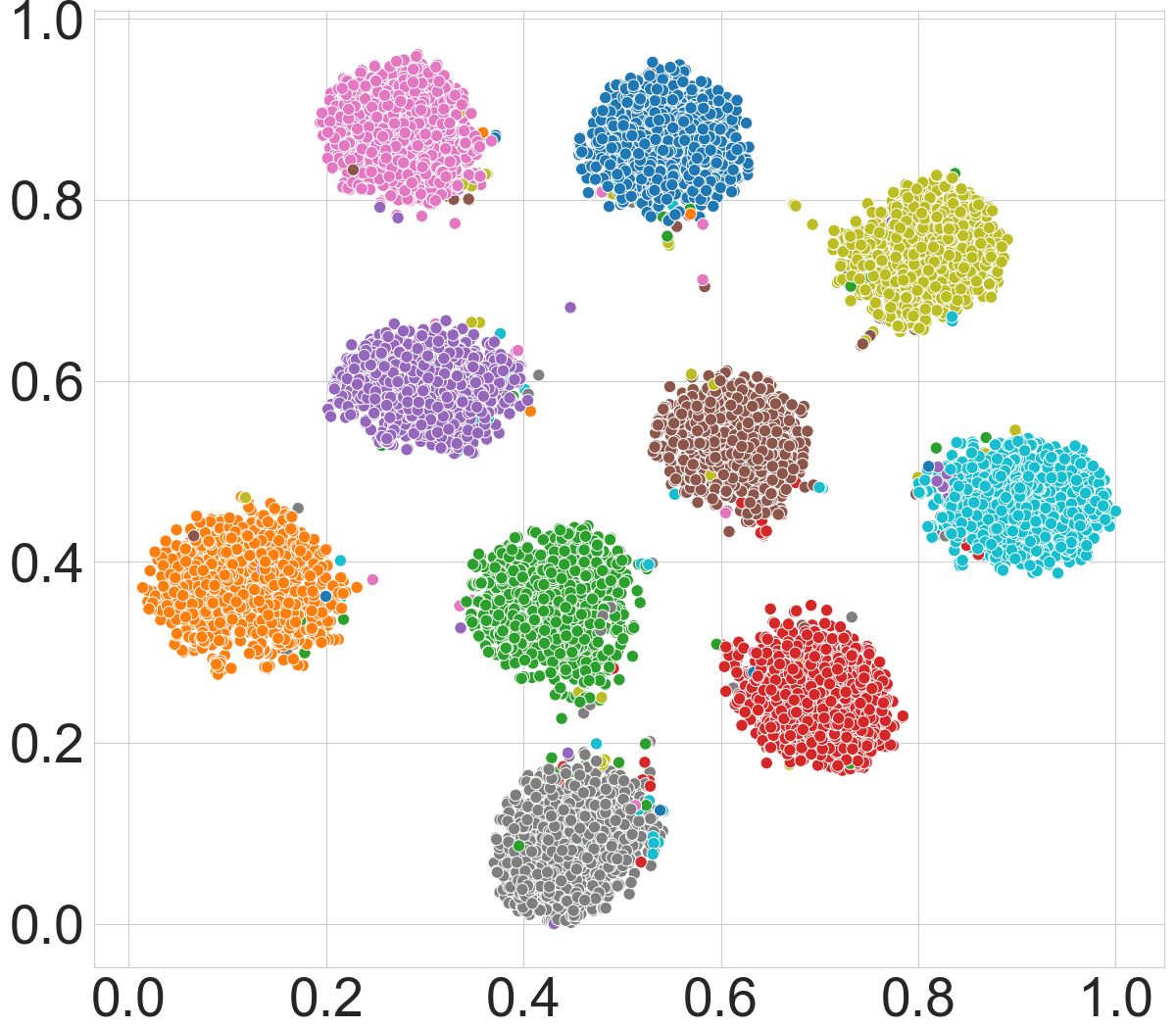}
    }
    \subfigure[GCE with $\eta=0.6$]{
    \label{ffig:GCE-mnist-s-0.6}
    \includegraphics[width=1.2in]{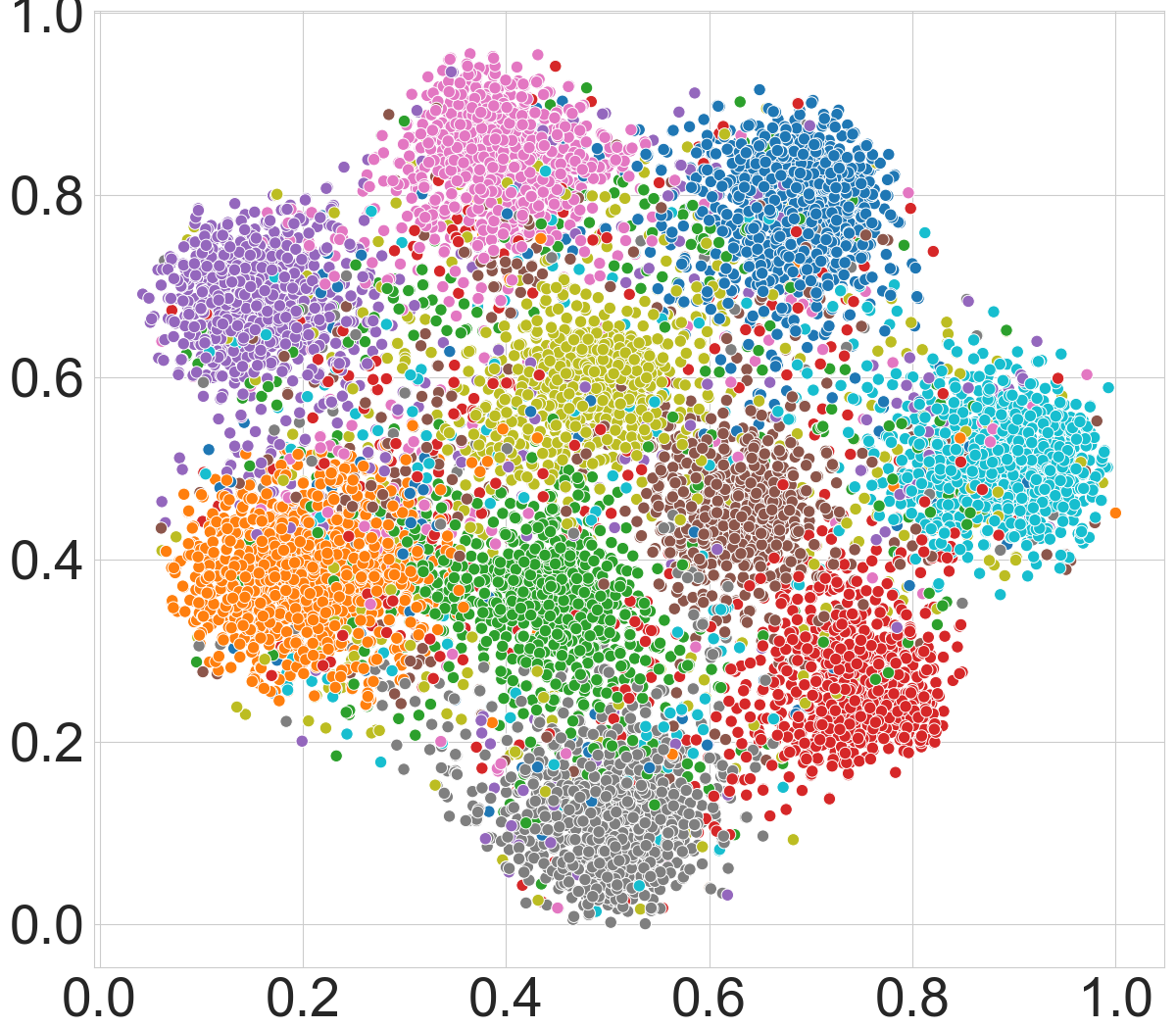}
    }
    \subfigure[GCE with $\eta=0.8$]{
    \label{ffig:GCE-mnist-s-0.8}
    \includegraphics[width=1.2in]{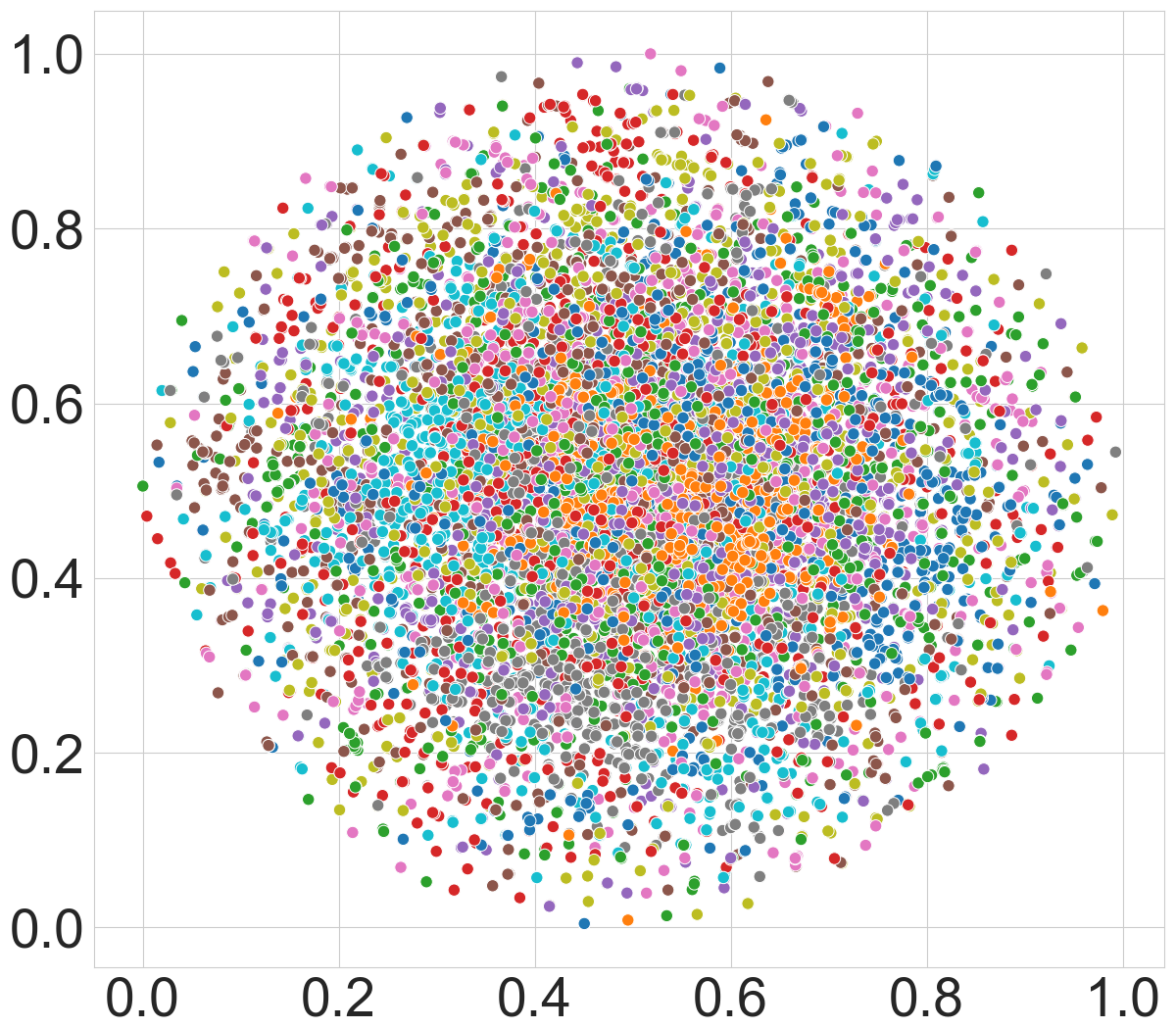}
    }
    \subfigure[AGCE with $\eta=0.0$]{
    \label{ffig:AGCE-mnist-s-0.0}
    \includegraphics[width=1.2in]{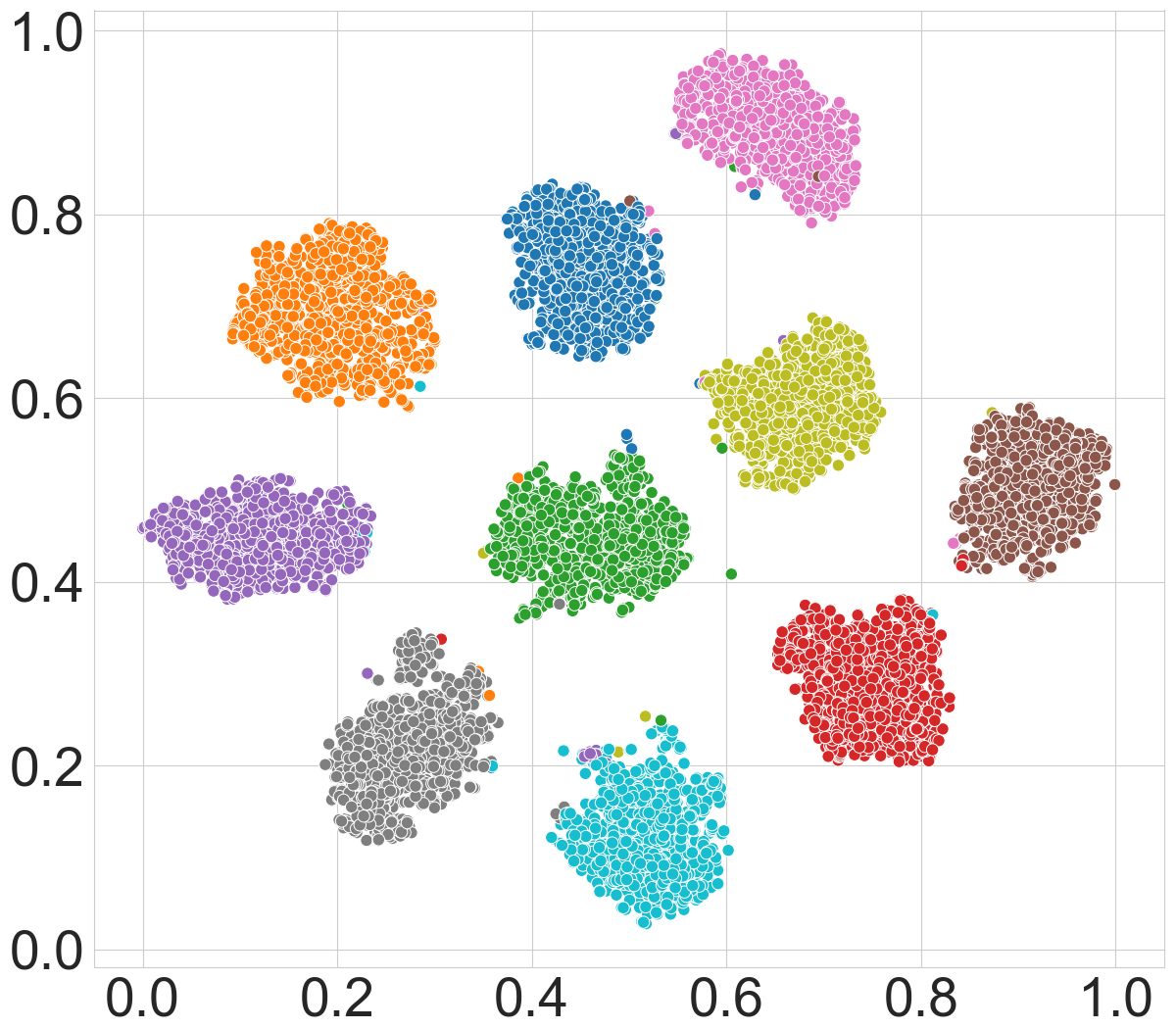}
    }
    \subfigure[AGCE with $\eta=0.2$]{
    \label{ffig:AGCE-mnist-s-0.2}
    \includegraphics[width=1.2in]{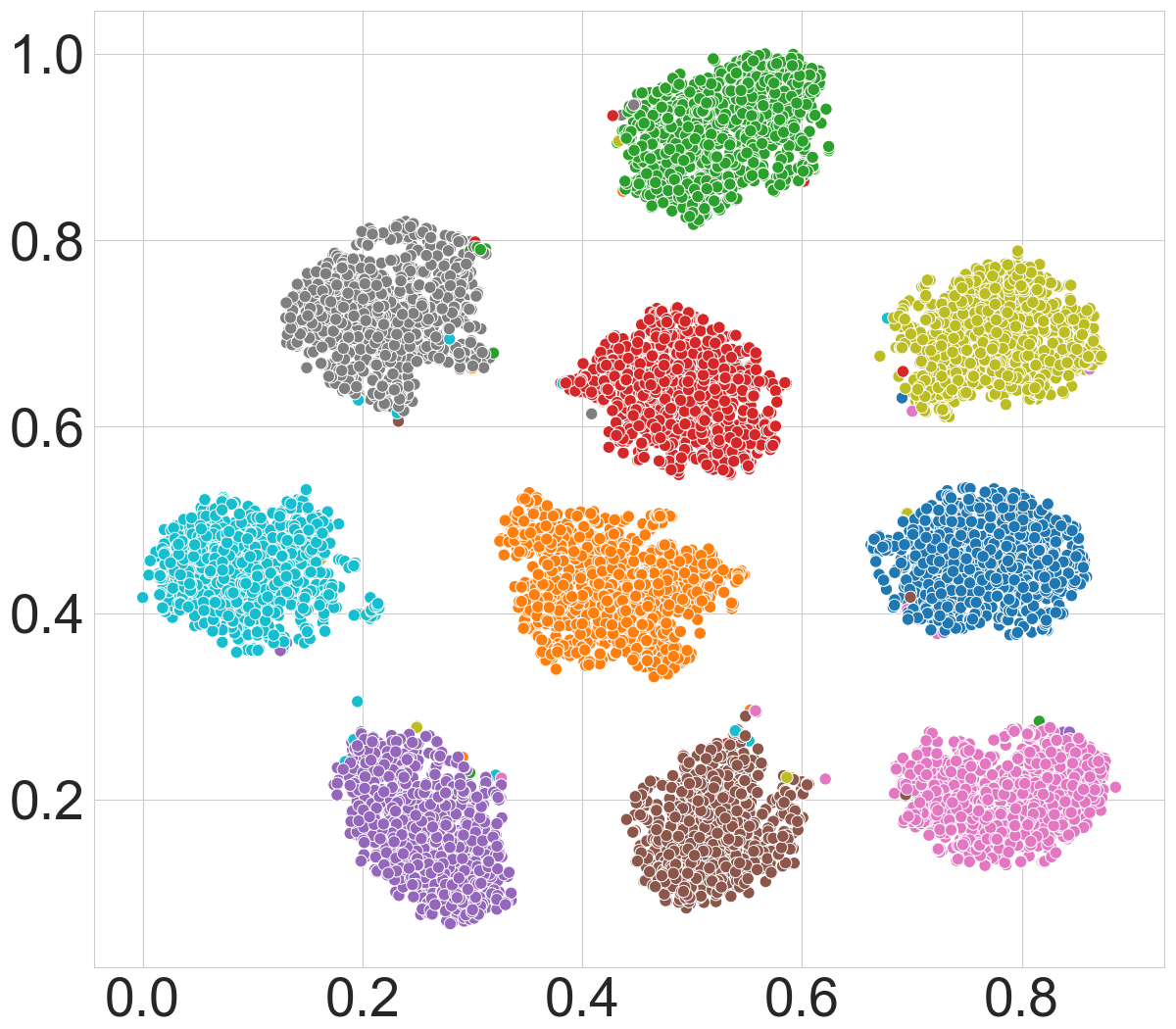}
    }
    \subfigure[AGCE with $\eta=0.4$]{
    \label{ffig:AGCE-mnist-s-0.4}
    \includegraphics[width=1.2in]{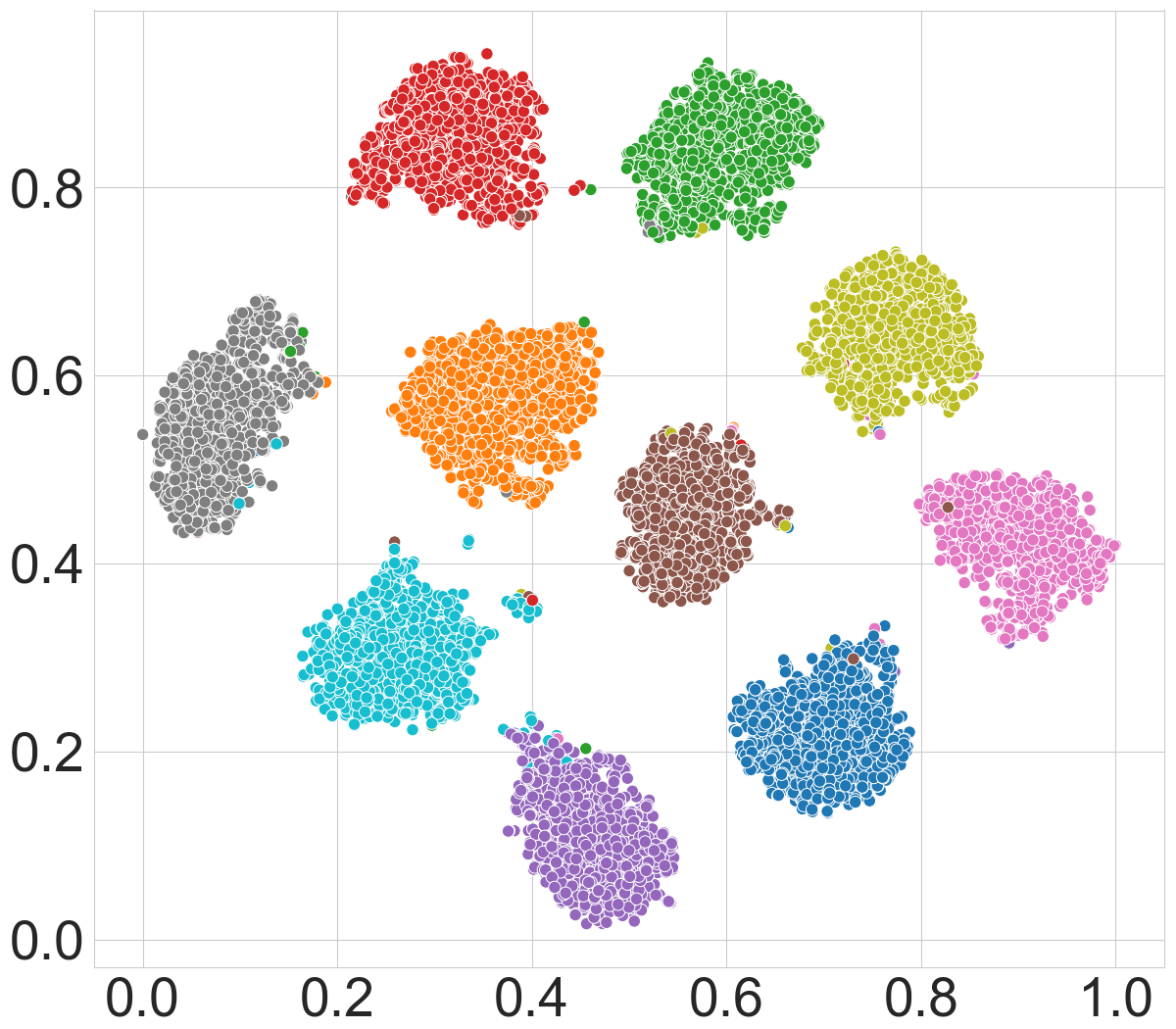}
    }
    \subfigure[AGCE with $\eta=0.6$]{
    \label{ffig:AGCE-mnist-s-0.6}
    \includegraphics[width=1.2in]{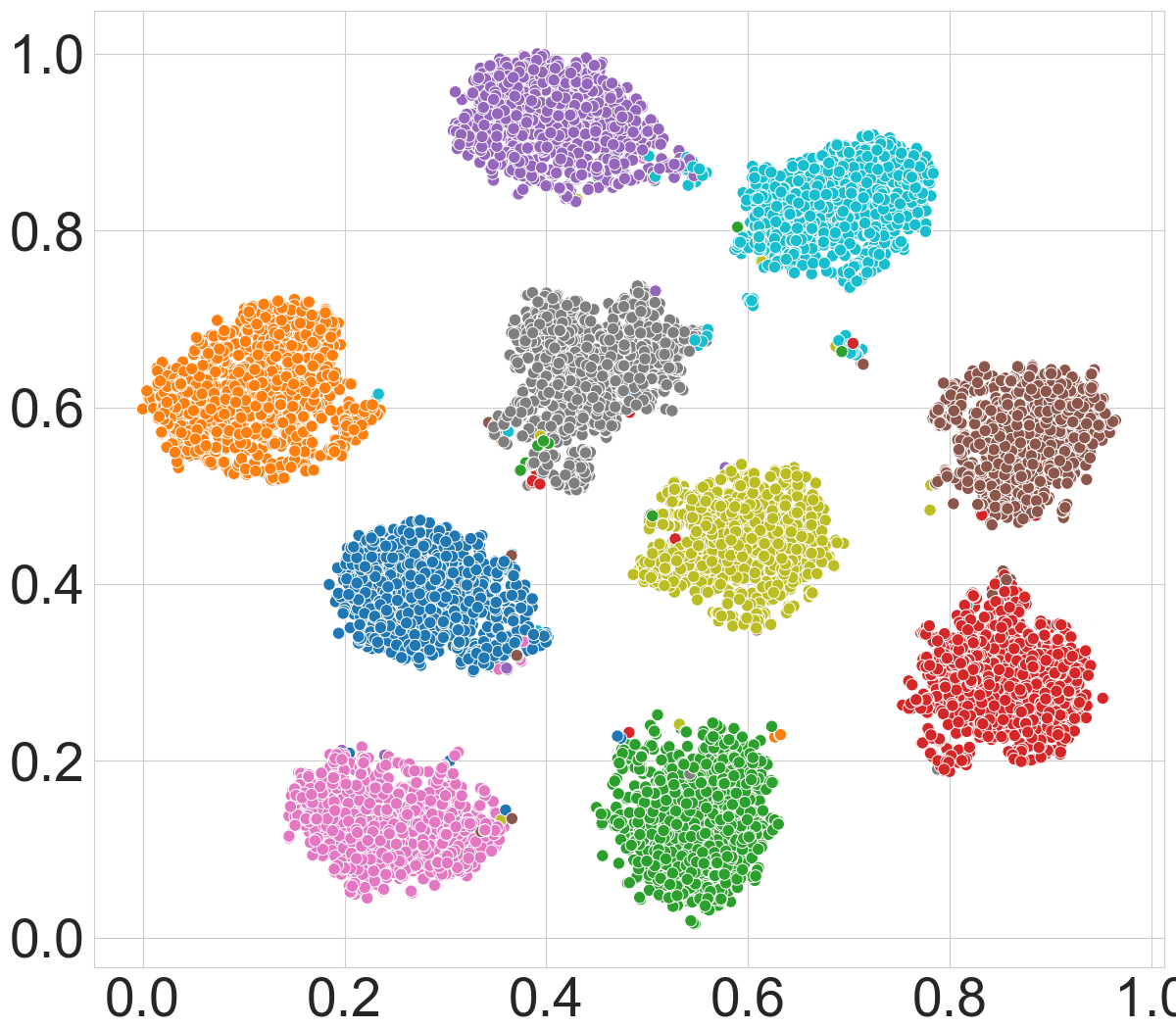}
    }
    \subfigure[AGCE with $\eta=0.8$]{
    \label{ffig:AGCE-mnist-s-0.8}
    \includegraphics[width=1.2in]{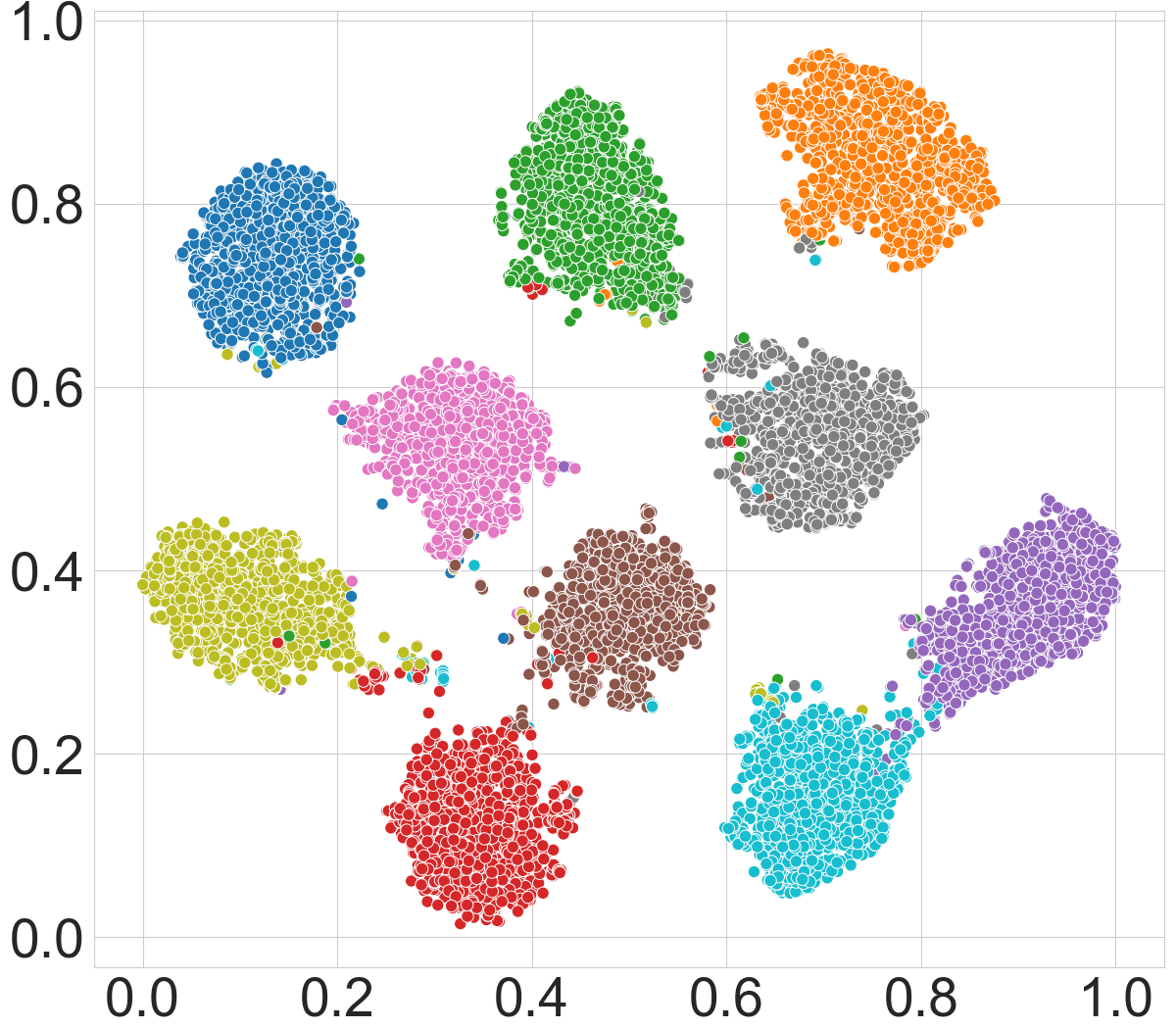}
    }
    \caption{Visualization for GCE (top) and AGCE (bottom) on MNIST with different symmetric noise ($\eta\in[0.0, 0.2, 0.4, 0.6, 0.8]$) by t-SNE \cite{tsne} 2D embeddings of deep features.}
    \label{ffig:tsne-mnist-s}
\end{figure}

\begin{figure}[htb]
    \centering
    \subfigure[CE]{
    \label{ffig:CE-cifar10-0.0}
    \includegraphics[width=1.2in]{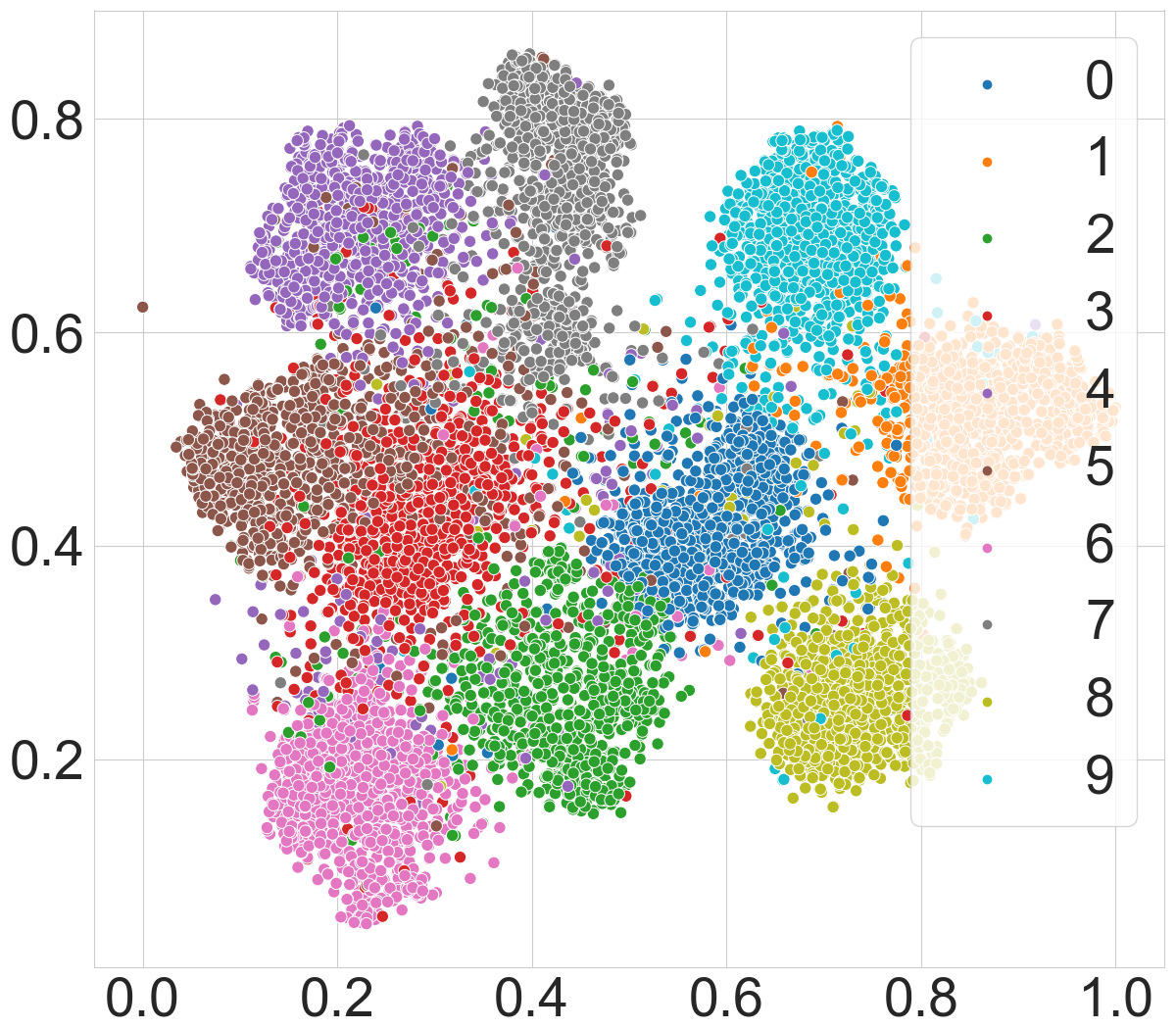}
    }
    \subfigure[GCE]{
    \label{ffig:GCE-cifar10-0.0}
    \includegraphics[width=1.2in]{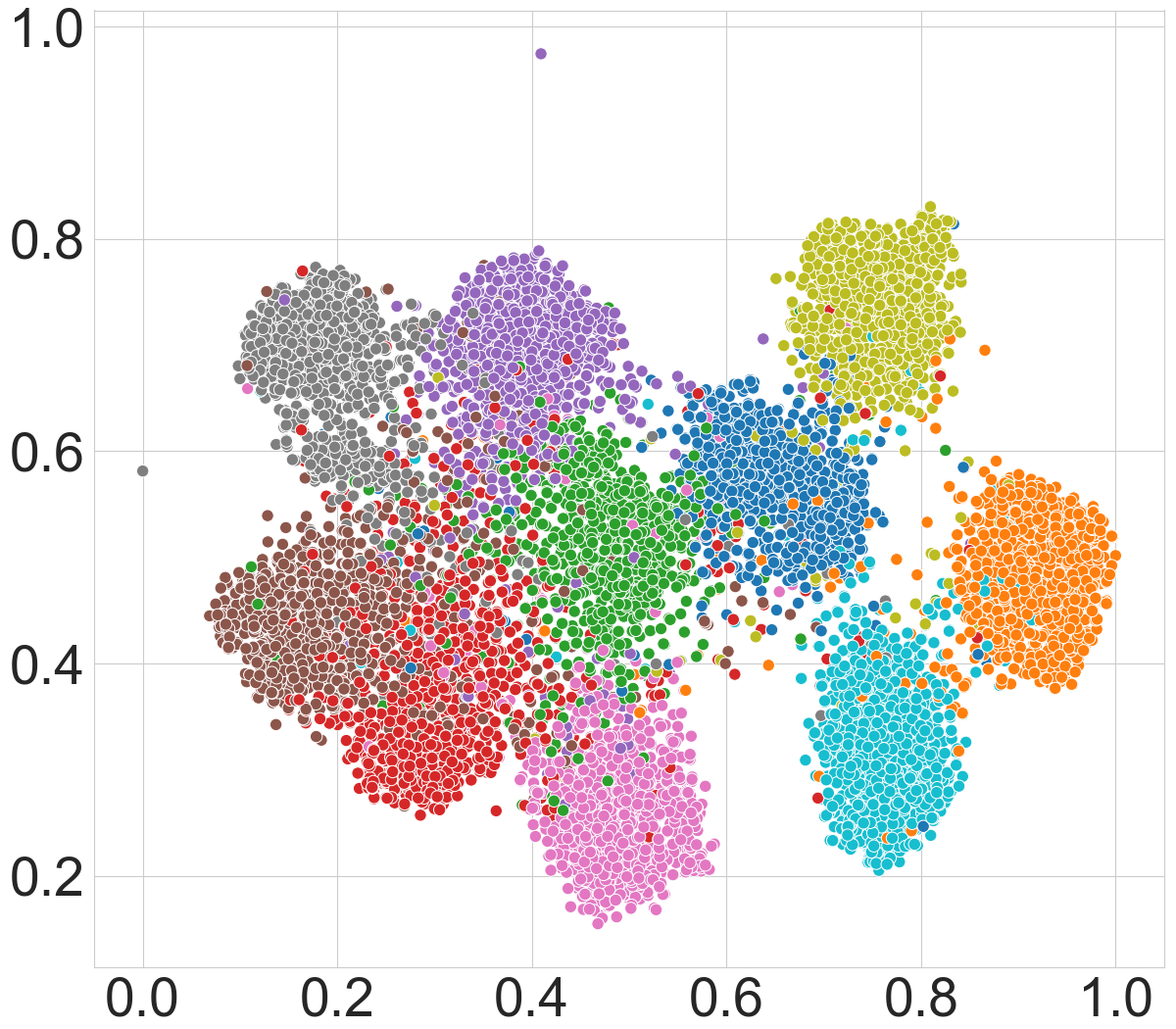}
    }
    \subfigure[AGCE]{
    \label{ffig:AGCE-cifar10-0.0}
    \includegraphics[width=1.2in]{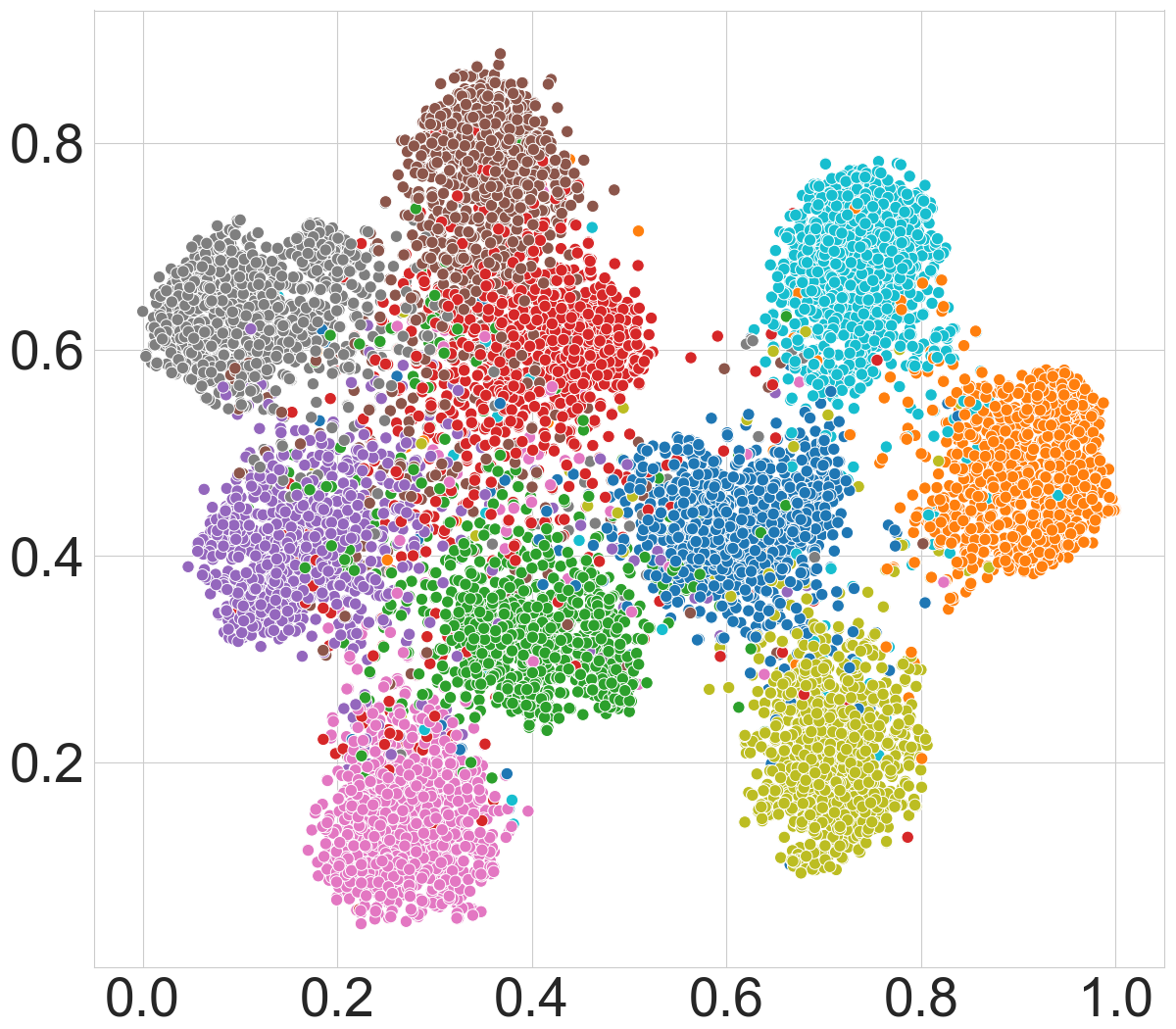}
    }
    \subfigure[AUL]{
    \label{ffig:AUL-cifar10-0.0}
    \includegraphics[width=1.2in]{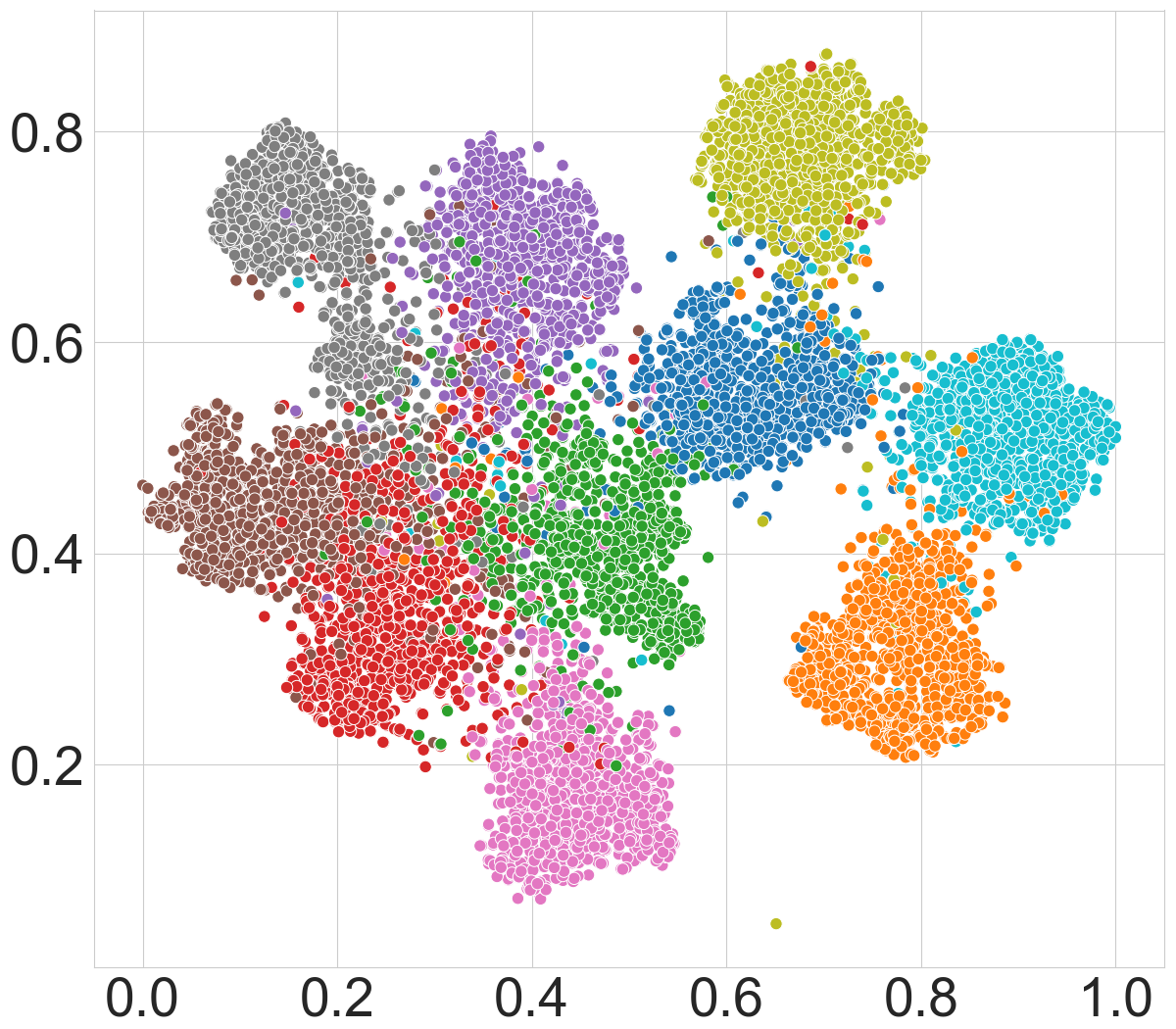}
    }
    \subfigure[AEL]{
    \label{ffig:AEL-cifar10-0.0}
    \includegraphics[width=1.2in]{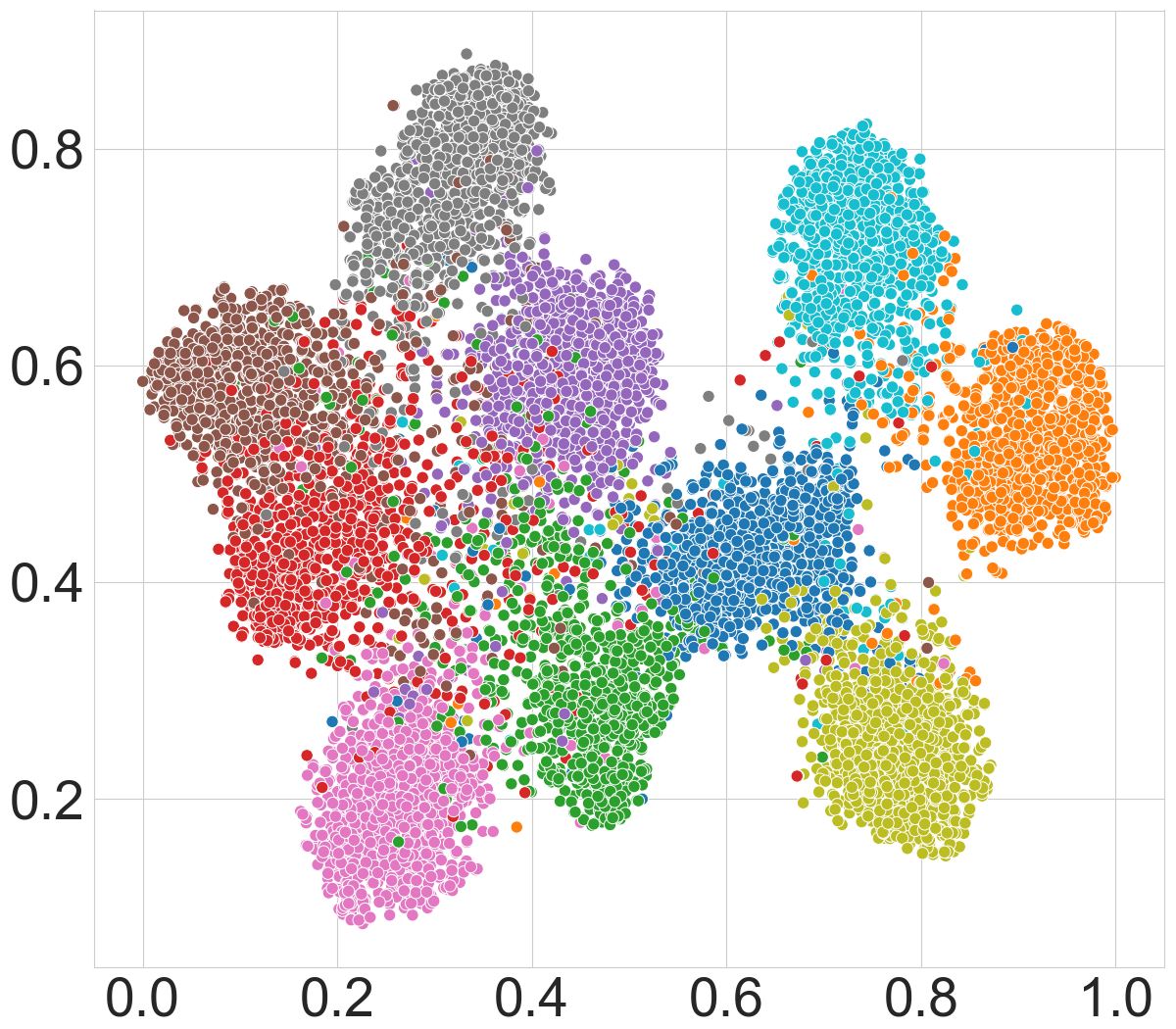}
    }
    \subfigure[CE]{
    \label{ffig:CE-cifar10-0.4}
    \includegraphics[width=1.2in]{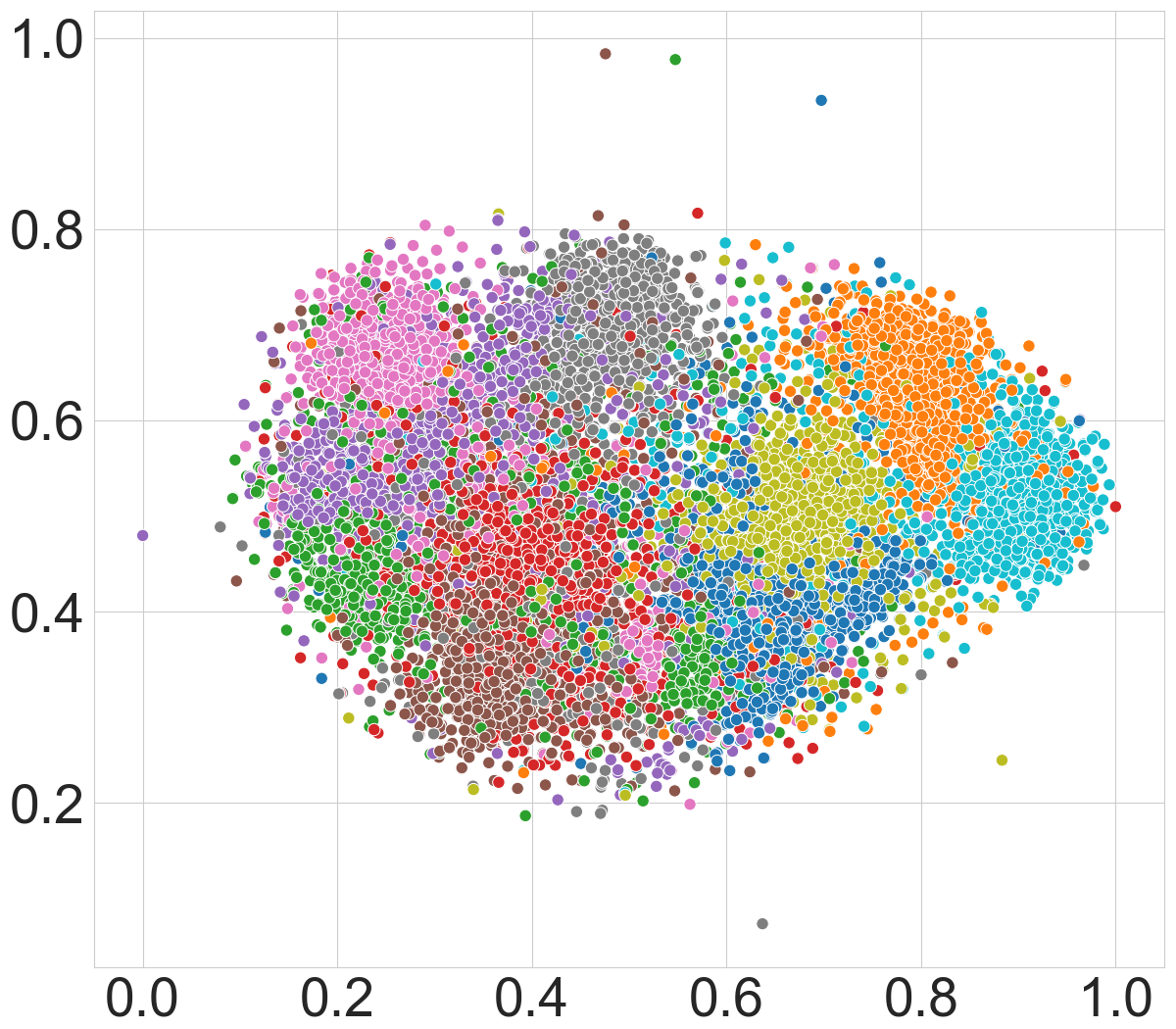}
    }
    \subfigure[GCE]{
    \label{ffig:GCE-cifar10-0.4}
    \includegraphics[width=1.2in]{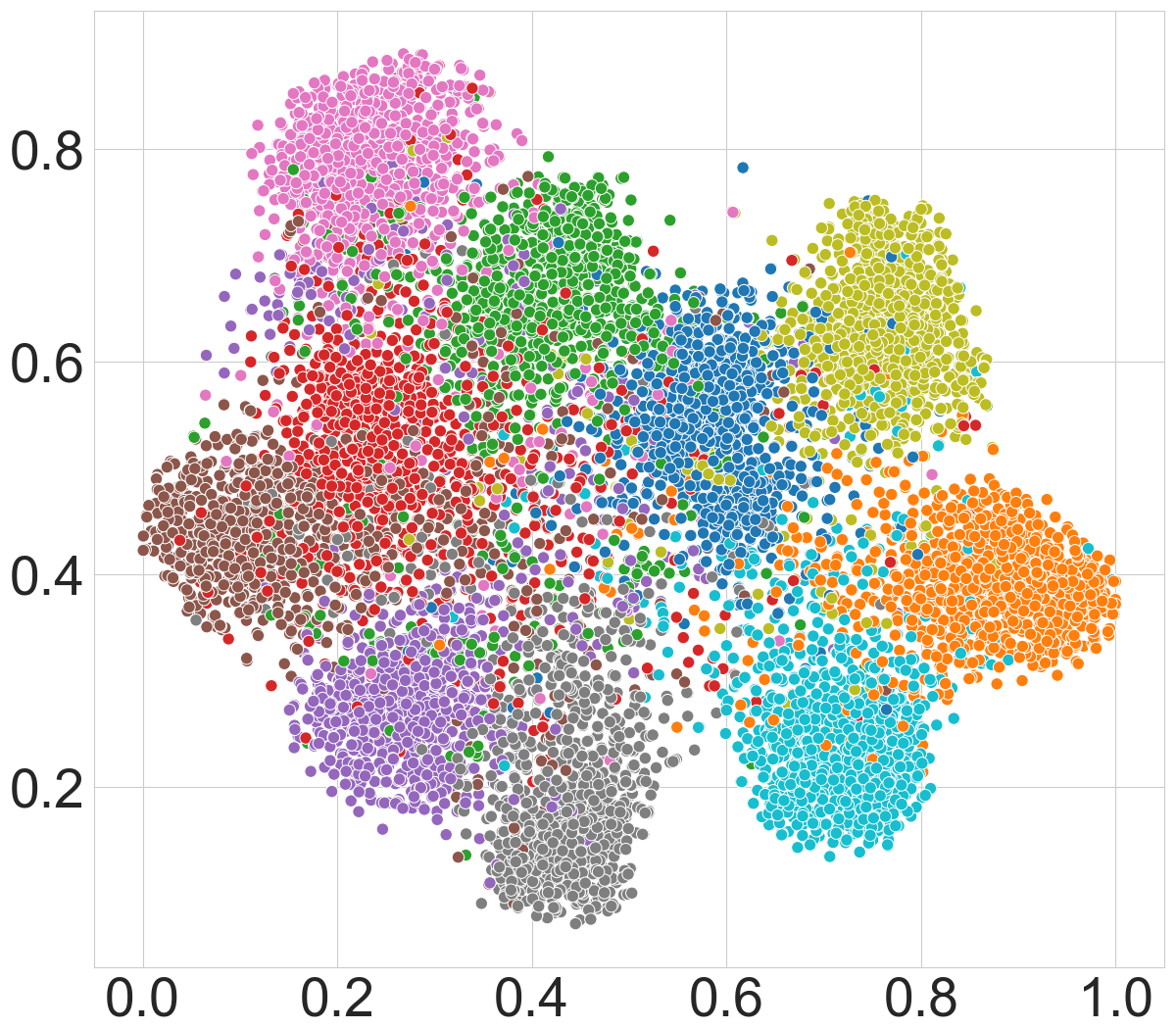}
    }
    \subfigure[AGCE]{
    \label{ffig:AGCE-cifar10-0.4}
    \includegraphics[width=1.2in]{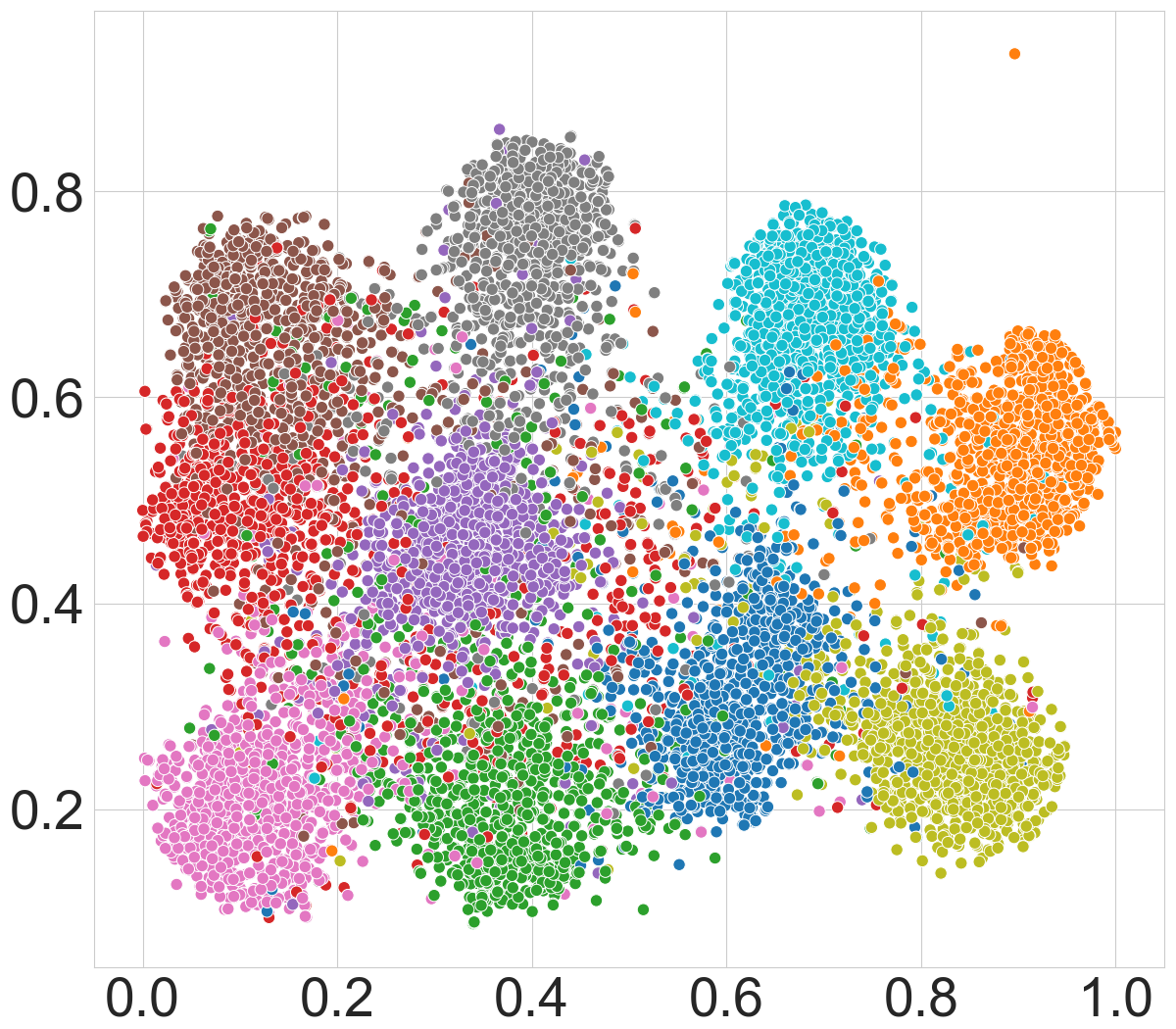}
    }
    \subfigure[AUL]{
    \label{ffig:AUL-cifar10-0.4}
    \includegraphics[width=1.2in]{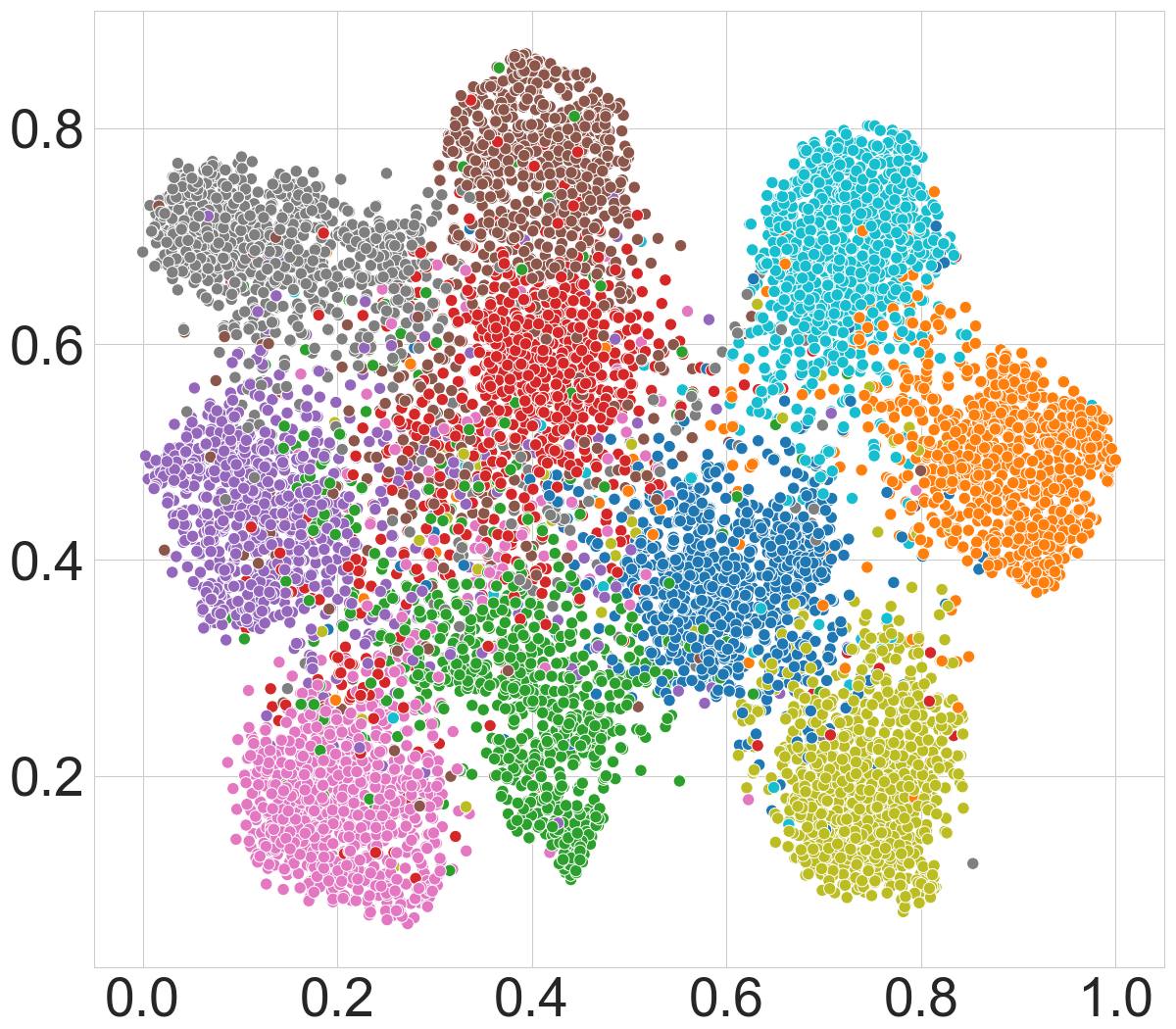}
    }
    \subfigure[AEL]{
    \label{ffig:AEL-cifar10-0.4}
    \includegraphics[width=1.2in]{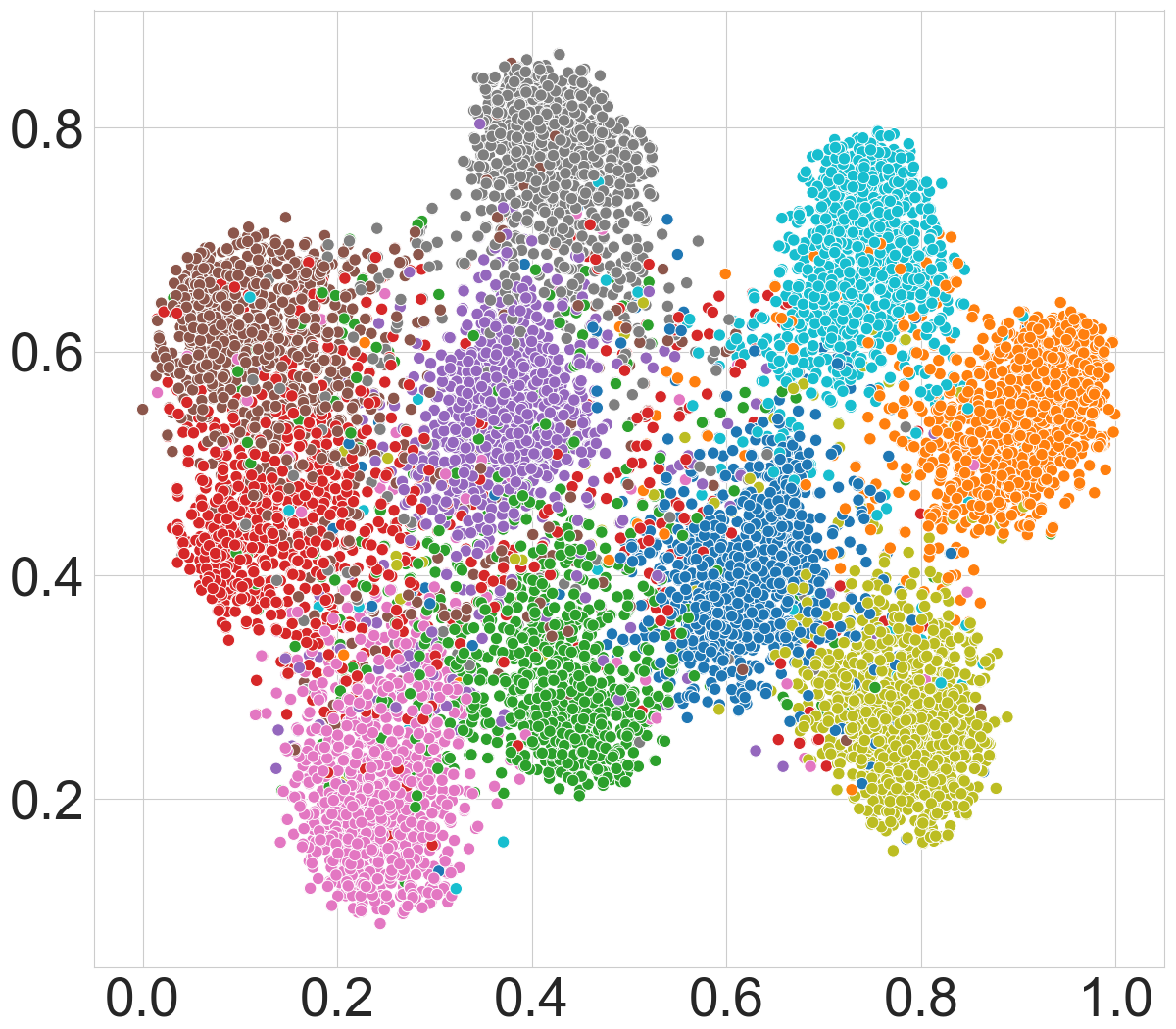}
    }
    \caption{Visualization for CE, GCE, AGCE, AUL, and AEL on CIFAR10 with different symmetric noise (0.0 for top, 0.4 for bottom) by t-SNE \cite{tsne} 2D embeddings of deep features.}
    \label{ffig:tsne-cifar10}
\end{figure}

\textbf{Results.} The experimental results of symmetric and asymmetric label noise are shown in Table \ref{full-symmetric-noise} and Table \ref{full-asymmetric-noise}, respectively. And we also visualize the learned features by the AGCE loss function and the GCE loss function. Figure \ref{ffig:tsne-mnist-s} validates AGCE's ability of separating samples and robustness to label noise with any noise rate $\eta\in\{0.0, 0.2, 0.4, 0.6, 0.8\}$. Figure \ref{ffig:tsne-cifar10} validates the different loss functions of separating samples and robustness to symmetric label noise with noise rates 0.0 and 0.4.

\subsection{Evaluation on Real-world Noisy Labels}
Here, we evaluate our asymmetric loss functions on large-scale real-world noisy dataset WebVision 1.0 \cite{li2017webvision}. It contains 2.4 million images of real-world noisy labels, crawled from the web using 1,000 concepts in ImageNet ILSVRC12. Since the dataset is very big, for quick experiments, we follow the training setting in \cite{jiang2018mentornet, ma2020normalized} that only takes the first 50 classes of the Google resized image subset. We evaluate the trained networks on the same 50 classes of WebVision 1.0 validation set, which can be considered as a clean validation. ResNet-50 \cite{he2016deep} is the model to be learnt. We compare our NCE+AGCE with GCE, SCE and NCE+RCE. The training details follow \cite{ma2020normalized}, where for each loss, we train a ResNet-50 \cite{he2016deep} using SGD for 250 epochs with initial learning rate 0.4, nesterov momentum 0.9 and weight decay $3\times 10^{-5}$ and batch size $512$. The learning rate is multiplied by $0.97$ after every epoch of training. All the images are resized to $224\times 224$. Typical data augmentations including random width/height shift, color jittering, and random horizontal flip are applied. Experiments can be reported in Table \ref{a-webvision}. 

\begin{table}[htb]
\centering
\caption{Top-1 validation accuracies (\%) on WebVision validation set of ResNet-50 models trained on WebVision using different loss functions, under the Mini setting in \cite{jiang2018mentornet, ma2020normalized}.}
\vskip2pt

\label{a-webvision}
\begin{tabular}{c|cccccc}
    \hline
    Loss & CE & GCE & SCE & NCE+RCE & \textbf{NCE+AGCE} & \textbf{AGCE}\\
    \hline
    Acc & 66.96 & 61.76 & 66.92 & 66.32 & \textbf{67.12} & \textbf{69.40}\\ 
    \hline

\end{tabular}
\end{table}

\begin{figure}[htb]
    \centering
    \subfigure[GCE with $\eta=0.1$]{
    \label{ffig:GCE-mnist-a-0.1}
    \includegraphics[width=1.2in]{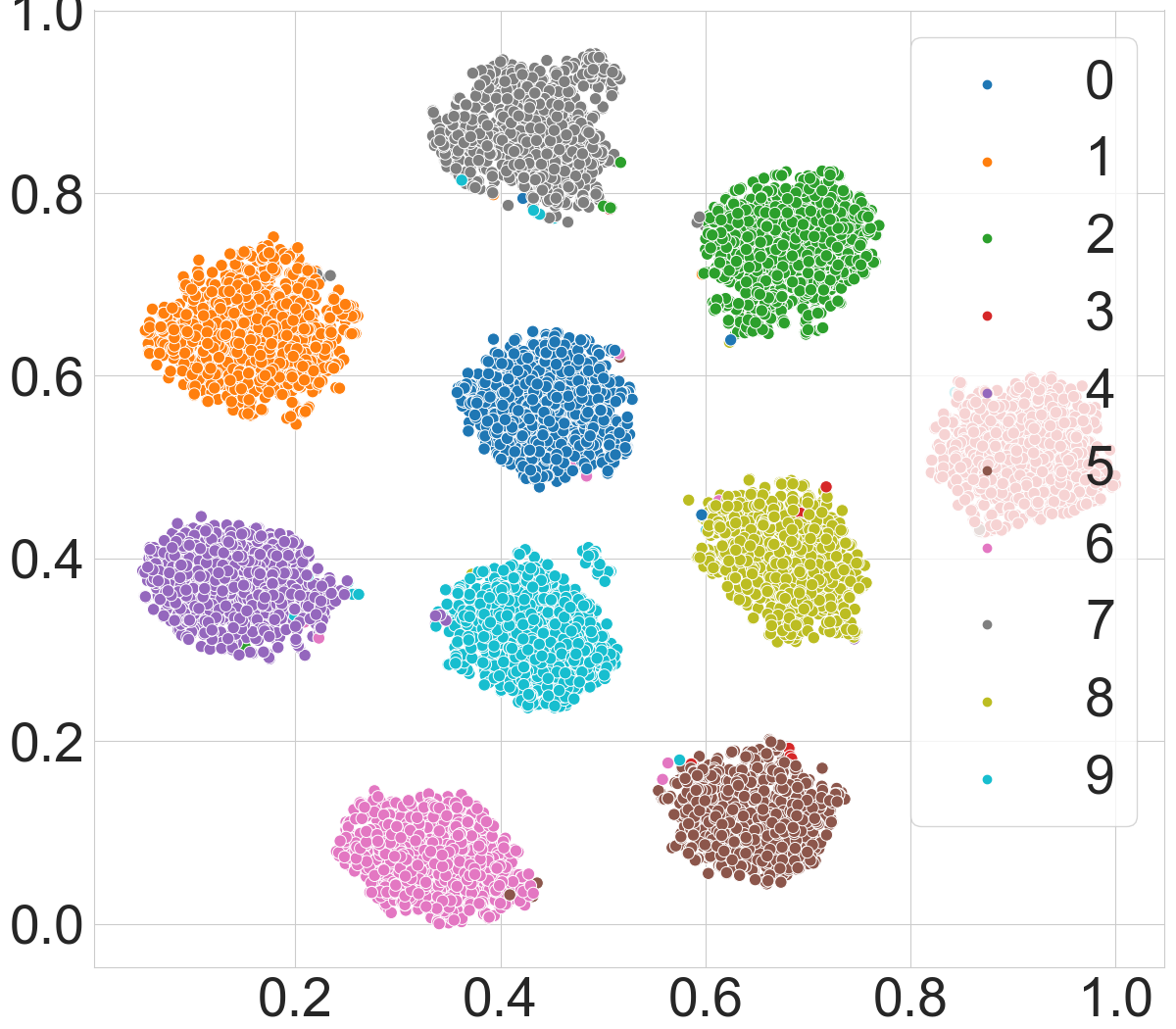}
    }
    \subfigure[GCE with $\eta=0.2$]{
    \label{ffig:GCEE-mnist-a-0.2}
    \includegraphics[width=1.2in]{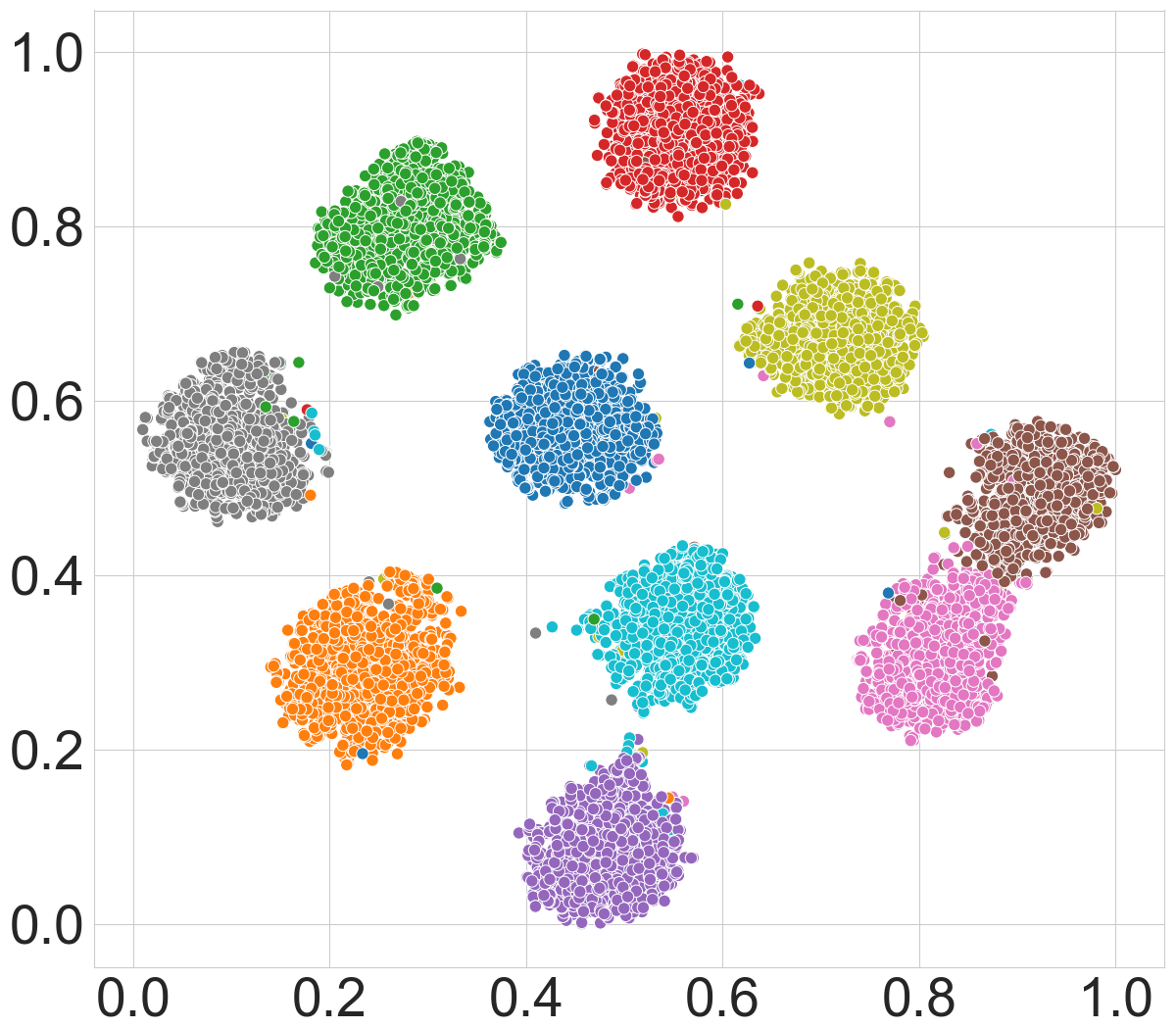}
    }
    \subfigure[GCE with $\eta=0.3$]{
    \label{ffig:GCEE-mnist-a-0.3}
    \includegraphics[width=1.2in]{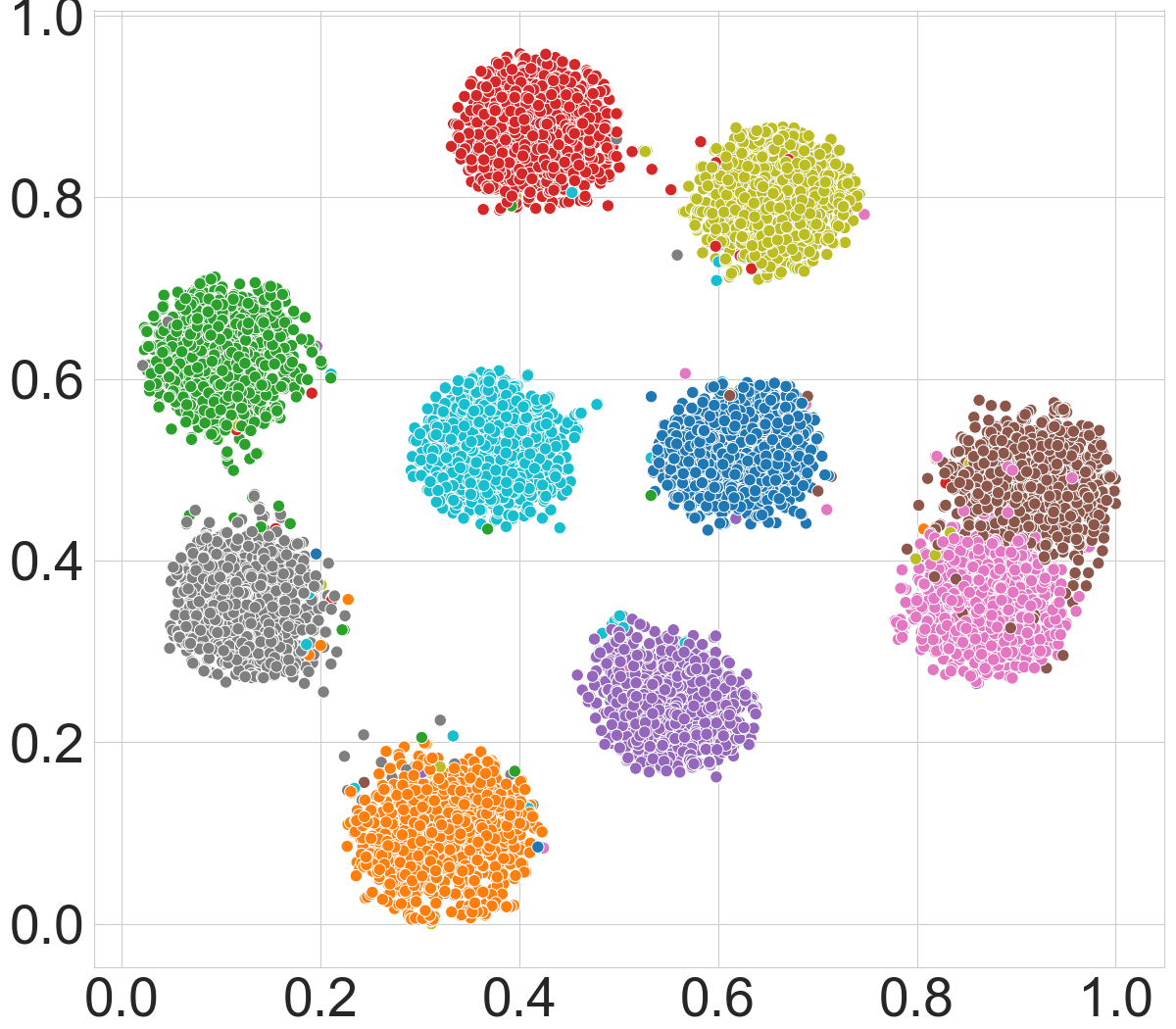}
    }
    \subfigure[GCE with $\eta=0.4$]{
    \label{ffig:GCEE-mnist-a-0.4}
    \includegraphics[width=1.2in]{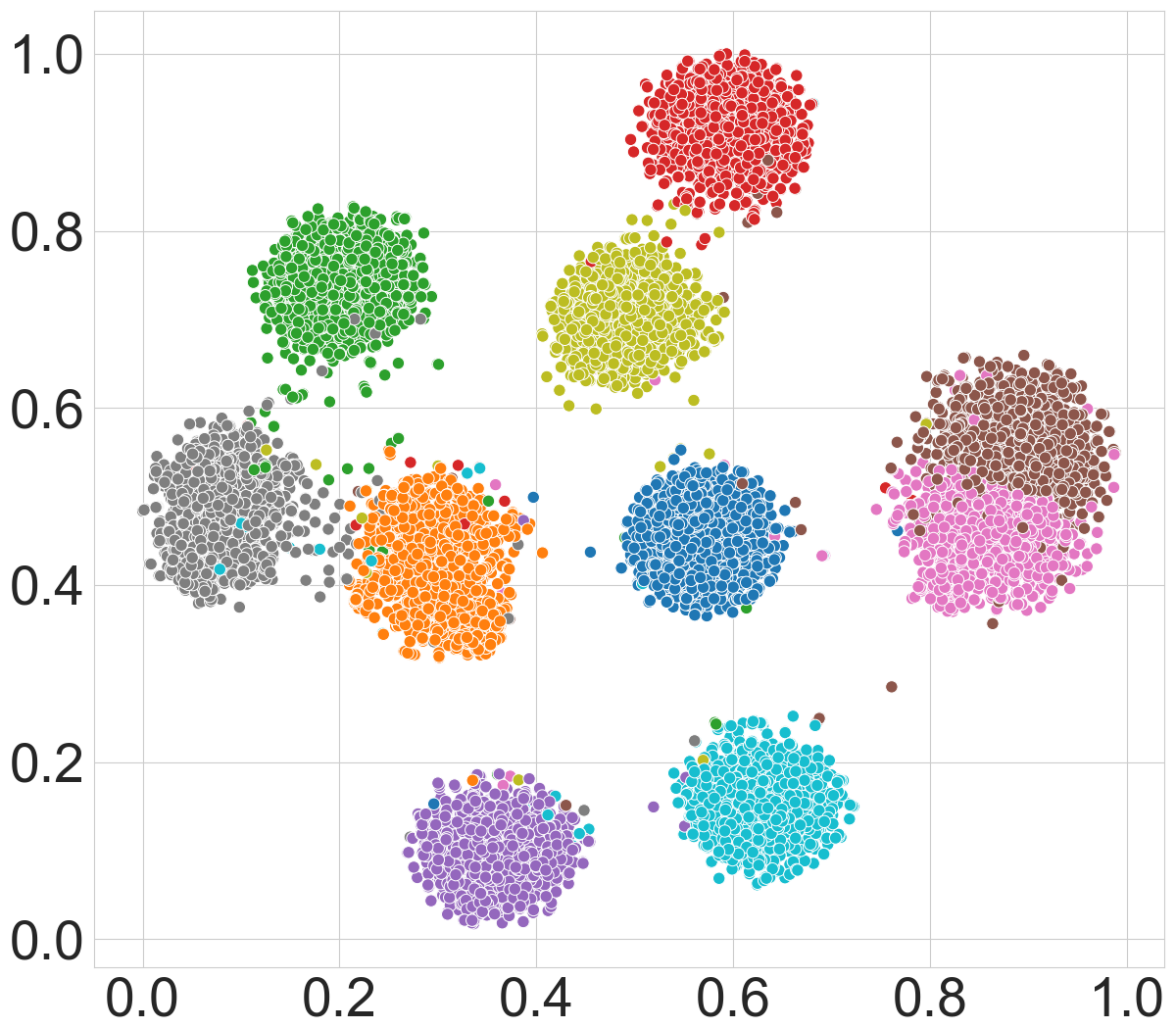}
    }
    \\
    \subfigure[AGCE with $\eta=0.1$]{
    \label{ffig:AGCEE-mnist-a-0.0}
    \includegraphics[width=1.2in]{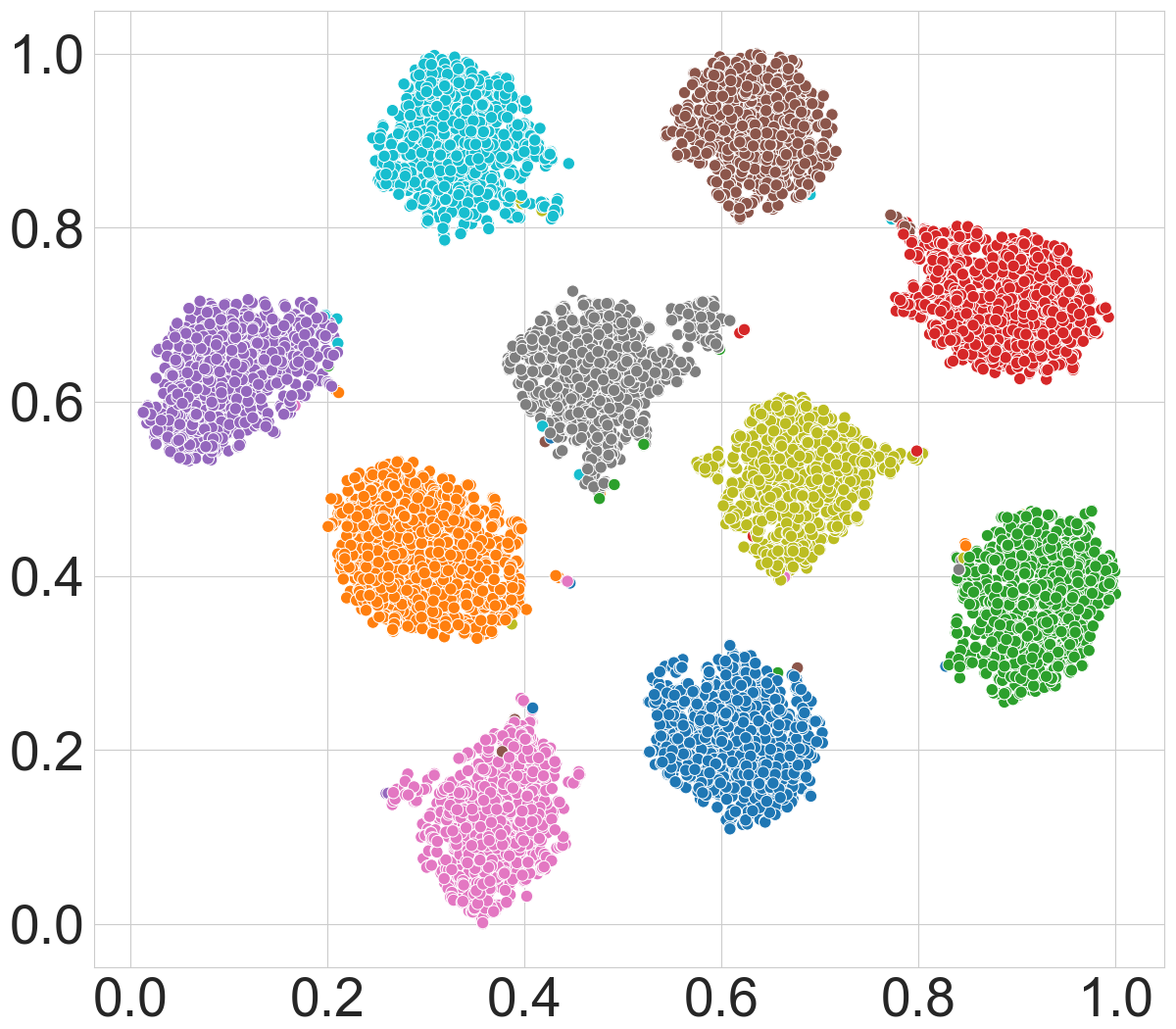}
    }
    \subfigure[AGCE with $\eta=0.2$]{
    \label{ffig:AGCEE-mnist-a-0.1}
    \includegraphics[width=1.2in]{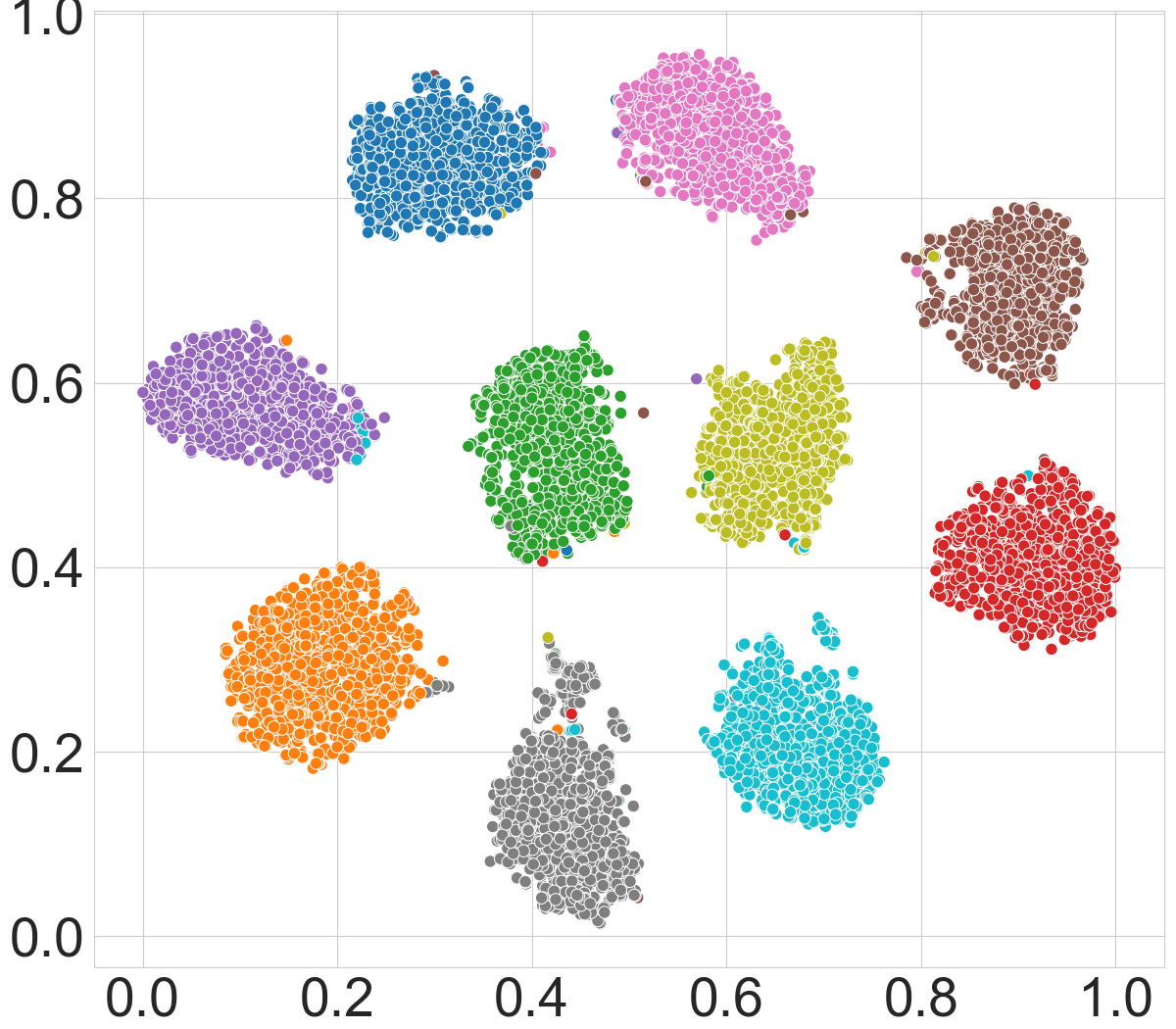}
    }
    \subfigure[AGCE with $\eta=0.3$]{
    \label{ffig:AGCEE-mnist-a-0.3}
    \includegraphics[width=1.2in]{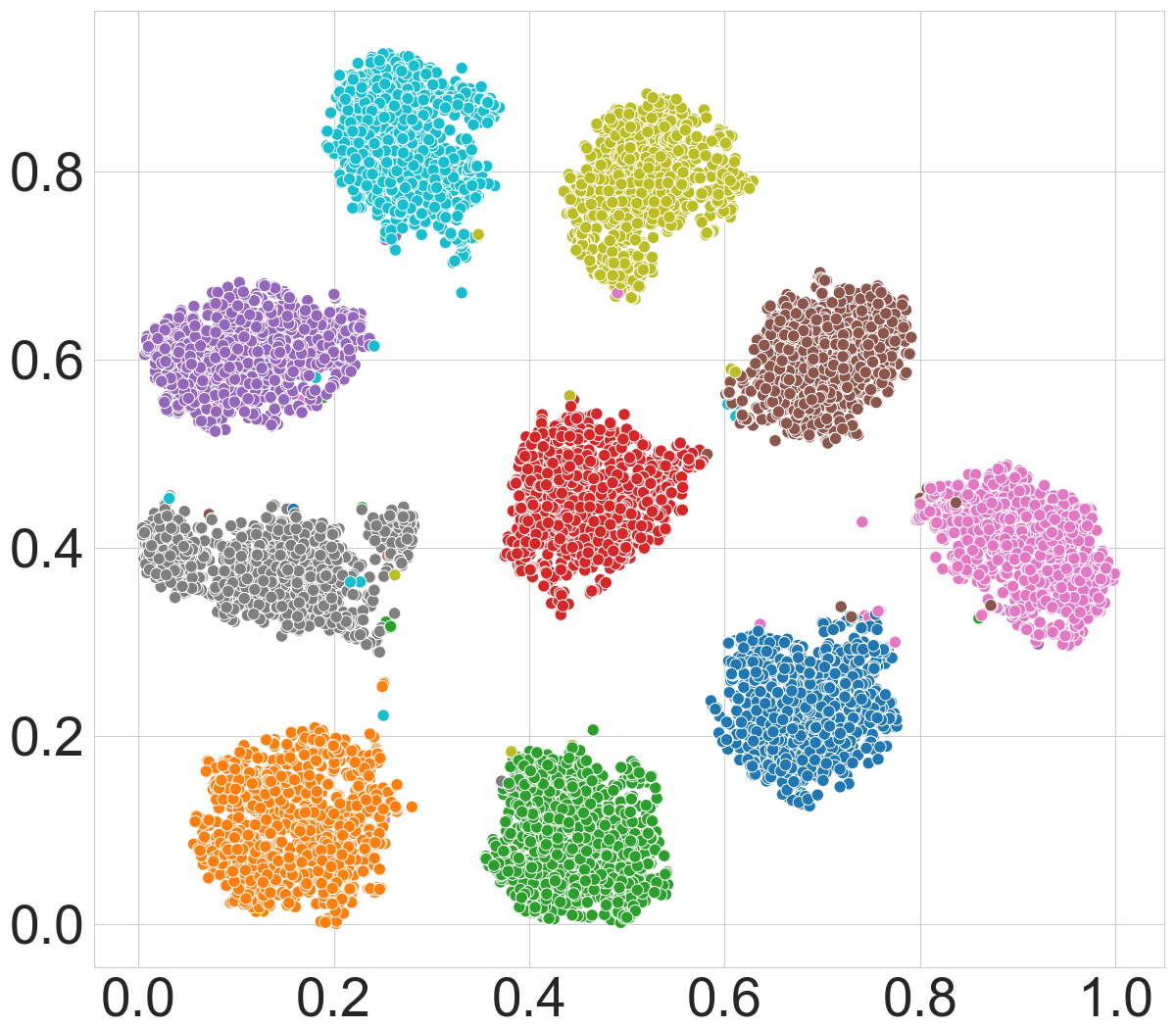}
    }
    \subfigure[AGCE with $\eta=0.4$]{
    \label{ffig:AGCEE-mnist-a-0.4}
    \includegraphics[width=1.2in]{AGCE-0.4.png}
    }
    \caption{Visualization for GCE (top) and AGCE (bottom) on MNIST with different asymmetric noise ($\eta\in[0.1, 0.2, 0.3, 0.4]$) by t-SNE \cite{tsne} 2D embeddings of deep features.}
    \label{ffig:tsne-mnist-a}
\end{figure}

\begin{figure}[tb]
    \centering
    \subfigure[AGCE with different parameters]{
    \label{ffig:AGCE}
    \includegraphics[width=2in]{agce.pdf}
    }
    \subfigure[AUL with different parameters]{
    \label{ffig:AUL}
    \includegraphics[width=2in]{aul.pdf}
    }
    \subfigure[AEL with different parameters]{
    \label{ffig:AEL}
    \includegraphics[width=2in]{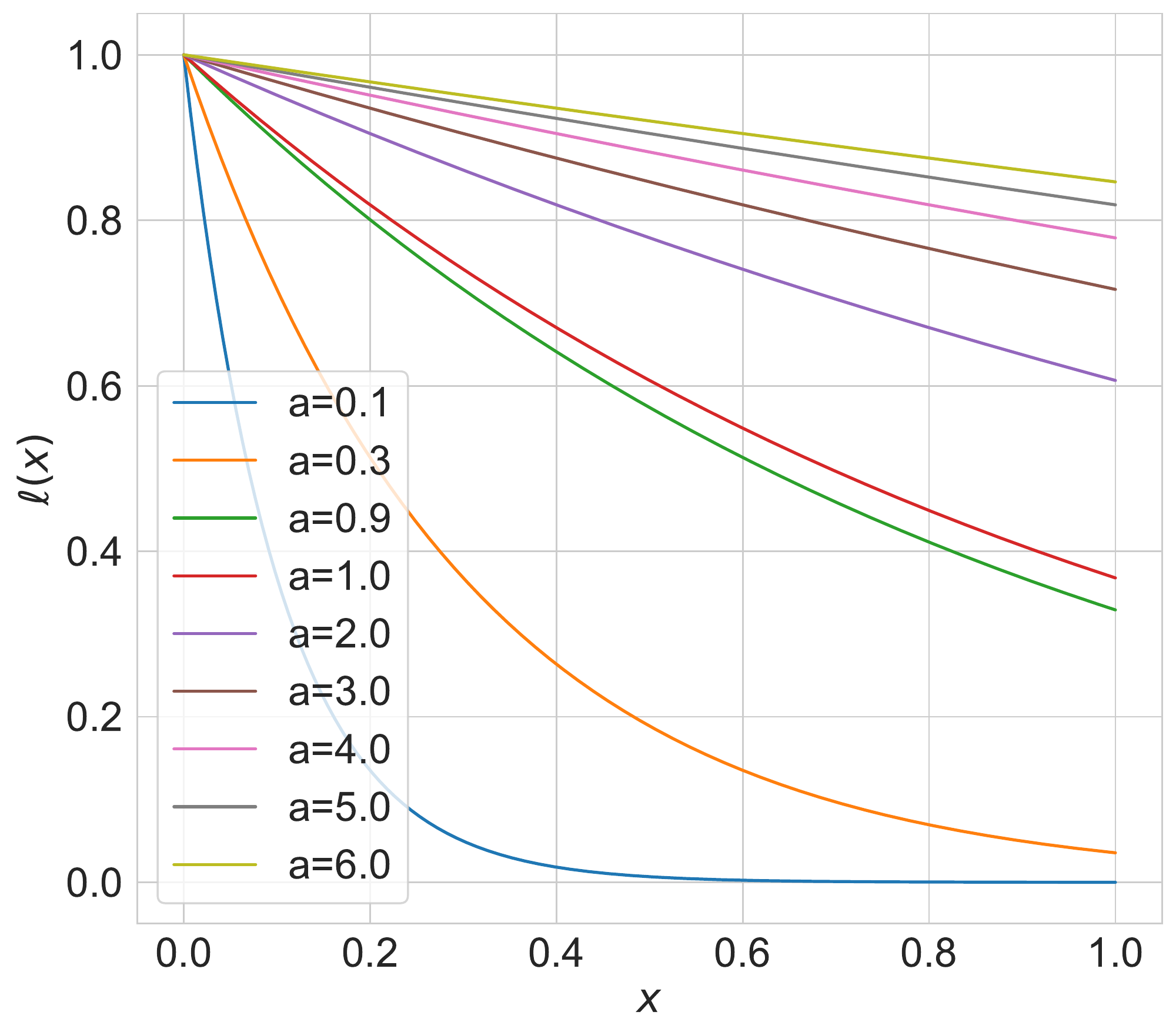}
    }
    \caption{Illustration of asymmetric loss functions.}
    \label{ffig:ALFs}
\end{figure}

\begin{figure}[htbp]
    \centering
    \subfigure[AGCE with $q=0.5$]{
        \label{ffig:AGCE-CIFAR10-a}
        \includegraphics[width=1.5in]{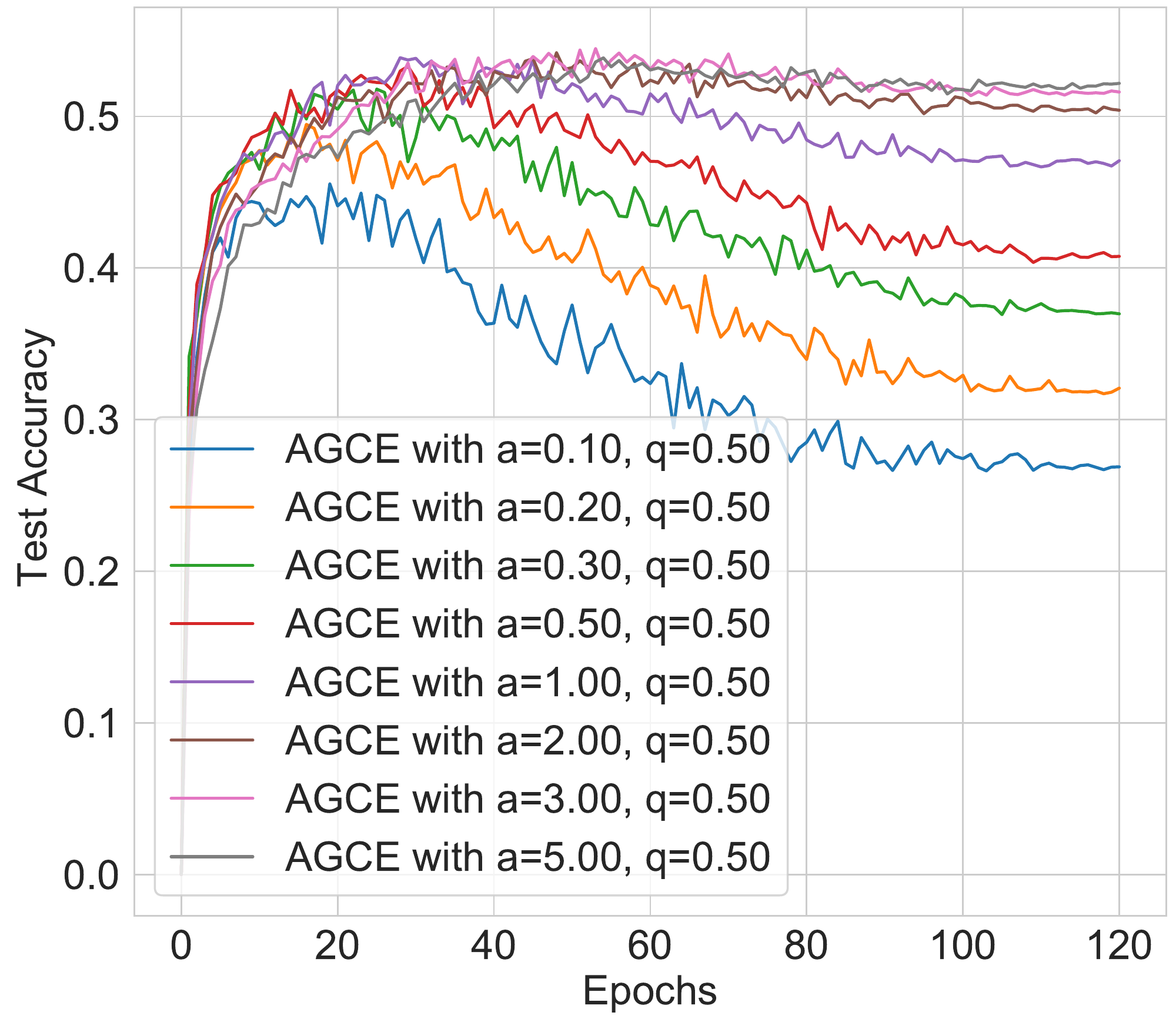}
    }
    \subfigure[AGCE with $q=1.5$]{
        \label{ffig:AGCE-CIFAR10-b}
        \includegraphics[width=1.5in]{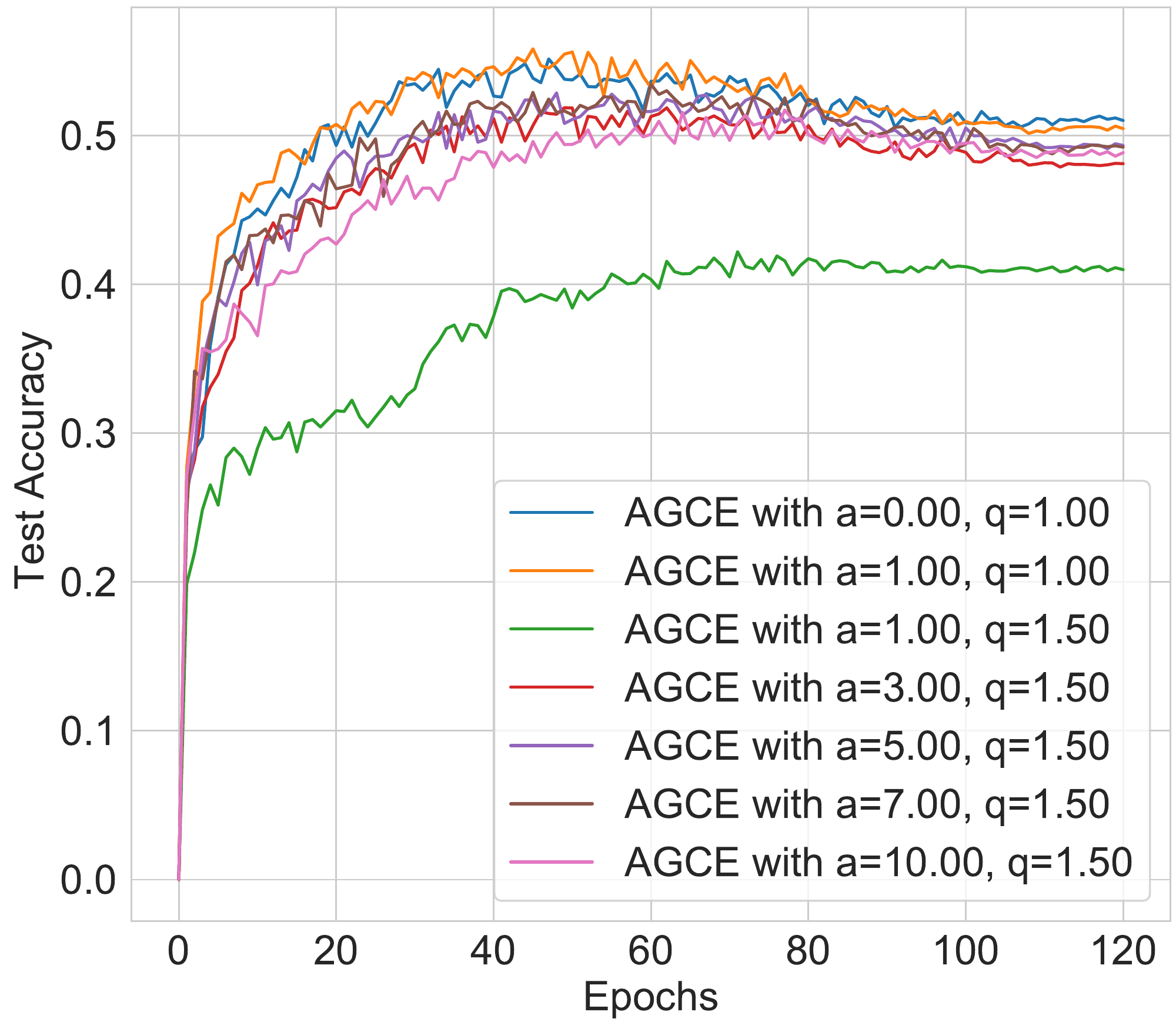}
    }
    \subfigure[AUL with $p<1$]{
        \label{ffig:AUL-CIFAR10-a}
        \includegraphics[width=1.5in]{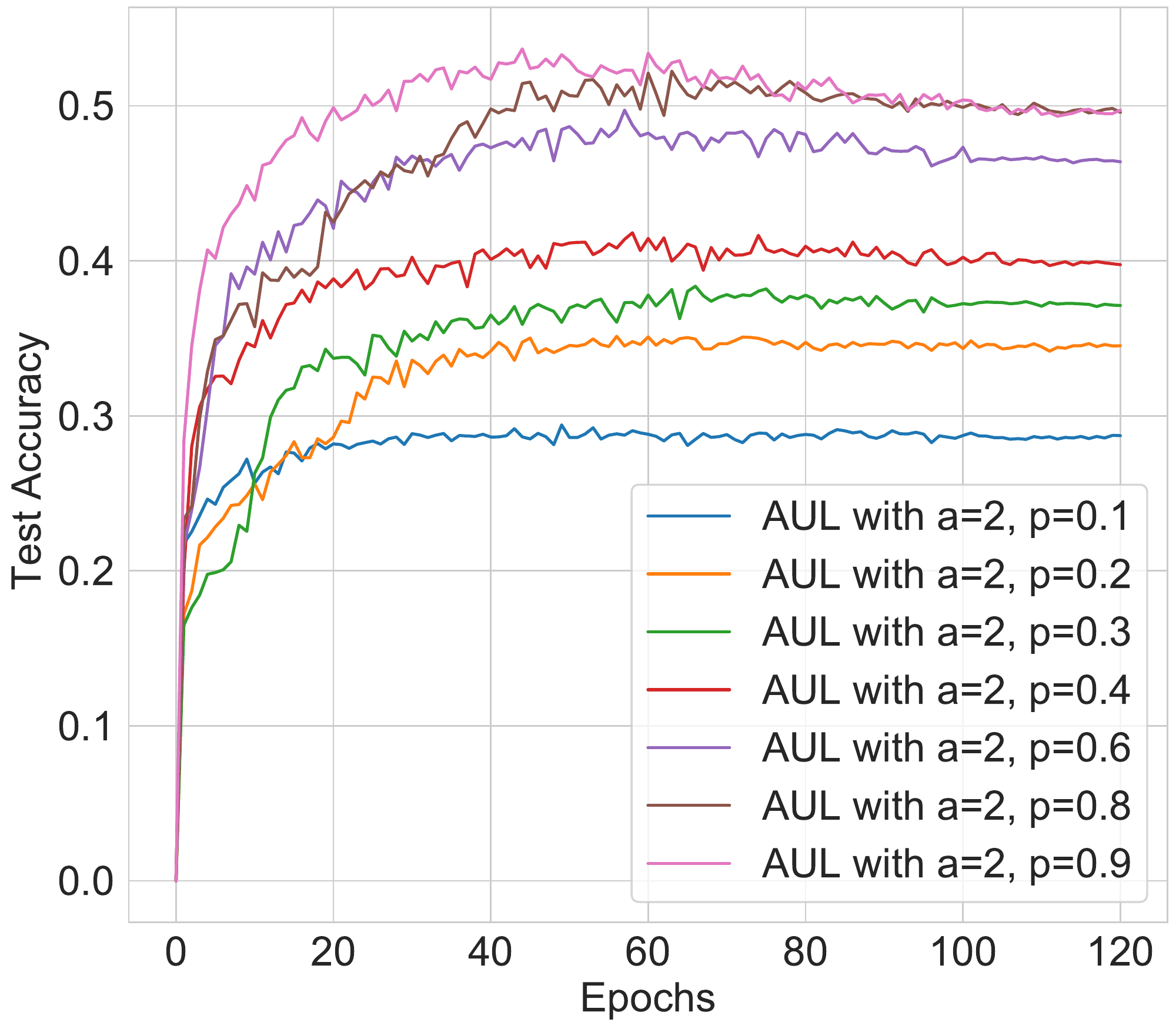}
    }
    \subfigure[AUL with $p=2$]{
        \label{ffig:AUL-CIFAR10-b}
        \includegraphics[width=1.5in]{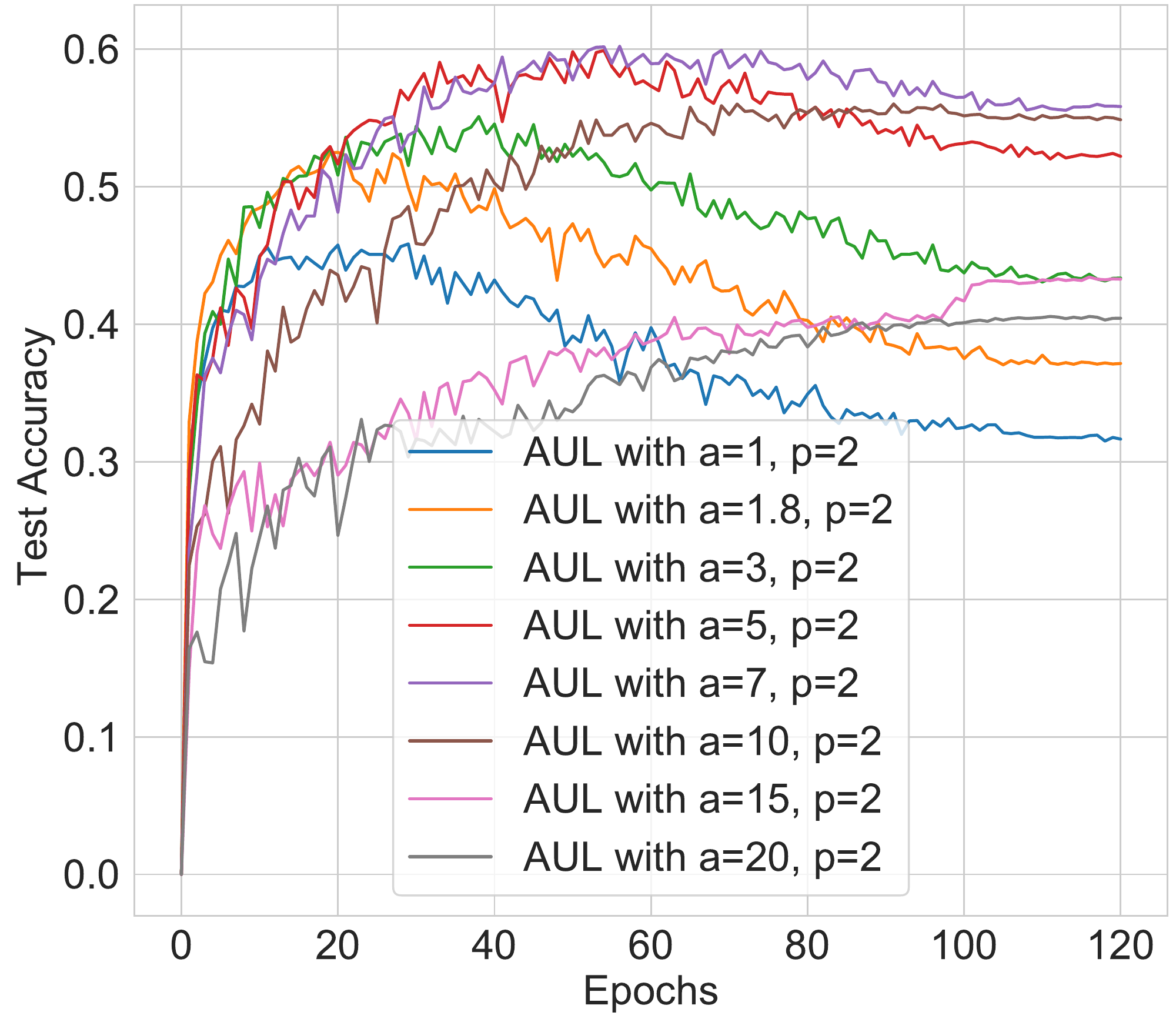}
    }
    \caption{Test accuracies of AGCE and AUL with different parameters on CIFAR-10 under 0.8 symmetric noise.}
    \label{ffig:AUL-CIFAR10}
\end{figure}
\begin{table*}[htb]
\centering
\caption{Test accuracies (\%) of different methods on benchmark datasets with clean or symmetric label noise ($\eta\in[0.2, 0.4, 0.6, 0.8]$). The results (mean$\pm$std) are reported over 3 random runs and the top 3 best results are \textbf{boldfaced}.}
\label{full-symmetric-noise}
\begin{tabular}{c|c|c|cccc}
    \hline
     \multirow{2}*{Datasets} & \multirow{2}*{Methods} & \multirow{2}*{Clean ($\eta=0.0$)} & \multicolumn{4}{c}{Symmetric Noise Rate ($\eta$)}  \\
     ~ & ~ & ~ & 0.2 & 0.4 & 0.6 & 0.8\\
     \hline\hline
     \multirow{14}*{MNIST} & CE & 99.15 $\pm$ 0.05 & 91.62 $\pm$ 0.39 & 73.98 $\pm$ 0.27 & 49.36 $\pm$ 0.43 & 22.66 $\pm$ 0.61\\
    ~ & FL &99.13 $\pm$ 0.09 &91.68 $\pm$ 0.14 &74.54 $\pm$ 0.06 &50.39 $\pm$ 0.28 &22.65 $\pm$ 0.26 \\
    ~ & GCE &99.27 $\pm$ 0.05 &98.86 $\pm$ 0.07 &97.16 $\pm$ 0.03 &81.53 $\pm$ 0.58 &33.95 $\pm$ 0.82\\
    ~ & NLNL &98.61 $\pm$ 0.13 &98.02 $\pm$ 0.14 &97.17 $\pm$ 0.09 &95.42 $\pm$ 0.30 &86.34 $\pm$ 1.43\\
    ~ & SCE &99.23 $\pm$ 0.10 &98.92 $\pm$ 0.12 &97.38 $\pm$ 0.15 &88.83 $\pm$ 0.55 &48.75 $\pm$ 1.54\\
    ~ & NCE &98.60 $\pm$ 0.06 &98.57 $\pm$ 0.01 &98.29 $\pm$ 0.05 &97.65 $\pm$ 0.08 &93.78 $\pm$ 0.41\\
    ~ & NFL &98.51 $\pm$ 0.03 &98.35 $\pm$ 0.07 &98.14 $\pm$ 0.06 &97.48 $\pm$ 0.09 &93.28 $\pm$ 0.40\\
    ~ & NGCE &98.72 $\pm$ 0.05 &98.65 $\pm$ 0.04 &98.42 $\pm$ 0.03 &97.67 $\pm$ 0.12 &94.76 $\pm$ 0.31\\
    ~ & NFL+RCE  &{99.41 $\pm$ 0.06} &\textbf{99.13 $\pm$ 0.07} &98.46 $\pm$ 0.07 &95.53 $\pm$ 0.36 &73.52 $\pm$ 1.39\\
    ~ & NCE+MAE &{99.34 $\pm$ 0.02} &\textbf{99.14 $\pm$ 0.0}5 &98.42 $\pm$ 0.09 &95.65 $\pm$ 0.13 &72.97 $\pm$ 0.34\\
    ~ & NCE+RCE &{99.36 $\pm$ 0.05} &\textbf{99.14 $\pm$ 0.03} &98.51 $\pm$ 0.06 &95.60 $\pm$ 0.21 &74.00 $\pm$ 1.68\\
    \cline{2-7}
      ~ & \textbf{AUL} & {99.14 $\pm$ 0.05} &{99.05 $\pm$ 0.09} &\textbf{98.90 $\pm$ 0.09} &\textbf{98.67 $\pm$ 0.04} &\textbf{96.73 $\pm$ 0.20}\\
    ~ & \textbf{AGCE} &99.05 $\pm$ 0.11 &{98.96 $\pm$ 0.10} &\textbf{98.83 $\pm$ 0.06} &\textbf{98.57 $\pm$ 0.12} &\textbf{96.59 $\pm$ 0.12}\\
    ~ & \textbf{AEL} & 99.03 $\pm$ 0.05 &98.93 $\pm$ 0.06 &\textbf{98.78 $\pm$ 0.13} &\textbf{98.51 $\pm$ 0.06} &\textbf{96.40 $\pm$ 0.11}\\
     \hline\hline
     \multirow{17}*{CIFAR10} & CE & 90.48 $\pm$ 0.11 & 74.68 $\pm$ 0.25 & 58.26 $\pm$ 0.21 & 38.70 $\pm$ 0.53 & 19.55 $\pm$ 0.49\\
     ~ & FL & 89.82 $\pm$ 0.20 & 73.72 $\pm$ 0.08 & 57.90 $\pm$ 0.45 & 38.86 $\pm$ 0.07 & 19.13 $\pm$ 0.28\\
     ~ & GCE & 89.59 $\pm$ 0.26 & 87.03 $\pm$ 0.35 & 82.66 $\pm$ 0.17 & 67.70 $\pm$ 0.45 & 26.67 $\pm$ 0.59\\
     ~ & SCE & 91.61 $\pm$ 0.19 & 87.10 $\pm$ 0.25 & 79.67 $\pm$ 0.37 & 61.35 $\pm$ 0.56 & 28.66 $\pm$ 0.27\\
     ~ & NLNL & 90.73 $\pm$ 0.20 & 73.70 $\pm$ 0.05 & 63.90 $\pm$ 0.44 & 50.68 $\pm$ 0.47 & 29.53 $\pm$ 1.55\\
     ~ & NCE & 75.65 $\pm$ 0.26 & 72.89 $\pm$ 0.25 & 69.49 $\pm$ 0.39 & 62.64 $\pm$ 0.18 & 41.49 $\pm$ 0.66\\
     ~ & NGCE & 80.92 $\pm$ 0.16 & 78.82 $\pm$ 0.09 & 75.52 $\pm$ 0.37 & 69.79 $\pm$ 0.27 & 52.03 $\pm$ 0.88\\
     ~ & NFL & 73.42 $\pm$ 0.35 & 70.93 $\pm$ 0.38 & 67.28 $\pm$ 0.24 & 60.30 $\pm$ 0.75 & 39.07 $\pm$ 0.40\\
     ~ & NFL+RCE &{90.97 $\pm$ 0.19} &{88.89 $\pm$ 0.14} &\textbf{86.03 $\pm$ 0.33} &79.65 $\pm$ 0.41 &\textbf{54.33 $\pm$ 0.80}\\
    ~ & NCE+MAE &{89.17 $\pm$ 0.09} &{86.98 $\pm$ 0.07} &83.74 $\pm$ 0.10 &76.02 $\pm$ 0.16 &46.69 $\pm$ 0.31\\
    ~ & NCE+RCE &{90.87 $\pm$ 0.37} &\textbf{89.25 $\pm$ 0.42} &85.81 $\pm$ 0.08 &\textbf{79.72 $\pm$ 0.20} &\textbf{55.74 $\pm$ 0.95}\\
    \cline{2-7}
     ~ & \textbf{AUL} & 91.27 $\pm$ 0.12 & 89.21 $\pm$ 0.09 & 85.64 $\pm$ 0.19 & 78,86 $\pm$ 0.66 & 52.92 $\pm$ 1.20\\
     ~ & \textbf{AGCE} & 88.95 $\pm$ 0.22 & 86.98 $\pm$ 0.12 & 83.39 $\pm$ 0.17 & 76.49 $\pm$ 0.53 & 44.42 $\pm$ 0.74\\
     ~ & \textbf{AEL} & 86.38 $\pm$ 0.19 & 84.27 $\pm$ 0.12 & 81.12 $\pm$ 0.20 & 74.86 $\pm$ 0.22 & 51.41 $\pm$ 0.32\\
     ~ & \textbf{NCE+AUL} & 91.10 $\pm$ 0.13 & \textbf{89.31 $\pm$ 0.20} & \textbf{86.23 $\pm$ 0.18} & \textbf{79.70 $\pm$ 0.08} & \textbf{59.44 $\pm$ 1.14}\\
     ~ & \textbf{NCE+AGCE} & {90.94 $\pm$ 0.12} & \textbf{89.21 $\pm$ 0.08} & \textbf{86.19 $\pm$ 0.15} & \textbf{80.13 $\pm$ 0.18} & {50.82 $\pm$ 1.46}\\
     ~ & \textbf{NCE+AEL} & {90.71 $\pm$ 0.04} & {88.57 $\pm$ 0.14} & {85.01 $\pm$ 0.38} & {77.33 $\pm$ 0.18} & 47.90 $\pm$ 1.21\\
     \hline\hline
     \multirow{14}*{CIFAR100} & CE &71.33 $\pm$ 0.43	&56.51 $\pm$ 0.39	&39.92 $\pm$ 0.10	&21.39 $\pm$ 1.17	&\ \ 7.59 $\pm$ 0.20\\
     ~ & FL &70.06 $\pm$ 0.70	&55.78 $\pm$ 1.55	&39.83 $\pm$ 0.43	&21.91 $\pm$ 0.89	&\ \ 7.51 $\pm$ 0.09\\
     ~ & GCE &63.09 $\pm$ 1.39	&61.57 $\pm$ 1.06	&56.11 $\pm$ 1.35	&45.28 $\pm$ 0.61	&17.42 $\pm$ 0.06\\
     ~ & SCE &69.62 $\pm$ 0.42	&52.25 $\pm$ 0.14	&36.00 $\pm$ 0.69	&20.14 $\pm$ 0.60	&\ \ 7.67 $\pm$ 0.63\\
     ~ & NLNL & 68.72 $\pm$ 0.60	&46.99 $\pm$ 0.91	& 30.29 $\pm$ 1.64	&16.60 $\pm$ 0.90	&11.01 $\pm$ 2.48\\
     ~ & NCE &29.96 $\pm$ 0.73	&25.27 $\pm$ 0.32	&19.54 $\pm$ 0.52	&13.51 $\pm$ 0.65	&\ \ 8.55 $\pm$ 0.37\\
     ~ & NGCE &22.83 $\pm$ 0.30	&18.96 $\pm$ 1.41	&15.09 $\pm$ 0.64	&11.07 $\pm$ 0.77	&\ \ 6.14 $\pm$ 0.50\\
     ~ & NFL &28.73 $\pm$ 0.08	&23.85 $\pm$ 0.24	&18.96 $\pm$ 0.58	&13.30 $\pm$ 0.80	&\ \ 8.20 $\pm$ 0.16\\
    ~ & NFL+RCE &67.90 $\pm$ 0.40	&64.53 $\pm$ 0.69	&57.85 $\pm$ 0.54	&44.79 $\pm$ 1.00	&\textbf{24.71 $\pm$ 0.93}\\
    ~ & NCE+MAE &67.60 $\pm$ 0.51	&52.30 $\pm$ 0.11	&36.09 $\pm$ 0.55	&18.63 $\pm$ 0.60	&\ \ 7.48 $\pm$ 1.35\\
    ~ & NCE+RCE &68.65 $\pm$ 0.40	&64.97 $\pm$ 0.49	&58.54 $\pm$ 0.13	&45.80 $\pm$ 1.02	&\textbf{25.41 $\pm$ 0.98}\\
     \cline{2-7}
     ~ & \textbf{NCE+AUL} & 68.96 $\pm$ 0.16 & \textbf{65.36 $\pm$ 0.20} & \textbf{59.25 $\pm$ 0.23} & \textbf{46.34 $\pm$ 0.21} & 23.03 $\pm$ 0.64\\
     ~ & \textbf{NCE+AGCE} & 69.03 $\pm$ 0.37 & \textbf{65.66 $\pm$ 0.46} & \textbf{59.47 $\pm$ 0.36} & \textbf{48.02 $\pm$ 0.58} & \textbf{24.72 $\pm$ 0.60}\\
     ~ & \textbf{NCE+AEL} & 68.70 $\pm$ 0.20 & \textbf{65.36 $\pm$ 0.14} & \textbf{59.51 $\pm$ 0.03} & \textbf{46.94 $\pm$ 0.07} & {24.48 $\pm$ 0.24}\\
     \hline
\end{tabular}
\end{table*}

\begin{table*}[htb]
\centering
\caption{Test accuracies (\%) of different methods on benchmark datasets with clean or asymmetric label noise ($\eta\in[0.1, 0.2, 0.3, 0.4]$). The results (mean$\pm$std) are reported over 3 random runs and the top 3 best results are \textbf{boldfaced}.}
\label{full-asymmetric-noise}
\begin{tabular}{c|c|cccc}
    \hline
     \multirow{2}*{Datasets} & \multirow{2}*{Methods} & \multicolumn{4}{c}{Asymmetric Noise Rate ($\eta$)}  \\
     ~ & ~ & 0.1 & 0.2 & 0.3 & 0.4\\
     \hline\hline
     \multirow{14}*{MNIST} & CE & 97.57 $\pm$ 0.22 & 94.56 $\pm$ 0.22 & 88.81 $\pm$ 0.10 & 82.27 $\pm$ 0.40\\
    ~ & FL &97.58 $\pm$ 0.09 &94.25 $\pm$ 0.15 &89.09 $\pm$ 0.25 &82.13 $\pm$ 0.49\\
    ~ & GCE &99.01 $\pm$ 0.04 &96.69 $\pm$ 0.12 &89.12 $\pm$ 0.24 &81.51 $\pm$ 0.19\\
    ~ & NLNL &98.63 $\pm$ 0.06 &98.35 $\pm$ 0.01 &97.51 $\pm$ 0.15 &95.84 $\pm$ 0.26\\
    ~ & SCE &{99.14 $\pm$ 0.04} &98.03 $\pm$ 0.05 &93.68 $\pm$ 0.43 &85.36 $\pm$ 0.17\\
    ~ & NCE &98.49 $\pm$ 0.06 &98.18 $\pm$ 0.12 &96.99 $\pm$ 0.17 &94.16 $\pm$ 0.19\\
    ~ & NFL &98.35 $\pm$ 0.07 &97.86 $\pm$ 0.16 &96.33 $\pm$ 0.21 &92.08 $\pm$ 0.28\\
    ~ & NGCE &98.73 $\pm$ 0.04 &98.67 $\pm$ 0.05 &98.32 $\pm$ 0.11 &97.27 $\pm$ 0.08\\
    ~ & NFL+RCE &\textbf{99.38 $\pm$ 0.02} &98.98 $\pm$ 0.10 &97.18 $\pm$ 0.14 &89.58 $\pm$ 4.81\\
    ~ & NCE+MAE &\textbf{99.32 $\pm$ 0.09} &98.89 $\pm$ 0.04 &96.93 $\pm$ 0.17 &91.45 $\pm$ 0.40\\
    ~ & NCE+RCE &\textbf{99.35 $\pm$ 0.03} &98.99 $\pm$ 0.22 &97.23 $\pm$ 0.20 &90.49 $\pm$ 4.04\\
    \cline{2-6}
     ~ & \textbf{AUL} & {99.15 $\pm$ 0.09} &\textbf{99.15 $\pm$ 0.02} &\textbf{98.98 $\pm$ 0.05} &\textbf{98.62 $\pm$ 0.09}\\
    ~ & \textbf{AGCE} & 99.10 $\pm$ 0.02 &\textbf{99.07 $\pm$ 0.09} &\textbf{98.95 $\pm$ 0.03} &\textbf{98.44 $\pm$ 0.11}\\
    ~ & \textbf{AEL} & 98.99 $\pm$ 0.05 &\textbf{99.06 $\pm$ 0.07} &\textbf{98.90 $\pm$ 0.15} &\textbf{98.34 $\pm$ 0.08}\\
     \hline\hline
     \multirow{17}*{CIFAR10} & CE & 87.55 $\pm$ 0.14 & 83.32 $\pm$ 0.12 & 79.316 $\pm$ 0.59 & 74.67 $\pm$ 0.38\\
     ~ & FL & 86.43 $\pm$ 0.30 & 83.37 $\pm$ 0.07 & 79.33 $\pm$ 0.08 & 74.28 $\pm$ 0.44\\
     ~ & GCE & 88.33 $\pm$ 0.05 & 85.93 $\pm$ 0.23 & 80.88 $\pm$ 0.38 & 74.29 $\pm$ 0.43\\
     ~ & SCE & 89.77 $\pm$ 0.11 & 86.20 $\pm$ 0.37 & 81.38 $\pm$ 0.35 & 75.16 $\pm$ 0.39\\
     ~ & NLNL & 88.54 $\pm$ 0.25 & 84.74 $\pm$ 0.08 & 81.26$\pm$ 0.43 & 76.97 $\pm$ 0.52\\
     ~ & NCE & 74.06 $\pm$ 0.27 & 72.46 $\pm$ 0.32 & 69.86 $\pm$ 0.51 & 65.66 $\pm$ 0.42\\
     ~ & NGCE & 80.18 $\pm$ 0.27 & 79.21 $\pm$ 0.08 & 76.76 $\pm$ 0.07 & 70.10 $\pm$ 1.82\\
     ~ & NFL & 72.28 $\pm$ 0.15 & 70.78 $\pm$ 0.13 & 68.27 $\pm$ 0.43 & 65.09 $\pm$ 0.40\\
     ~ & NFL+RCE & {89.91 $\pm$ 0.17} & {88.24 $\pm$ 0.16} & \textbf{85.81 $\pm$ 0.23} & {79.25 $\pm$ 0.25}\\
     ~ & NCE+MAE &{88.31 $\pm$ 0.20} &86.50 $\pm$ 0.31 & 83.34 $\pm$ 0.39 & 77.14 $\pm$ 0.33\\
     ~ & NCE+RCE & \textbf{90.06 $\pm$ 0.13} & \textbf{88.45 $\pm$ 0.16} & {85.42 $\pm$ 0.09} & \textbf{79.33 $\pm$ 0.15}\\
    \cline{2-6}
     ~ & \textbf{AUL} & 90.19 $\pm$ 0.16 & 88.17 $\pm$ 0.11 & 84.87 $\pm$ 0.04 & 56.33 $\pm$ 0.07 \\
     ~ & \textbf{AGCE} & 88.08 $\pm$ 0.06 & 86.67 $\pm$ 0.14 & 83.59 $\pm$ 0.15 & 60.91 $\pm$ 0.20\\
     ~ & \textbf{AEL} & 85.22 $\pm$ 0.15 & 83.82 $\pm$ 0.15 & 82.43 $\pm$ 0.16 & 58.81 $\pm$ 3.62\\
     ~ & \textbf{NCE+AUL} & \textbf{90.05 $\pm$ 0.20} & \textbf{88.72 $\pm$ 0.26} & \textbf{85.48 $\pm$ 0.18} & \textbf{79.26 $\pm$ 0.05}\\
     ~ & \textbf{NCE+AGCE} & \textbf{90.35 $\pm$ 0.15} & \textbf{88.48 $\pm$ 0.16} & \textbf{85.96 $\pm$ 0.24} & \textbf{80.00 $\pm$ 0.44}\\
     ~ & \textbf{NCE+AEL} & {89.95 $\pm$ 0.04} & {87.93 $\pm$ 0.06} & {84.81 $\pm$ 0.26} & {77.27 $\pm$ 0.11}\\
     \hline\hline
     \multirow{14}*{CIFAR100} & CE &64.85 $\pm$ 0.37	&58.11 $\pm$ 0.32	&50.68 $\pm$ 0.55	&40.17 $\pm$ 1.31\\
     ~ & FL &64.78 $\pm$ 0.50	&58.05 $\pm$ 0.42	&51.15 $\pm$ 0.84	&\textbf{41.18 $\pm$ 0.68}\\
     ~ & GCE &63.01 $\pm$ 1.01	&59.35 $\pm$ 1.10	&53.83 $\pm$ 0.64	&40.91 $\pm$ 0.57\\
     ~ & SCE &61.63 $\pm$ 0.84	&53.81 $\pm$ 0.42	&45.63 $\pm$ 0.07	&36.43 $\pm$ 0.20\\
     ~ & NLNL & 59.55 $\pm$ 1.22 & 50.19 $\pm$ 0.56 & 42.81 $\pm$ 1.13 & 35.10 $\pm$ 0.20\\
     ~ & NCE &27.59 $\pm$ 0.54	&25.75 $\pm$ 0.50	&24.28 $\pm$ 0.80	&20.64 $\pm$ 0.40\\
     ~ & NGCE &20.89 $\pm$ 0.52	&19.28 $\pm$ 0.23	&17.77 $\pm$ 2.32	&13.15 $\pm$ 2.90\\
     ~ & NFL &26.46 $\pm$ 0.31	&25.39 $\pm$ 0.87	&23.18 $\pm$ 0.80	&20.10 $\pm$ 0.21\\
     ~ & NFL+RCE &65.97 $\pm$ 0.18	&62.77 $\pm$ 0.31	&\textbf{55.60 $\pm$ 0.25}	&\textbf{41.66 $\pm$ 0.20}\\
     ~ & NCE+MAE &60.22 $\pm$ 0.37	&52.20 $\pm$ 0.41	&44.50 $\pm$ 0.46	&35.82 $\pm$ 0.27\\
     ~ & NCE+RCE &66.38 $\pm$ 0.16	&\textbf{62.97 $\pm$ 0.24}	&\textbf{55.38 $\pm$ 0.49}	&\textbf{41.68 $\pm$ 0.56}\\
     \cline{2-6}
     ~ & \textbf{NCE+AUL} & \textbf{66.62 $\pm$ 0.09} & \textbf{63.86 $\pm$ 0.18} & 50.38 $\pm$ 0.32 & 38.59 $\pm$ 0.48\\
     ~ & \textbf{NCE+AGCE}& \textbf{67.22 $\pm$ 0.12} & \textbf{63.69 $\pm$ 0.19} & \textbf{55.93 $\pm$ 0.38} & \textbf{43.76 $\pm$ 0.70}\\
     ~ & \textbf{NCE+AEL} & \textbf{66.92 $\pm$ 0.22} & 62.50 $\pm$ 0.23 & {52.42 $\pm$ 0.98} & 39.99 $\pm$ 0.12
     \\
     \hline
\end{tabular}
\end{table*}

\end{document}